\documentclass{article}
\usepackage[preprint]{neurips_2025_custom}
\usepackage{fix-cm}

\usepackage{xcolor}         
\definecolor{linkColor}{rgb}{0.2,0.4,0.6}
\definecolor{myblue}{HTML}{0379AC}
\definecolor{myred}{HTML}{A50E50}
\usepackage[utf8]{inputenc} 
\usepackage[T1]{fontenc}    
\usepackage[table]{xcolor}
\usepackage[colorlinks=true,linkcolor=linkColor,citecolor=linkColor,filecolor=linkColor,urlcolor=linkColor]{hyperref}       
\usepackage{url}            
\usepackage{booktabs}       
\usepackage{amsfonts}       
\usepackage{nicefrac}       
\usepackage{microtype}      
\usepackage{graphicx}
\graphicspath{{./}{Figures/}}
\usepackage{booktabs}
\usepackage{multirow}
\usepackage{caption}
\usepackage{subcaption}
\usepackage{makecell}
\usepackage{csquotes}
\usepackage{epigraph}

\usepackage{multirow}
\usepackage{amsmath}
\usepackage{capt-of}
\usepackage{needspace}
\usepackage{placeins}
\usepackage{tabularx}
\usepackage{epsfig}
\usepackage{amssymb}
\usepackage{amsfonts}
\usepackage{booktabs}
\usepackage{scalerel}
\usepackage[inline]{enumitem}
\usepackage{listings}
\usepackage{fvextra}
\usepackage{varwidth}
\usepackage[export]{adjustbox}
\usepackage{tikz}
\usetikzlibrary{tikzmark, positioning, shapes.geometric, arrows.meta, fit, calc}

\usepackage{stmaryrd}
\usepackage{bbm}
\usepackage{wrapfig}
\usepackage{pifont}
\usepackage[noabbrev]{cleveref}

\usepackage{algorithm}
\usepackage[noend]{algpseudocode}

\definecolor{deepblue}{rgb}{0,0,0.5}
\definecolor{officeblue}{RGB}{0,102,204}
\definecolor{deepred}{rgb}{0.6,0,0}
\definecolor{deepgreen}{rgb}{0,0.5,0}
\definecolor{mybrickred}{RGB}{182,50,28}

\definecolor{fillcolor}{RGB}{216,217,252}

\newcommand*\AlgCommentInLine[1]{{\color{deepblue}{$\triangleright$ \textit{#1}}}}

\usepackage{etoolbox}
\usepackage{framed}
\tcbuselibrary{breakable,skins}

\newif\ifxetexorluatex
\ifxetex
  \xetexorluatextrue
\else
  \ifluatex
    \xetexorluatextrue
  \else
    \xetexorluatexfalse
  \fi
\fi
\newcommand*\quotesize{60} 
\newcommand*{\openquote}
   {\tikz[remember picture,overlay,xshift=-4ex,yshift=-2.5ex]
   \node (OQ) {\fontsize{\quotesize}{\quotesize}\selectfont``};\kern0pt}

\newcommand*{\closequote}[1]
  {\tikz[remember picture,overlay,xshift=4ex,yshift={#1}]
   \node (CQ) {\fontsize{\quotesize}{\quotesize}\selectfont''};}

\colorlet{shadecolor}{white}
\definecolor{appreportnavy}{HTML}{1A365D}
\definecolor{appreportaccent}{HTML}{2563EB}
\definecolor{appreportlight}{HTML}{F8FAFC}
\definecolor{appreportline}{HTML}{CBD5E1}
\definecolor{appreporttext}{HTML}{1E293B}

\newcommand*\shadedauthorformat{\emph} 

\tcbset{
  appendixreportbox/.style={
    enhanced,
    breakable,
    colback=white,
    colframe=appreportline,
    boxrule=0.8pt,
    arc=1mm,
    left=7mm,
    right=7mm,
    top=6mm,
    bottom=6mm,
    borderline west={1.2pt}{0pt}{appreportnavy!75},
    before upper={\parindent0pt\setlength{\parskip}{0.65em}\small\linespread{1.05}\selectfont},
  },
  appendixreportheader/.style={
    enhanced,
    colback=appreportnavy,
    colframe=appreportnavy,
    boxrule=0pt,
    arc=1mm,
    left=6mm,
    right=6mm,
    top=4mm,
    bottom=4mm,
  },
  appendixreportpanel/.style={
    enhanced,
    colback=appreportlight,
    colframe=appreportline,
    boxrule=0.5pt,
    arc=0.8mm,
    left=5mm,
    right=5mm,
    top=4mm,
    bottom=4mm,
  }
}

\newcommand{\appendixreportsectiontitle}[1]{%
  \par\medskip
  {\color{appreportline}\rule{\linewidth}{0.5pt}}\par
  \vspace{0.5em}
  {\normalsize\sffamily\bfseries\color{appreportnavy}#1\par}
  \vspace{0.4em}
}

\newcommand*\authoralign[1]{%
  \if#1l
    \def\authorfill{}\def\quotefill{\hfill}
  \else
    \if#1r
      \def\authorfill{\hfill}\def\quotefill{}
    \else
      \if#1c
        \gdef\authorfill{\hfill}\def\quotefill{\hfill}
      \else\typeout{Invalid option}
      \fi
    \fi
  \fi}
{\authoralign{#1}
\ifblank{#2}
   {\def\shadequoteauthor{}\def\yshift{-2ex}\def\quotefill{\hfill}}
   {\def\shadequoteauthor{\par\authorfill\shadedauthorformat{#2}}\def\yshift{2ex}}
\begin{snugshade}\begin{quote}\openquote}
{\shadequoteauthor\quotefill\closequote{\yshift}\end{quote}\end{snugshade}}

\newfloat{Code}{htbp}{Code}

\usepackage{amsmath,amsfonts,bm}

\def\eqref#1{equation~\ref{#1}}

\def\1{\bm{1}}

\DeclareMathAlphabet{\mathsfit}{\encodingdefault}{\sfdefault}{m}{sl}
\SetMathAlphabet{\mathsfit}{bold}{\encodingdefault}{\sfdefault}{bx}{n}

\newcommand{\nomad}[0]{Nomad}

\title{\nomad: Autonomous Exploration and Discovery}

\author{
\textbf{Bokang Jia}$^{1}$ \quad
\textbf{Samta Kamboj}$^{1}$ \quad
\textbf{Satheesh Katipomu}$^{1}$\\
\textbf{Seung Hun Han}\thanks{Work done during internship at Inception, G42.}\hspace{0.25em}$^{3}$ \quad
\textbf{Neha Sengupta}$^{1}$ \quad
\textbf{Andrew Jackson}$^{2}$\\
$^{1}$Inception, G42 \qquad
$^{2}$G42 \qquad
$^{3}$MBZUAI
}

\begin{document}

\maketitle

\begin{abstract}
We introduce \nomad{}, a system for autonomous data exploration and insight discovery. Given a corpus of documents, databases, or other data sources, users rarely know the full set of questions, hypotheses, or connections that could be explored. As a result, query-driven question answering and prompt-driven deep-research systems remain limited by human framing and often fail to cover the broader insight space.
\nomad{} addresses this problem with an exploration-first architecture. It constructs an explicit Exploration Map over the domain and systematically traverses it to balance breadth and depth. It generates and selects hypotheses and investigates them with an explorer agent that can use document search, web search, and database tools. Candidate insights are then checked by an independent verifier before entering a reporting pipeline that produces cited reports and higher-level meta-reports.
We also present a comprehensive evaluation framework for autonomous discovery systems that measures trustworthiness, report quality, and diversity. Using corpora of selected UN and WHO reports and arXiv papers on LLM agents, we show that \nomad{} produces reports with strong numeric grounding, higher overall quality and actionability than baselines, and more diverse insights over several runs.
\nomad{} is a step toward autonomous systems that not only answer user questions or conduct directed research, but also discover which questions, research directions, and insights are worth surfacing in the first place.
\end{abstract}

\section{Introduction}

\begin{figure*}[t]
\centering
\includegraphics[width=0.95\textwidth]{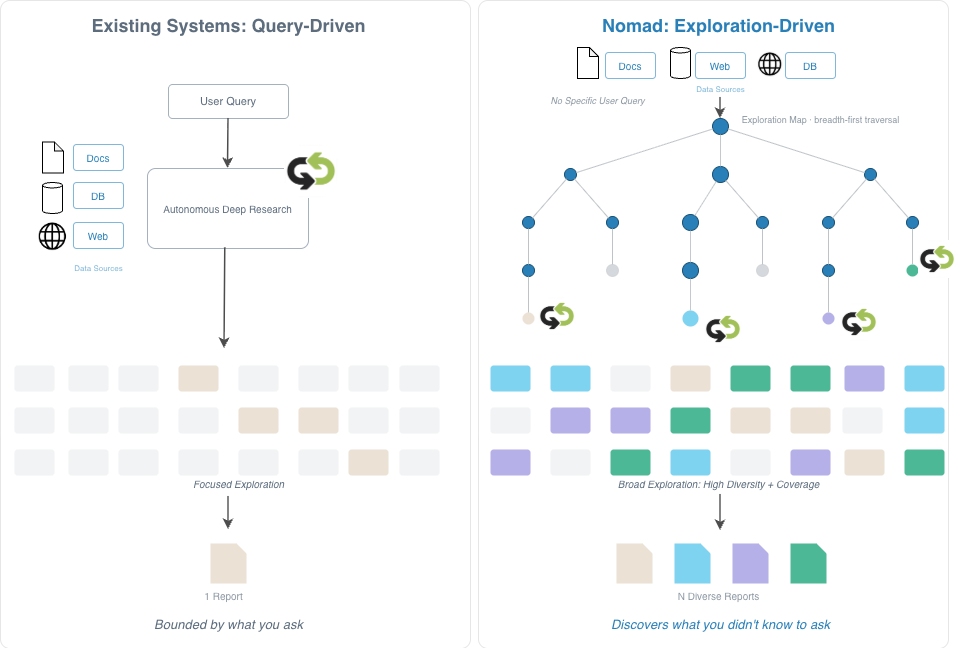}
\caption{Query-driven systems (left) explore only the narrow slice of a corpus that a user thinks to ask about, producing a single report shaped by the initial question. \nomad{} (right) constructs an Exploration Map over the domain and systematically traverses it, generating and verifying hypotheses across the full breadth of the corpus. This exploration-first approach yields diverse, evidence-backed reports covering directions the user may never have thought to pursue.}
\label{fig:hero}
\end{figure*}

Large document collections contain far more potential insights than a human analyst can enumerate as explicit questions. This is true even for corpora with only a few hundred documents. A user may know some of the questions to ask, but rarely knows the full space of useful questions, hidden connections, or contradictions in advance. As a result, the value extracted from a corpus is often bounded by the analyst's initial framing rather than by the information present in the corpus itself.

Recent agentic research systems have made strong progress on long-horizon tool use, web search, and report generation \citep{zheng2025deepresearcher,li2025searcho1,he2025openwebvoyager,shao2024storm,openai2025deepresearch,gptresearcher2025deepresearch}. However, these systems are still mainly organized around a user-specified task, query, or browsing goal. This framing is appropriate for question answering and deep research assistants, but it is not sufficient for autonomous discovery. If the system follows only user-provided directions, the burden remains on the user to know which directions matter. On the other hand, if an LLM-driven system follows only its own highest-probability reasoning path, it tends to revisit obvious directions and leave large parts of the corpus unexplored. We therefore study a different problem: how to build a system that autonomously explores a corpus, expands the space of candidate research directions, and produces many distinct, evidence-backed insights (Figure~\ref{fig:hero}).

This problem introduces three core challenges. The first is \emph{coverage}. The space of possible topics, hypotheses, and cross-document connections is combinatorial, so naive sampling does not explore it well. The second is \emph{verifiability}. A discovery system must not only generate interesting claims; it must also filter unsupported or weakly grounded ones. The third is \emph{evaluation}. A system designed for repeated discovery cannot be judged only by whether one report looks plausible. It must also be evaluated for diversity across runs, trustworthiness of its evidence, and the quality of the reports it produces.

In this paper we introduce \nomad{}, a system for autonomous data exploration and discovery. \nomad{} addresses coverage by constructing an \emph{Exploration Map}, a structured search space over topics, concepts, hypotheses, and insights. The map can be built top-down from a goal, bottom-up from a corpus, and expanded over time as new data becomes available. Exploration then proceeds by breadth-first topic selection, so the system covers heterogeneous parts of the domain before spending more depth on any one branch. At the frontier of exploration, \nomad{} generates new hypotheses, selects promising ones, and investigates them with an explorer agent that can use web search, document search, and database tools.

A second key design choice is to separate discovery from verification. \nomad{} couples the explorer with an independent verifier that revisits proposed insights using the same data sources and rejects or refines weak claims. Only verified insights are sent to the reporting stack. From there, the system generates self-contained reports with citations, evaluates them for trustworthiness and quality, and periodically synthesizes multiple reports into higher-level meta-reports. This design lets \nomad{} treat discovery as an end-to-end pipeline: search and reasoning produce candidate insights, verification filters them, and reporting turns them into usable artifacts for human readers.

\nomad{} builds on several active lines of work, including research agents, corpus structure induction, grounded long-form writing, and factuality control \citep{edge2024graphrag,kargupta2025taxoadapt,shao2024storm,zhang2025longcite}. The key difference is the objective. Prior systems usually optimize performance on a given task. \nomad{} instead optimizes for diversity, novelty, coverage, and verifiability over a shared corpus with an unspecified or underspecified task. This shift in objective determines the architecture. The exploration map acts as the control state for exploration, driving diversity and coverage. The explorer drives novelty, while the verifier actively helps to generate more verifiable and trustworthy insights. The report generator drives quality of final outcomes presented to readers.

\paragraph{Contributions.}
\begin{itemize}
    \item We formulate autonomous data exploration as a task distinct from query-driven question answering and deep research, and present \nomad{}, an end-to-end system built for repeated insight discovery over a corpus.
    \item We introduce the Exploration Map, together with topic selection, hypothesis generation, and web-based expansion mechanisms that balance breadth and depth under fixed compute budgets.
    \item We introduce an explorer--verifier loop that couples autonomous investigation with explicit evidence checking, and connect it to a reporting stack that generates cited reports and higher-level meta-reports.
    \item We propose an evaluation framework for autonomous discovery systems that measures trustworthiness, diversity, report quality, and redundancy at both section and report levels, and evaluate \nomad{} across two domains and two settings. We further ablate the explorer model, verifier model, and hypothesis-generation module.
\end{itemize}

The remainder of the paper describes the architecture of \nomad{}, its reporting and synthesis pipeline, and the evaluation framework used to assess it.

\begin{table}[t]
\centering
\small
\caption{A few example report titles generated by each system on the UN and WHO documents in the bottom-up setting. \nomad{} produces topically diverse titles, whereas \texttt{o3-deep-research} and GPT Researcher converge on similar themes and framings.}
\label{tab:titles-who}
\begin{tabular}{p{0.96\linewidth}}
\toprule
\textbf{\nomad{}} \\
\midrule
Financing Healthier Electric Transport \\
Cyclones, Surveillance, and Hunger \\
Malaria's Disability Data Gap \\
Shifting Stocks, Strained Seas \\
Stateless Women's Twin Exposure \\
Data Readiness Boosts Funding \\
\midrule
\textbf{o3-deep-research} \\
\midrule
A Converging Polycrisis: Reversals in Global Health, Climate, and Development \\
Converging Crises and a Global Development Backslide: An Underappreciated Cross-Domain Risk \\
Interconnected Crises Threaten a ``Lost Decade'' of Global Progress \\
Converging Crises: A Polycrisis Undermining Global Progress \\
Navigating an Underappreciated Polycrisis: Reversals and Risks Across Health, Climate, Economy, and Technology \\
A ``Polycrisis'' Reversing Global Progress Across Health, Climate, and Development \\
\midrule
\textbf{GPT Researcher} \\
\midrule
Rural Digital Bottlenecks as a Systemic Risk to Health Security, Climate-Smart Agriculture and Inclusive Growth \\
Convergence Risk 2026: How Rapid Digitalisation, Climate-Accelerated Urbanisation and Weak One-Health Systems Are Creating a New Category of Compound Threats \\
Under-the-Radar but Systemic: The Global Financing Gap for One Health and Antimicrobial-Resistance Control and its Spill-over Risks for Trade, Food Security and Sustainable Development \\
Emerging Polycrises at the Intersection of Digitalisation, Gender and Climate-Sensitive Health Systems \\
Converging Crises: Antimicrobial Resistance in a Warming, Fragmenting World \\
Converging Pressures: How Climate-Aligned Trade Measures and Digital Disparities Exacerbate Post-Pandemic Health and Food-Security Risks in Developing Countries \\
\bottomrule
\end{tabular}
\end{table}

\FloatBarrier

\section{Related Work}
\label{sec:related-work}

\paragraph{Agentic research and information-seeking systems.}
Recent work has shown that LLM agents benefit from interleaving reasoning with external actions. ReAct established the basic observe--reason--act pattern for tool-using language agents \citep{yao2023react}. Later systems push this pattern toward open-web research, search-enhanced reasoning, and autonomous task discovery, including DeepResearcher, WebDancer, Search-o1, Search-R1, OpenWebVoyager, and Auto-Intent \citep{zheng2025deepresearcher,wu2025webdancer,li2025searcho1,jin2025searchr1,he2025openwebvoyager,kim2024autointent}. Open and commercial systems now package similar capabilities directly as deep-research products, including GPT Researcher, Open Deep Research, OpenAI Deep Research, Gemini Deep Research, and Perplexity Research \citep{gptresearcher2025deepresearch,huggingface2025opendeepresearch,openai2025deepresearch,google2024geminideepresearch,perplexity2026researchmode}. These systems show that multi-step tool use, web search, and source-backed synthesis are central to modern research assistants. However, they are still organized around executing a user-specified topic or request. Robin moves slightly closer to autonomous discovery, but it is specialized for scientific workflows with domain-specific experimental steps and requires a specific target in the input \citep{futurehouse2025robin}. ARI broadens deep research toward enterprise use over mixed public and private data, but remains framed as an analyst-style product rather than an exploration-first discovery engine \citep{youcom2025ari}. \nomad{} differs in objective: it treats a corpus as an open discovery space and explicitly seeks many diverse, non-obvious insights rather than only the best answer to one request.

\paragraph{Structured exploration over corpora.}
Another line of work argues that broad coverage requires explicit corpus structure. GraphRAG and HippoRAG build graph-based representations that connect entities, passages, and global summaries across a collection \citep{edge2024graphrag,gutierrez2024hipporag}. Work on hierarchical taxonomy induction further shows that LLMs can organize large scientific corpora into multi-level topic structures, both at construction time and as the corpus evolves \citep{zhu2025contexttaxonomy,kargupta2025taxoadapt}. These papers motivate \nomad{}'s exploration map, but their main goal is indexing, retrieval, or taxonomy quality. In \nomad{}, the map is not only a representation of the corpus. It is the control state for exploration: topic selection, breadth-first coverage, hypothesis generation, and later synthesis all operate over the same shared structure.

\paragraph{Retrieval and long-form synthesis.}
A third line studies how systems should retrieve evidence and turn it into long-form outputs. Self-RAG learns when to retrieve, when to critique, and when to revise generation \citep{asai2024selfrag}. STORM studies grounded article writing through research, outline generation, and drafting \citep{shao2024storm}. GPT Researcher and Open Deep Research package related retrieve-and-synthesize loops into open deep-research systems, while Elicit Reports specializes this workflow to auditable literature review over scientific papers \citep{gptresearcher2025deepresearch,huggingface2025opendeepresearch,elicit2025reports}. RAPID and LongWriter focus on planning, information discovery, and coherent long-form generation once relevant evidence is available \citep{gu2025rapid,bai2024longwriter}. This work is directly relevant to \nomad{}'s document-search, report-generation, and meta-report modules. The difference is again the system boundary. Prior work in this area typically assumes that the writing task or research question is already given. \nomad{} first discovers candidate insights through autonomous exploration and only then writes reports and cross-insight syntheses for the subset that survive verification.

\paragraph{Verification, citation grounding, and evaluation.}
Another line focuses on making long-form outputs auditable and measurable. Chain-of-Verification inserts explicit checking steps into generation \citep{dhuliawala2024cov}. LongCite studies fine-grained citation generation, DnDScore decomposes long-form claims for factuality verification, and ReportEval benchmark introduced in \cite{fan2025understanding} evaluates deep-research reports along quality, redundancy, and factuality dimensions, including claim--source checking and paragraph-pair redundancy judgments \citep{zhang2025longcite,wanner2025dndscore,fan2025understanding}. Complementary evaluation work decomposes outputs into atomic facts or uses LLM judges to better align automatic scoring with human judgments \citep{min2023factscore,liu2023geval}. These papers motivate our verifier, citation auditing, and report evaluation stack. Their typical use, however, is to improve or score answers for a fixed task after the main reasoning path has already been chosen. \nomad{} makes verification part of discovery itself: an explorer proposes candidate insights, an independent verifier re-checks them against the same evidence sources, and only verified outputs enter the reporting pipeline.

\FloatBarrier

\section{Nomad}
\label{sec:nomad}

This section presents the design and implementation of \nomad{}, our system for autonomous data exploration and discovery. We first summarize the architecture and core components, then describe each module in detail.

\paragraph{Illustrative Example:} To illustrate the outcomes of the individual modules, we use a running example of a \nomad{} instance representing a WHO Research Analyst, tasked with surfacing non-obvious insights, including but not limited to emerging trends and risks in health systems and related areas. We optionally provide the system with a corpus of curated UN and WHO reports, and it can also use Web search for both exploration map construction and expansion, as well as for evidence retrieval during exploration and verification.
The provided corpus contains official UN and WHO policy and statistical reports spanning health, climate, development, trade, labour, migration, food security, education, and gender. We assembled it from publicly available PDFs on official UN and WHO sources using curated report lists and selected WHO topic pages. We use a public WHO/UN corpus in this paper so that the running example can be shared and inspected. We have also used the same workflow on confidential enterprise document collections, but do not include those materials here because they contain non-public business information.

For the second evaluation domain, we use an arXiv Researcher instance tasked with analyzing research on agentic LLM systems to surface novel insights, identify gaps in current approaches, and highlight unexplored research directions. Its corpus contains 50 publicly available arXiv papers on LLM-based agents, with arXiv identifiers spanning November 2025 through March 2026. The papers cover agent planning and reasoning, multi-agent systems, tool use and orchestration, memory and retrieval, safety and security, evaluation, reinforcement learning, and domain-specific applications.

\subsection{Overview}

\nomad{} operates autonomously over a corpus to generate many distinct, high-quality insights. A central challenge is that LLM-driven agents concentrate on high-probability responses, producing redundant observations even with stochastic decoding over several hundred generations. We therefore include explicit mechanisms for broad exploration, rigorous verification, and non-obvious discovery.

\begin{figure}[ht]
    \centering
    \includegraphics[width=0.95\linewidth]{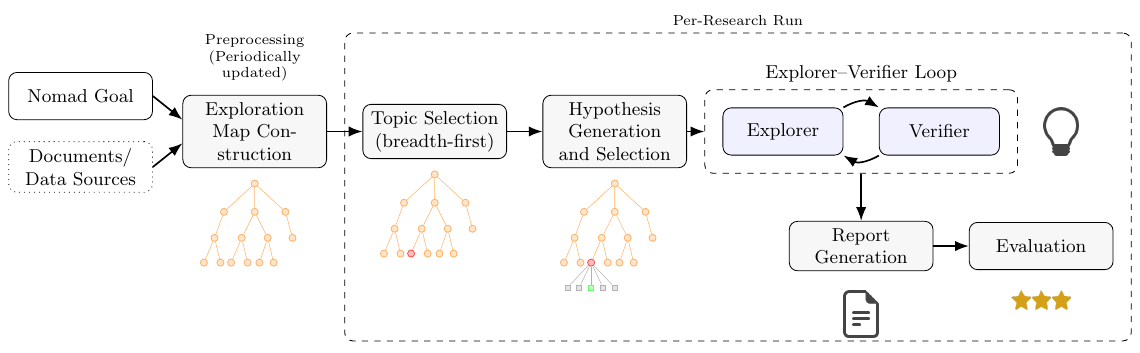}
    \caption{Overview of the \nomad{} pipeline. The exploration map and topic selection feed hypothesis generation, which enters the explorer--verifier loop. Verified insights are reported and evaluated, with periodic synthesis in meta-reports.}
    \label{fig:nomad-overview}
\end{figure}

\noindent\textbf{Key Goals:} Diversity, Verifiability, Novelty. These goals shape the architecture and workflow. Diversity is implemented through an exploration map: a tree-structured search space traversed breadth-first to cover heterogeneous regions of the corpus before committing to depth \citep{edge2024graphrag,kargupta2025taxoadapt}. Verifiability is enforced by a verifier agent that evaluates candidate insights from the explorer, filtering unsupported claims and elevating evidence-backed findings. Novelty is encouraged by a hypothesis-generation module that prompts non-obvious hypotheses and connections beyond surface-level retrieval. Together, these mechanisms balance coverage, validation, and creative hypothesis formation.

Given a high-level goal and/or a set of documents, \nomad{} begins by constructing an \emph{Exploration Map}. This is a one-time preprocessing step that captures the space of topics to explore in the corpus. Although potentially expensive, it enables large-scale and efficient exploration in the insight generation phase, with more meaningful outcomes obtained by each run of \nomad{}.
The Topic Tree, a subgraph of the exploration map, provides a hierarchical view of the domain, with main subtopics at the first level and finer subtopics at deeper levels. It is built using LLM-based topic extraction and clustering over available documents, and it serves as the backbone of exploration. The Topic Tree serves two primary purposes. First, it allows the system to prioritize exploration across different topics and subtopics, thus promoting diversity and coverage. Second, it provides a structured framework for organizing insights and hypotheses, enabling the system, as well as the human users consuming its outcomes, to build a coherent mental map of the domain as it explores.

Once the exploration map (and therefore, the Topic Tree) is constructed, insight generation begins with topic selection. The topic selection module moves level-wise down the Topic Tree, selecting the least explored subtopic at each level. This breadth-first traversal promotes diversity by exploring widely before going deep.

At a leaf node, the system checks for unexplored hypotheses. If none exist, the hypothesis generation module creates new ones. Depending on configuration, it performs Web or document search for the topic, then proposes a batch of hypotheses from the retrieved evidence. Each hypothesis is graded for impact and relevance, and also scored for diversity against previously explored hypotheses. The hypotheses are added as children of the topic node and thus to the `unexplored pool'. The best hypothesis is selected for exploration, and the system enters the explorer--verifier loop.

The explorer--verifier loop is the core of insight generation. The explorer runs a ReAct-style loop and can issue multiple tool calls per step (Web search, database queries, or document queries). It forms observations, decides next actions, and may revise the hypothesis. When it reaches a sufficiently interesting, novel, and \emph{surprising} insight, it submits it for verification. The verifier checks evidence support using the same data sources, runs a similar loop, and assigns a score. If the score passes, the loop exits. Otherwise, verifier feedback is appended to the explorer history and the loop repeats. This process iteratively refines insights and filters unsupported claims.

At this stage, the insight is typically short and dependent on the preceding conversation. To make it self-contained and accessible, the system enters a report generation phase.
The report generation module consumes the full explorer trace and its citations to produce an end-to-end report. The generated report is passed to the evaluation module, which scores the report on metrics encompassing quality and reliability. These scores are shown to the user alongside the report.

Finally, the system periodically generates \emph{meta-reports} that aggregate selected insights into a higher-level synthesis. Because insights span diverse subtopics, meta-reports are organized around themes, trends, or emerging questions that cut across the exploration map. Insights are selected automatically based on common themes and content.

\subsection{Exploration Map}

Given an overall goal, and potentially a corpus or search tools to operate on, \nomad{} must first capture the set of potential topics that may be interesting to explore. 
For example, given the same tool, e.g., a database comprising enterprise data, the topics of interest may be different for an HR executive vs a Marketing Analyst. 

Therefore, before commencing multiple deep research threads on a given use case, \nomad{} starts by constructing an \emph{Exploration Map} of interesting topics. We denote this map as a graph $G = (V, E)$. A subgraph of $G$ is the Topic Tree, where the root node represents the overall goal, and the first level of nodes are the main subtopics. Each subsequent level contains finer subtopics. 

The exploration map contains the following types of nodes, and edges between them:
\begin{itemize}
\item \textbf{Root}: The root node represents the overall goal or domain of interest. There is only one root node in the graph.
\item \textbf{Topic nodes}: These nodes represent topics and subtopics in the domain. They are organized hierarchically under the root node. A topic node may have multiple topic nodes or concept nodes as children. Topic nodes that do not have any concept or topic children may have hypotheses  children. The parent of a topic node may be the root node or another topic node. The root node and the topic nodes together form the Topic Tree.
\item \textbf{Concept nodes}: These nodes represent key concepts, entities, or keywords extracted from the documents. If there are no documents to analyze, there are no concept nodes in the graph. Concept nodes can have only topic nodes as parents, and they may have at most one hypothesis node as a child.
\item \textbf{Hypothesis nodes}: These nodes represent hypotheses generated for a concept or topic node, and are children of concept or topic nodes. They may have at most 1 child - which is an insight node. A Hypothesis node is \emph{explored} if it has an insight child, and is \emph{unexplored} otherwise. 
\item \textbf{Insight nodes}: These nodes represent validated insights derived from hypotheses. They are children of hypothesis nodes and have no children.
\item \textbf{Document nodes}: These nodes represent documents in the corpus. They are connected to the concept nodes that were extracted from them.
\end{itemize}

\subsubsection{Exploration Map Variations}

The construction of the exploration map $G$ follows two separate paths depending on the availability of a corpus. The case where a corpus is not available follows the typical setting in which users today use deep-research tools \citep{openai2025deepresearch,google2024geminideepresearch,perplexity2026researchmode,gptresearcher2025deepresearch}: they do have a (potentially specific) research goal, but they typically rely on external information only.

If a corpus is not available, the map is constructed \emph{top-down}: the user's research goal, augmented by any relevant Web information, is used to directly construct the Topic Tree using an LLM - a one-shot hierarchical breakdown of the goal.

If a corpus is available, the map is built \emph{bottom-up}: The system first extracts concepts (important topics, key phrases) from the documents, and then organizes these concepts into a hierarchy. 

Many real settings also require a temporal layer on top of either case. A corpus or a written goal may capture the core domain, but recent events from the Web can still be critical.
For example, a company may have internal research on marketing, yet a new trade deal, sanction, or conflict can change what matters.

\nomad{} supports this temporal update by periodically augmenting the exploration map with information from the Web. The map is still built top-down or bottom-up depending on corpus availability, but augmentation can add new topics and sub-topics over time. The probability of Web expansion is a hyperparameter that can be tuned per use case to balance stability and freshness.

In the following, we discuss the construction of the exploration map $G$ in each setting.

\subsubsection{Top-Down Exploration Map Construction}
In cases where a corpus is not provided, the overall goal itself is used to generate meaningful and diverse keywords for Web search. Together with the provided goal and any other information about the user or domain available, the retrieved documents from Web search are used to prompt an LLM to construct a hierarchical Topic Tree. We prompt the model to maintain diversity and coverage, have at least 3 levels of hierarchy, and to ensure that the topics at each level are semantically coherent. The resulting Topic Tree is the exploration map in this case, and the system proceeds to the topic selection phase. 

\subsubsection{Bottom-Up Exploration Map Construction}

In cases where a corpus is available, the exploration map follows broadly two high-level steps: First, it constructs a \emph{Concept Layer}, the set of all potentially interesting concepts (entities, keywords) discovered in the corpus, followed by the \emph{Topic Tree}, a hierarchical organization of these concepts.

\paragraph{Step 1: Concept Layer Construction}
Given a set of documents $\mathcal{D}$, the exploration map construction module operates in bottom-up fashion. First, for each document \footnote{In a practical implementation, documents need to be chunked. For the rest of this paper, we use \emph{document} to refer to individual chunks. The metadata associated with each document stores the source information - i.e. which filename and page number, or URL, it comes from.} it extracts key concepts - topics, entities, important dates, or other key phrases using an LLM-based extraction method. For each such extracted concept, the LLM is prompted to return the concept name, the phrases mentioning the concept, and a description of it based on the document. 

Extracted concepts are disambiguated across documents (for example, UAE $\rightarrow$ United Arab Emirates) to point to a single unified form. Resolving concept names starts by a surface level step - in this step, we only resolve concepts that have a clear surface form match -- addressing case variations and abbreviations (in the case of person names). In the next step, we use an LLM-based disambiguation prompt that takes in two concept mentions and their descriptions, and outputs whether they refer to the same concept or not. For each concept in the extracted concept set, we retrieve the top-k similar concepts based on text embeddings of the concept name and its description, and run the disambiguation prompt with an LLM to disambiguate these pairs. This results in a unified set of concepts $\mathcal{C}$. The concept layer of the graph is then constructed by adding these concepts as nodes, and connecting them to the documents they were extracted from. For each concept in $\mathcal{C}$, we construct a unified description based on the various descriptions obtained from the documents that mention it.
Algorithm \ref{alg:concept-extraction} describes the process of concept layer construction.

\begin{figure}[t]
    \centering
    \includegraphics[width=\linewidth]{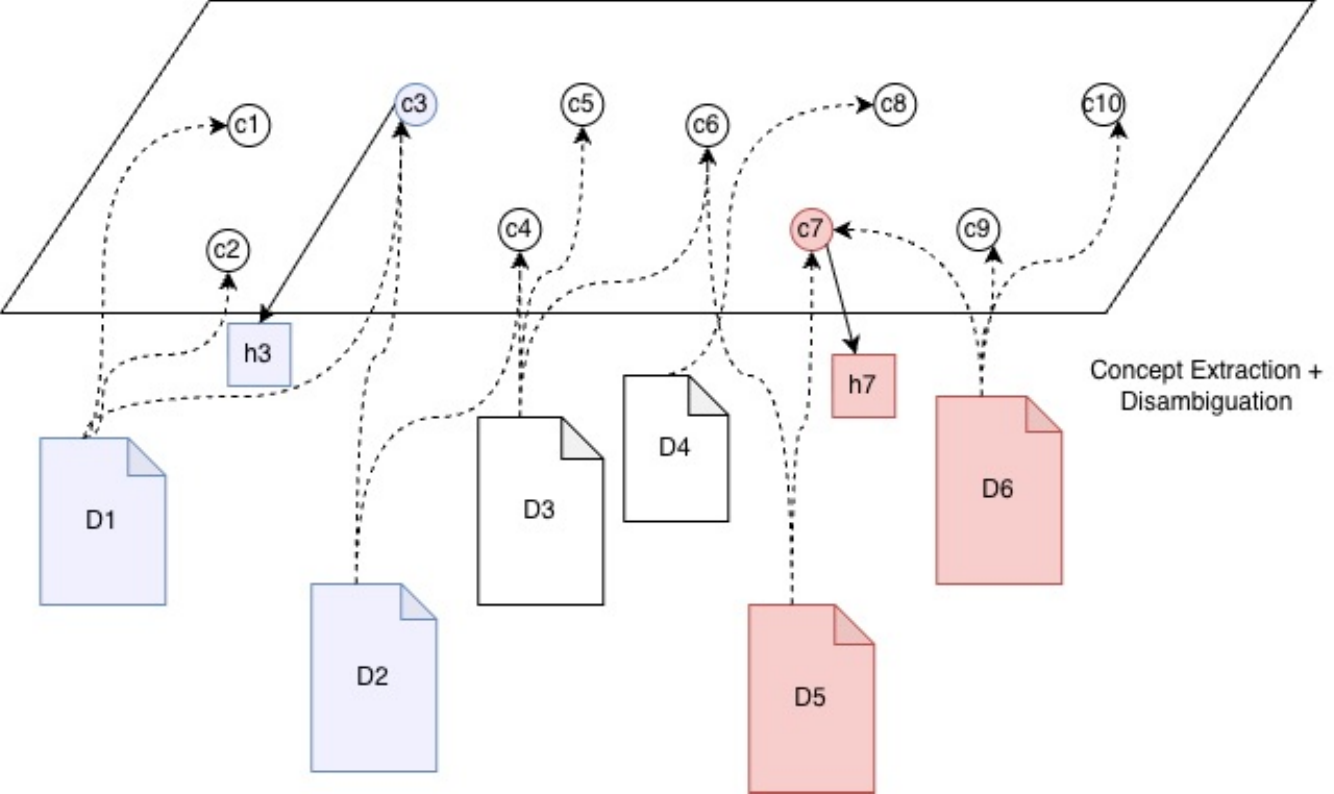}
    \caption{Interim state of the Exploration Map after concept layer construction. Documents are connected to the concepts extracted from them. Concepts are disambiguated across documents to point to a unified concept. The red (blue) highlighted nodes become part of a single LLM call during insight potential evaluation of concept $c7$ ($c3$). This results in the generation of hypothesis $h7$ ($h3$) as a child of the concept node.}
    \label{fig:concept-layer}
\end{figure}

\begin{algorithm}[t]
\caption{Concept Layer Construction}
\label{alg:concept-extraction}
\begin{algorithmic}[1]
\Require Document set $\mathcal{D}$
\Ensure Graph $G = (V, E)$
\State Initialize $V \gets \mathcal{D}$, $E \gets \emptyset, \mathcal{C} \gets \emptyset$
\ForAll{$d \in \mathcal{D}$}
    \State Extract concepts $\mathcal{C}(d)$ from $d$
    \State Extract structured metadata $m(d)$ and attach to $d$
    \State $\mathcal{C} \gets \mathcal{C} \cup \mathcal{C}(d)$
\EndFor
\State Disambiguate concepts in $\mathcal{C}$ to get unified concept set $\mathcal{C}$
\State Construct unified description $Desc(c)$ for each concept $c \in \mathcal{C}$ based on individual descriptions obtained from documents mentioning $c$
\State $V \gets V \cup \mathcal{C}$
\ForAll{$d \in \mathcal{D}$}
    \ForAll{$c \in \mathcal{C}(d)$}
        \State $E \gets E \cup \{(d, c, \textsc{doc-concept})\}$
    \EndFor
\EndFor
\State Identify candidate concepts $\mathcal{C}_\text{cand}$ with near-median document frequency
\ForAll{$c \in \mathcal{C}_\text{cand}$}
    \State Score insight potential and generate hypothesis $g(c)$ using $Desc(c)$ and documents mentioning $c$ (with metadata)
    \If{score is positive}
        \State Construct hypothesis node $h$
        \State $h.\text{hypothesis} \gets g(c)$
        \State $V \gets V \cup \{h\}$
        \State $E \gets E \cup \{(c, h, \textsc{parent-child})\}$
        \State Set potential score of $c$ to $1$
    \EndIf
\EndFor
\State \Return $G = (V, E)$
\end{algorithmic}
\end{algorithm}

Next, as part of the concept layer construction, for each document $d \in \mathcal{D}$, we also extract metadata related to the document. Here, given the existing metadata for the document (source filename, URL etc), the LLM is prompted to derive as much additional metadata in a structured format as possible. This includes information such as the type of document, the date it was created, the author, the organization it is related to, and any other relevant metadata. This metadata is added to the graph as attributes of the relevant document nodes. For system prompts used in concept layer construction: extraction, disambiguation, and metadata extraction, please refer to Appendix \ref{app:exploration-map-system-prompts}, under the Concept Layer System Prompts subsection.

\paragraph{Identifying Interesting Concepts}

Note that for even a few hundred documents, the number of concepts extracted in this manner can easily be in the order of thousands. Our eventual objective is to guide the system towards interesting and novel insights \emph{first}, while also ensuring diversity in exploration.

To achieve this, we qualify concepts based on their \emph{interestingness} or insight potential. There exist concepts with a large fraction of documents mentioning them (for example, the name of the organization in a corpus of enterprise documents), and there also exist a large number of concepts which are mentioned in only one document. Both types are unlikely to lead to novel or interesting information. The former are almost `stop-words' in this domain, and the latter are often specific dates mentioned in passing or similar. We therefore focus on concepts that have a median or close to the median number of documents referencing them. We refer to the set of these concepts as `candidate concepts'. For each candidate concept, we provide a reasoning model with the concept name and description, along with the contents of the documents that mention them, including document metadata. The reasoning model is prompted to generate a score for the potential of this concept to lead to interesting insights based on the documents. It also needs to state the reasoning for arriving at that score, the key connection drawn between the documents, and the resulting starting hypothesis that a relevant research may begin with (for the prompt used, see Appendix \ref{app:exploration-map-system-prompts}, under the Concept Layer System Prompts subsection).

The above step is key to \nomad{}. Having been provided \textbf{only} the set of documents that refer to this concept at a time, as well as the connecting theme to focus on (the concept itself), the reasoning model is expected to identify clear contradictions, surprising connections, or other interesting patterns better than if all documents were in its context at once. While today's frontier models have very large contexts and could potentially fit large fractions of the corpus in their context at a time, this deliberate focus on a small subset at a time allows the surfacing of a much larger and diverse set of potentially interesting research topics.

All candidate concept nodes that get a positive insight potential score get assigned a potential score $1$, and the generated hypothesis by the reasoning model becomes a child node of it, as an unexplored hypothesis, and thus part of the hypothesis pool. Other concept nodes (non-candidate concept nodes) automatically have a score of $0$ and no hypothesis children.

Figure \ref{fig:concept-layer} illustrates the concept layer after the construction completes -- this captures documents, concepts extracted from them, and any generated hypotheses.

Tables \ref{tab:concept-layer-hypothesis-1}, \ref{tab:concept-layer-hypothesis-2}, \ref{tab:concept-layer-hypothesis-3} show examples of hypotheses generated for the concepts `COVID-19', `ILO', and also a temporal concept like `2018' in the WHO Research Analyst instance. The first hypothesis identifies feedback loops linking COVID-19 to diabetes morbidity, antibiotic use, antimicrobial resistance, and socioeconomic conditions. The reasoning draws on four documents that mention COVID-19 in different contexts, connecting them into novel and actionable research threads. The second hypothesis explores the possibility of training more healthcare workers to offset the impact of automation led by generative AI on the labor economy. The concept `2018', on the other hand, links three thematically diverse documents on a common temporal anchor point with unmet policy commitments and worsening trends. 
These examples illustrate how the concept layer construction can surface non-obvious insights by focusing on a single concept and its associated documents at a time.

\begin{table}[t]
\centering
\caption{Concept-layer hypothesis example: COVID-19. The hypothesis and reasoning are generated from the documents associated with the concept, and the decisive snippets are highlighted in \textcolor{myblue}{blue}.}
\label{tab:concept-layer-hypothesis-1}
\begingroup
\footnotesize
\setlength{\tabcolsep}{2pt}
\renewcommand{\arraystretch}{1.12}
\begin{tabularx}{\linewidth}{@{}X@{}}
\toprule
\textbf{Concept Name} : COVID-19 \\
\midrule
\textbf{Generated Hypothesis}: The COVID-19 pandemic set off a reinforcing feedback loop in low- and middle-income countries wherein (a) economic shocks and falling labour-income shares increased poverty-related risk factors for diabetes and reduced access to routine care; (b) the resulting rise in severe infections among diabetics prompted higher, often inappropriate, antibiotic use; and (c) this surge in antibiotic consumption accelerated antimicrobial resistance, which in turn raises treatment costs and further strains household and health-system finances.  Breaking this loop requires integrated policies that simultaneously address NCD management, antibiotic stewardship, and social protection. \\
\midrule
\textbf{Reasoning}: All four documents speak of COVID-19 as a shock that spills over into very different domains: 1. the clinical vulnerability of people with diabetes and the disruption of their care, 2. the obligation of Member States to keep diabetes services running despite the pandemic, 3. the rise in antimicrobial resistance (AMR) partly fuelled by inappropriate antibiotic use during COVID-19, and 4. the pandemic-triggered fall in the global labour-income share that deepens inequality.  These statements are not merely parallel references; they outline a potential causal chain that links a health-system stressor (COVID-19) to (i) higher diabetes morbidity/mortality, (ii) increased antibiotic use and AMR, and (iii) worsening socioeconomic conditions.  The possibility that economic hardship and reduced labour income further exacerbate diabetes prevalence (through poverty-related risk factors) and drive inappropriate antibiotic use (through weakly regulated informal markets) suggests a feedback loop.  Investigating this loop could uncover leverage points where integrated economic, stewardship, and NCD policies break the cycle – an area that is currently under-explored and highly actionable, especially for LMICs.  Hence the seed set can credibly launch a multi-sectoral research thread with game-changing policy implications. \\
\midrule
\textbf{Supporting Documents} \\
\textbf{D1.} \emph{WHO Health Topics Diabetes A74 R4, p.~2.} 
\ldots increased across both periods; [2] \textcolor{myblue}{Concerned that people living with noncommunicable diseases, including diabetes, have a higher risk of becoming severely ill or dying from coronavirus disease (COVID-19), and are among those most impacted by the COVID-19 pandemic; [3] Concerned also that complete or partial disruptions to diabetes prevention and control due to the COVID-19 pandemic, including in respect of early detection and diabetic complication management services, represent significant threats to the life and health of people living with diabetes;} 1 See document A74/10 Rev.1. \ldots\\
\textbf{D2.} \emph{WHO Health Topics Diabetes A74 R4, p.~4.} \ldots for developing type 2 diabetes; \textcolor{myblue}{(5) to ensure a continued focus on maintaining a high level of treatment and care for all people, regardless of the COVID-19 pandemic, including for people living with diabetes, especially in low- and middle-income countries, recognizing that necessary diabetes prevention and control efforts are hampered by, inter alia, lack of universal access to quality, safe, effective, affordable essential health services, medicines, diagnostics and health technologies, as well as by a global shortage of qualified health workers;} (6) to ensure that national \ldots \\
\textbf{D3.} \emph{WHO Health Topics Antimicrobial Resistance A77-5, p.~3.} \ldots focus on drug-resistant bacterial infections. \textcolor{myblue}{However, other infections, such as viral and fungal infections, may elicit the inappropriate use of antibiotics and are also considered. The pandemic of coronavirus disease (COVID-19), for example, contributed to antimicrobial resistance.} As further evidence emerges on \ldots \\
\textbf{D4.} \emph{ILO World Employment Update, p.~2.} \ldots stood at 52.9 per cent. \textcolor{myblue}{After a short-lived increase in 2020, the share in 2021 had already returned to pre pandemic levels as the global economy recovered from the worst impacts of the COVID-19 pandemic. In 2022, it declined to 52.3 per cent.} Incorporating the latest macroeconomic data, \ldots\\
\bottomrule
\end{tabularx}

\endgroup
\end{table}

\begin{table}[h]
\centering
\caption{Concept-layer hypothesis example: ILO}
\label{tab:concept-layer-hypothesis-2}
\begingroup
\footnotesize
\setlength{\tabcolsep}{2pt}
\renewcommand{\arraystretch}{1.12}
\begin{tabularx}{\linewidth}{@{}X@{}}
\toprule
\textbf{Concept Name} : ILO \\
\midrule
\textbf{Generated Hypothesis}: Strategic public investment in expanding the health and care workforce—as advocated by the WHO-ILO-OECD `Working for Health' action plan—can offset automation-induced declines in the labour-income share, reduce youth NEET rates (especially for women), and mitigate rising inequality over the next decade. Empirical modelling that integrates sectoral labour-intensity profiles, automation susceptibility indices, and macro labour-income-share trends can quantify the potential distributional and growth effects of such health-employment expansion. \\
\midrule
\textbf{Reasoning}: The documents build a coherent narrative: (a) the global labour-income share has been falling; (b) automation-oriented technological change is a statistically significant contributor; (c) this erosion risks worsening inequality and youth NEET rates, especially for young women.  A separate WHO resolution jointly adopted with ILO and OECD promotes large-scale investment in the health workforce - particularly community health workers - as a lever for `health employment and inclusive economic growth'.  The ILO report itself notes that some forms of technological innovation can *raise* the labour-income share when they `create labour-intensive tasks'.  Health and care services are well-known to be comparatively resistant to automation and are highly labour-intensive, offering significant employment opportunities for youth and women. \\
Hence, there is a non-trivial, cross-document connection: while automation is squeezing labour's share in many sectors, coordinated global expansion of the health-care workforce (championed by the joint WHO-ILO-OECD plan) could offset that squeeze by generating labour-intensive, less-automatable jobs, simultaneously addressing NEET rates and gender gaps.  This link is neither explicit in the ILO update nor in the WHO resolution, yet combining them suggests an actionable research avenue: can strategic investment in health employment act as a macro-level counterbalance to automation-induced inequality and youth unemployment?\\
Because the documents provide complementary empirical findings, theoretical framing (Acemoglu and Restrepo), and a concrete policy programme (Working for Health plan), they form a strong seed for high-impact, multidisciplinary research that could model and quantify such a compensatory strategy, explore financing mechanisms, and guide policy design. \\
\midrule
\textbf{Supporting Documents} \\
\textbf{D2.} \emph{ILO World Employment Update Sep 2024, p.~3.} \ldots Outlook: September 2024 Update 3 \textcolor{myblue}{While the decrease appears modest in percentage points (a reduction of 1.6 p.p. in two decades, of which 0.6 p.p. in the last 5 years), the effect is sizeable. This decline represents \$2.4 trillion (in constant PPP) in labour income globally in 2024.} Notwithstanding the distributional significance of \ldots \\
\textbf{D3.} \emph{ILO World Employment Update Sep 2024, p.~4.} \textcolor{myblue}{In that framework, different types of technological innovations have opposite effects on the labour income share. Certain innovations are predicted to lower it (such as automation) whereas others will tend to increase it (such as the creation of labour-intensive tasks), some innovations (capital or labour augmenting innovations) have an ex-ante ambiguous effect.} \ldots \\
\textbf{D5.} \emph{ILO World Employment Update Sep 2024, p.~8.} \ldots  access to education and employment. \textcolor{myblue}{Globally in 2024, the female youth NEET incidence is estimated at 28.2 per cent, more than double the incidence among young men, which stands at 13.1 per cent.} Two decades ago, the gap \ldots \\
\textbf{D6.} \emph{ILO World Employment Update Sep 2024, p.~9.} \ldots Outlook: September 2024 Update 9 \textcolor{myblue}{across a wide range of relevant domains, the recent breakthroughs in generative AI could exert further downward pressure on the labour income share.} Nonetheless, the results presented should \\
\textbf{D7.} \emph{Universal Health Coverage A72 R3, p.~5.} \ldots human resources for health and resolution WHA70.6 (2017) in \textcolor{myblue}{which the Health Assembly adopted “Working for Health”: the ILO, OECD, WHO five-year action plan for health employment and inclusive economic growth (2017-2021), and future work on community health worker programmes;} (7) to submit a report \ldots\\
\bottomrule
\end{tabularx}
\endgroup
\end{table}

\begin{table}[h]
\centering
\caption{Concept-layer hypothesis example: 2018}
\label{tab:concept-layer-hypothesis-3}
\begingroup
\footnotesize
\setlength{\tabcolsep}{2pt}
\renewcommand{\arraystretch}{1.12}
\begin{tabularx}{\linewidth}{@{}X@{}}
\toprule
\textbf{Concept Name} : 2018 \\
\midrule
\textbf{Generated Hypothesis}: Insufficient progress on the 2018 UNGA diabetes commitments—particularly inadequate insulin accessibility in low- and middle-income countries—has indirectly contributed to the continued global rise of antimicrobial resistance observed in 2018-2023 by increasing infection incidence and antibiotic consumption among people with poorly controlled diabetes. Integrated strategies that simultaneously expand diabetes care (insulin supply, lifestyle intervention) and AMR stewardship could yield greater impact than siloed approaches. \\
\midrule
\textbf{Reasoning}: The three documents are thematically different (diabetes, antimicrobial resistance policy, and AMR surveillance) yet all anchor key data or policy commitments to the year 2018. [Document 1] cites the 2018 UNGA resolution 73/2 that called for intensified primary-care diabetes action. [Document 2] cites the 2018 UNEP resolution on AMR and urges stronger One-Health AMR plans. [Document 3] presents hard surveillance data showing that, despite policy efforts, median annual AMR for several critical pathogens has risen steadily from the 2018 baseline through 2023. Because diabetes substantially increases infection risk and antibiotic consumption (e.g., diabetic foot infections, urinary-tract infections), inadequate diabetes control—exacerbated by poor insulin access highlighted in [Document 1]—can indirectly fuel inappropriate or prolonged antibiotic use, thereby contributing to AMR. Meanwhile, [Document 3] reports that AMR is highest where surveillance (and, by implication, health-system capacity) is weakest—often the same low- and middle-income countries that [Document 1] notes have poor insulin access. Thus the documents can be linked to form a non-trivial research thread: failures to meet the 2018 diabetes-care commitments may be an unrecognised driver undermining the 2015-2022 AMR action plans, helping to explain why AMR trends since 2018 are worsening despite policy pledges. Exploring this intersection could reveal synergistic interventions (e.g., increasing affordable insulin and glycaemic control to curb infection incidence, thereby reducing antibiotic pressure and slowing AMR).\\
\midrule
\textbf{Supporting Documents} \\
\textbf{D1.} \emph{Diabetes A74 R4 - p.~3.} \ldots segments of the population; [1] \textcolor{myblue}{Reaffirming also our commitment in United Nations General Assembly resolution 73/2 (2018) to further strengthen efforts to address diabetes as part of universal health coverage through intensified interventions at the primary health care level, including in low- and middle-income countries, on prevention and control of diabetes;} Emphasizing the importance of prevention \ldots \\
\textbf{D2.} \emph{Antimicrobial Resistance A72 R5, p.~2.} \ldots and promote their prudent use; \textcolor{myblue}{Recalling FAO resolution 4/2015 on antimicrobial resistance, World Organisation for Animal Health (OIE) resolution No. 36 (2016) on combating antimicrobial resistance through a One Health approach: actions and OIE strategy, and the UNEP resolution UNEP/EA.3/Res.4 (2018) on environment and health; Noting the importance of providing opportunities for Member States to engage meaningfully with and provide input into reports, recommendations, and relevant actions from WHO, FAO, and OIE, together with UNEP, and from the Interagency Coordination Group on Antimicrobial Resistance aimed at combating antimicrobial resistance; Reaffirming the global commitment to combat antimicrobial resistance with continued, high-level political efforts as a coordinated international community, emphasizing the critical need to accelerate Member States' development and implementation of their national action plans with a One Health approach}, 1. WELCOMES the new tripartite \ldots \\
\textbf{D3.} \emph{Global Health GLASS AMR Surveillance Report 2025, p.~11.} \ldots  report higher levels of AMR. \textcolor{myblue}{The frequency of AMR is highest in countries with low surveillance coverage. In fact, there is a strong inverse correlation between a country's AMR surveillance coverage and its reported median AMR (Pearson correlation coefficient, $r = -0.74, P < 0.0001$, Fig. 4).} This pattern may reflect both \ldots \\
\bottomrule
\end{tabularx}
\endgroup
\end{table}
Once the concept layer is constructed with concepts and their associated hypotheses, the next step is to build the Topic Tree on top of this concept layer. 

\paragraph{Step 2: Topic Tree Construction}

A simple approach to building the Topic Tree is to prompt an LLM to construct the complete hierarchy of topics in one shot given all the concept names, while instructing it to cover all concepts and maintain the semantic hierarchy as well. However, this approach usually doesn't work well, because given a large number of topics (typically in the order of at least thousands for even 100-200 documents) LLMs, even frontier ones, struggle to include all provided concepts in the hierarchy, and often miss a large fraction of them. Another potential approach is to first cluster the concepts based on their text embeddings, and then prompt the LLM to name each cluster based on its members while constructing the Topic Tree. 
However, since text embeddings often cannot capture the fine-grained semantics of the concepts, this approach results in clusters that are not semantically coherent, and thus the LLM cannot name them in a meaningful way, and also cannot construct a meaningful Topic Tree based on these clusters.

\begin{figure}[t]
    \centering
    \includegraphics[width=0.9\linewidth]{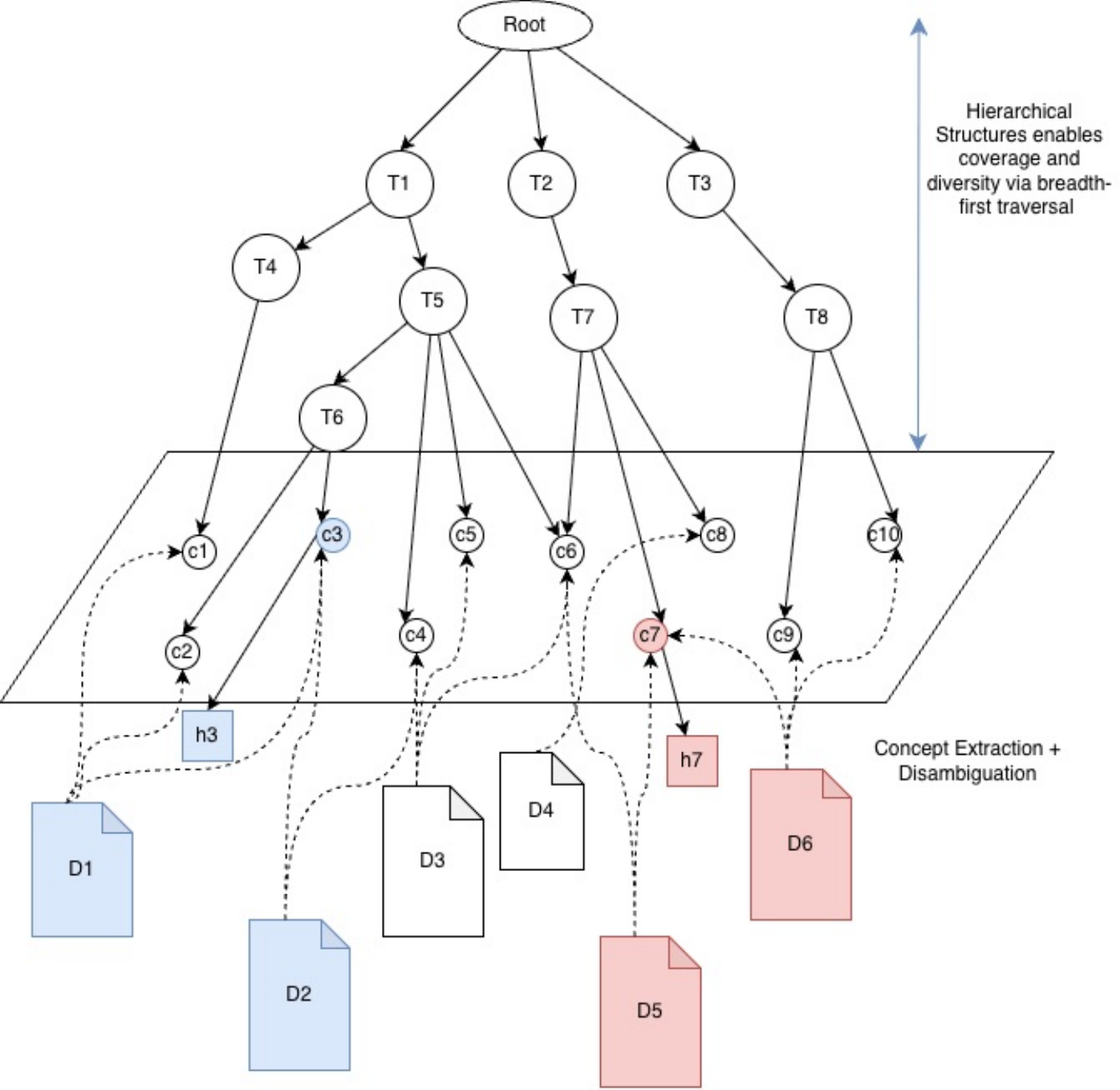}
    \caption{The full exploration map: A Topic Tree is constructed on top of the concept layer, and concepts are connected to their closest leaf topic node. The Topic Tree is traversed breadth-first for exploration.}
    \label{fig:exploration-map}
\end{figure}

We therefore follow a third, hybrid approach. We generate text embeddings for each concept based on its name and description, and we cluster these embeddings to get a set of concept clusters\footnote{We follow a standard clustering algorithm - we use \texttt{UMAP} to reduce dimensionality with \texttt{n\_components=20} and \texttt{n\_neighbors=3}, followed by \texttt{HDBSCAN} with a minimum cluster size of 5. These choices were part of a small scale experiment conducted over a selected dataset.}. However, instead of treating these potentially erroneous clusters as final, we use them as a suggestion for the LLM to construct the Topic Tree. We prompt the LLM to generate the Topic Tree without the concept memberships (higher level topics only), and we present all concepts at once, simply \emph{ordering them} based on their cluster membership, thus providing the LLM with a strong signal of which concepts may be related to each other, while still allowing it the freedom to deviate from these clusters if it deems fit (see the Topic Tree System Prompts subsection in Appendix \ref{app:exploration-map-system-prompts}). Moreover, since no explicit mention of any clusters is given, the LLM is not required to name clusters. Once a Topic Tree has been generated, each concept is connected to its closest leaf topic node based on text embedding similarity between itself and that of the topic node (based on its topic). This results in a Topic Tree that is more accurate and less prone to errors than the previous two approaches.
Figure \ref{fig:exploration-map} shows the full exploration map after this step. Algorithm \ref{alg:topic-tree} summarizes this hybrid topic-tree construction.

\begin{algorithm}[t]
\caption{Hybrid Topic Tree Construction}
\label{alg:topic-tree}
\begin{algorithmic}[1]
\Require Concept set $\mathcal{C}$ with names and descriptions
\Ensure Topic Tree $T$ with concept assignments
\State Compute embeddings $\mathbf{e}_c$ for each $c \in \mathcal{C}$ from name and description
\State Cluster $\{\mathbf{e}_c\}$ into clusters $\{\mathcal{K}_1,\ldots,\mathcal{K}_m\}$
\State Order concepts by cluster membership to form list $\mathcal{L}$
\State Prompt LLM with $\mathcal{L}$ to generate Topic Tree $T$ with leaf nodes $L(T)$.
\ForAll{$c \in \mathcal{C}$}
    \State Assign $c$ as a child Concept node to $\ell^* = \arg\max_{\ell \in L(T)} \cos\_\mbox{sim}(\mathbf{e}_c, \mathbf{e}_\ell)$
\EndFor
\State \Return $T$
\end{algorithmic}
\end{algorithm}

\begin{figure}[t]
    \centering
    \includegraphics[width=\linewidth]{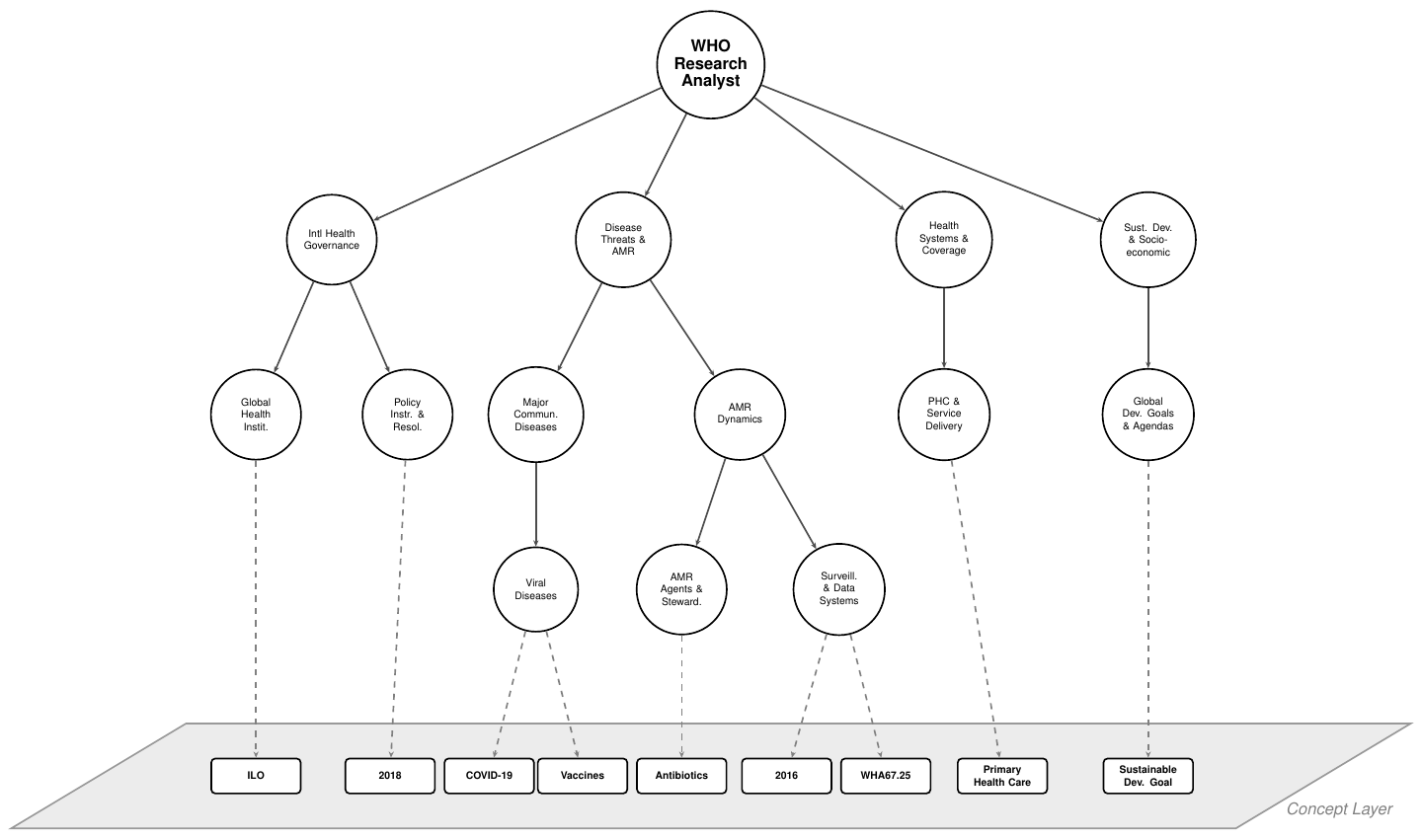}
    \caption{A real exploration map from the WHO Research Analyst instance. The Topic Tree (circle nodes) is constructed over the concept layer (rectangles), with each concept connected to its closest leaf topic node via dashed lines. Note the non-uniform branching factor and depth across different branches of the tree.}
    \label{fig:who-exploration-map-example}
\end{figure}

\begin{figure}[t]
    \centering
    \includegraphics[width=\linewidth]{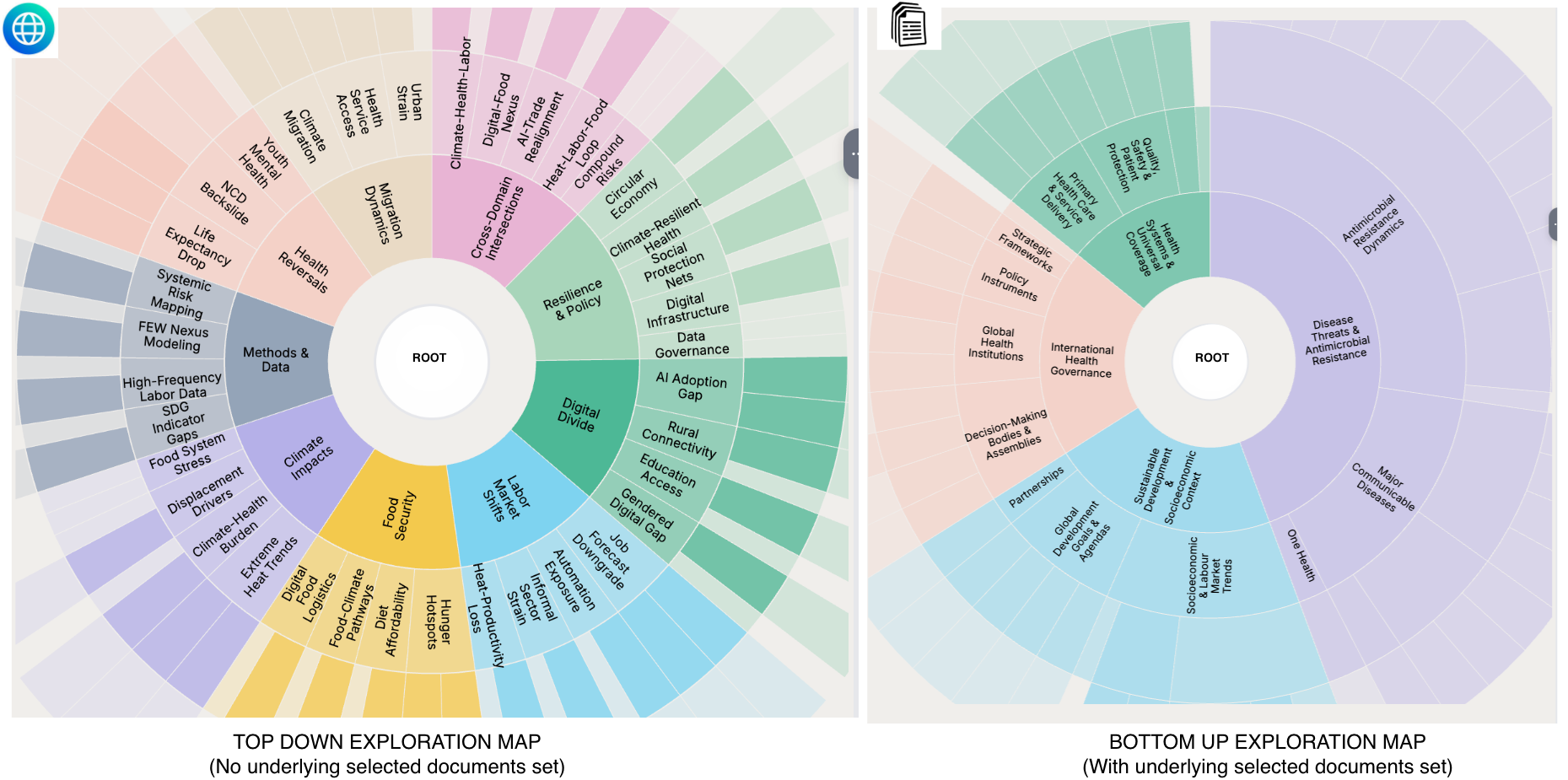}
    \caption{Running example of exploration map construction for the WHO Research Analyst use case. The figure shows both top-down construction from the research goal alone (left) and bottom-up construction from a provided corpus (right). While the top-down map is more general and covers more diverse topics (using the LLMs world knowledge about topics relevant to the goal), the bottom-up map is more detailed and specific to the topics found in the small set of documents provided - demonstrating the relevance and relative advantages of both approaches. Moreover, the bottom-up map has non-uniform branching factor and depth, which is a direct consequence of the insight potential evaluation step during concept layer construction, which awards individual concepts with suitable insight potential with hypotheses before the Topic Tree construction, thus potentially resulting in topic nodes with more (or fewer) such concept nodes in their subtree.}
    \label{fig:who-exploration-maps}
\end{figure}

Figure \ref{fig:who-exploration-map-example} shows a running example from the WHO Research Analyst use case for the bottom up exploration map. Figure \ref{fig:who-exploration-maps} shows a different view of the Topic Tree (as a sunburst diagram) and illustrates the differences between the constructed maps in both settings: bottom-up when a corpus is available, and top-down when no corpus is provided.

\textbf{Incremental Document Addition}: While we do not focus on this in the current version of \nomad{}, we note that the bottom-up exploration map can be updated incrementally as new documents are added. When a new document is added, we can extract concepts from it, disambiguate them with existing concepts, and then connect them to the Topic Tree based on their similarity to existing concepts and topics. An additional LLM-based step may also insert a new branch into the Topic Tree if required. This allows \nomad{} to continuously evolve its understanding of the domain as new information becomes available.

\subsection{Web Search-Based Exploration Map Expansion}

In both cases, top-down and bottom-up, the exploration map is built once, and remains \emph{static} for the entire duration of the research. This has the drawback of not being able to capture new information that may become available during the research process, which may be crucial to integrate. This is especially relevant for \nomad{} instances that run over long periods of time, such as weeks or months, and for domains that are rapidly evolving, such as geopolitics or current events.

We therefore introduce a mechanism for exploration map expansion based on Web search. Recent agentic search work suggests that deciding when and how to branch search is itself a major part of long-horizon performance \citep{jin2025searchr1}. With some probability at each step, \nomad{} can expand the exploration map using Web search (with keywords generated using the provided goal), and add the newly discovered concepts and topics to the existing exploration map. Given a set of seeds derived from Web search results and a starting node in the Topic Tree, the system traverses the tree level-wise. At each level, it collects the names of the current node's children (both topic and concept nodes) and prompts an LLM with these candidates, the seed set, and the super-topic path (see the Exploration Map Expansion System Prompt subsection in Appendix \ref{app:exploration-map-system-prompts}). The LLM selects the most relevant existing child when possible; otherwise it proposes a new topic name that is not already present. If the selection matches an existing child, the system descends into that child. If not, it creates a new topic child and descends into it. This process repeats until the maximum allowed depth is reached, ensuring new topics are inserted at an appropriate level rather than always being attached to the root.

\begin{algorithm}[t]
\caption{Seed-Guided Topic Insertion.}
\label{alg:seed-guided-insertion}
\begin{algorithmic}[1]
\Require Topic Tree $T$, start node $v_0$, seed set $S$, max levels $L_\text{topic}, L_\text{concept}$
\Ensure Most relevant node $v$
\State $v \gets v_0$
\State $L \gets \max(L_\text{topic}, L_\text{concept})$
\While{$\text{level}(v) < L$}
    \State $C \gets$ names of children of $v$ (topic and concept nodes)
    \State $P \gets$ super-topic path from root to $v$
    \State $t \gets$ LLMSelect$(C, P, S)$ \Comment{Prefer existing topics; propose new if none fit}
    \If{$t$ matches an existing child of $v$}
        \State $v \gets$ that child
    \Else
        \State Create new topic node $v'$ with name $t$ at level $\text{level}(v)+1$
        \State Add $v'$ as a child of $v$
        \State $v \gets v'$
    \EndIf
\EndWhile
\State \Return $v$
\end{algorithmic}
\end{algorithm}

\begin{figure}[t]
    \centering
    \includegraphics[width=0.8\linewidth]{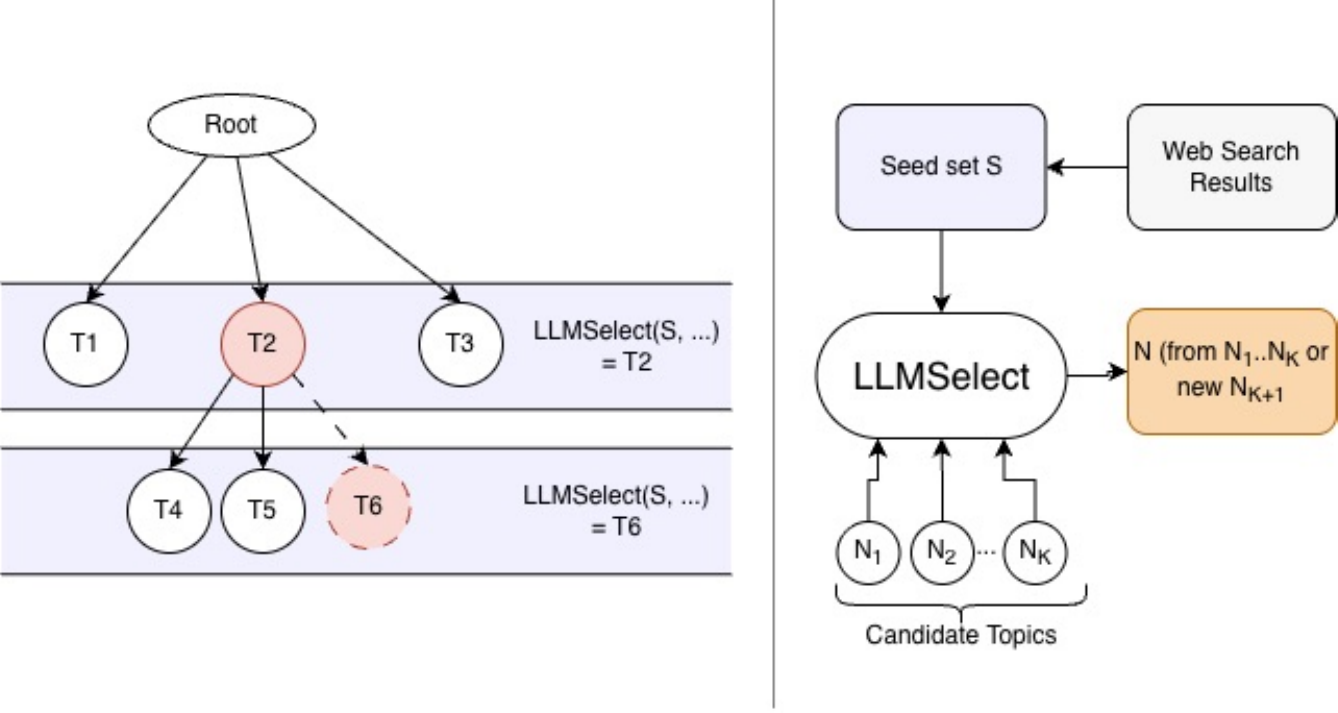}
    \caption{Seed-guided topic insertion during Web search expansion of the exploration map. At the first level, $T2$ is selected based on relevance to the seed set. At the second level, since no existing child matches the seed, a new topic $T6$ is proposed and inserted as a child of $T2$.}
    \label{fig:seed-guided-expansion}
\end{figure}

\subsection{Topic Selection}

Once \nomad{} constructs the exploration map, it needs to select which topic to explore first. 

Note that depending on whether the exploration map was constructed top-down or bottom-up, or the last time it was augmented with Web search, the Topic Tree may or may not have unexplored hypotheses at the leaf level. The topic selection module works the same way in either case.

First, we define for each topic node, two types of scores: the \emph{insight potential score} and the \emph{exploration score}. The insight potential score of a topic node is the total number of unexplored hypotheses in its subtree. The exploration score of a topic node is the number of explored hypotheses in its subtree.

The topic selection module starts from the root. At each node, it first collects the set of child topic nodes with a non-zero insight potential score. If such nodes exist, it selects the one with the lowest exploration score, thus prioritizing less explored topics. If no such node exists, it selects the child topic node with the lowest exploration score among all child nodes. Thus, it balances utilizing past work (of generating hypotheses) and diverse exploration. This process continues until a leaf topic node or concept node is reached. If the reached node has unexplored hypotheses, the system selects one of them for exploration. If not, it triggers the hypothesis generation module to generate new hypotheses for this topic or concept, which are then added to the exploration map as children of this node, and one of them is selected for exploration. 

Algorithm \ref{alg:topic-selection} summarizes the topic selection procedure.

\begin{algorithm}[t]
\caption{Topic Selection}
\label{alg:topic-selection}
\begin{algorithmic}[1]
\Require Topic Tree $T$ with insight potential and exploration scores; hypothesis pools at leaf topics and concept nodes
\Ensure Selected hypothesis $h$
\State $v \gets \textsc{Root}(T)$
\While{$v$ is a topic node and has topic children}
    \State $C \gets$ child topic nodes of $v$
    \State $C_+ \gets \{c \in C \mid \textsc{InsightPotential}(c) > 0\}$
    \If{$C_+ \neq \emptyset$}
        \State $v \gets \arg\min_{c \in C_+} \textsc{ExplorationScore}(c)$
    \Else
        \State $v \gets \arg\min_{c \in C} \textsc{ExplorationScore}(c)$
    \EndIf
\EndWhile
\If{$\textsc{UnexploredHypotheses}(v) \neq \emptyset$}
    \State Select $h$ from $\textsc{UnexploredHypotheses}(v)$
\Else
    \State Generate hypotheses for $v$ and add them as children
    \State Select $h$ from the newly generated hypotheses
\EndIf
\State \Return $h$
\end{algorithmic}
\end{algorithm}

\subsection{Hypothesis Generation}

As we have noted above, when the topic selection module arrives at a leaf topic node, it may or may not have an associated unexplored hypothesis. If not, the hypothesis generation module must be triggered.

Note, however, that hypothesis generation is an optional step. It is less likely to be triggered for corpus-based exploration, since the concept layer construction step already generates hypotheses for a large number of concepts. It may be triggered more often for Web based exploration, since the Topic Tree construction step does not generate hypotheses. In either case, the hypothesis generation module is critical to ensure that the system can continue generating insights even after exhausting the pool of hypotheses generated during exploration map construction or augmentation.

A simple approach to hypothesis generation is to prompt an LLM with the topic or concept name and description, and Web search results related to it, and ask it to generate a hypothesis. In our implementation, we instead prompt the LLM to generate multiple hypotheses at once for the same topic or concept (see the Hypothesis Generation System Prompt subsection in Appendix \ref{app:exploration-map-system-prompts}). In generating multiple hypotheses at once, the LLM must make them all distinct from each other, thus encouraging diversity. We denote $H = h_1, \ldots, h_k$ as the generated hypotheses batch.
Each generated hypothesis $h \in H$ is graded for the following criteria (see the Hypothesis Scoring System Prompts subsection in Appendix \ref{app:exploration-map-system-prompts}):
\begin{itemize}
    \item \textbf{Relevance}: How relevant is this hypothesis to the goal. In our implementation, an LLM first scores each hypothesis in isolation for relevance to the goal and retrieved evidence, and then assigns relative scores across the batch. This results in the relevance score $r(h)$.
    \item \textbf{Impact}: How impactful is this hypothesis if it turns out to be true. An LLM assigns all hypotheses in the batch relative scores for impact. This results in the impact score $i(h)$.
    \item \textbf{Diversity}: How different is this hypothesis from previously explored hypotheses. This is computed using text embedding similarity between the generated hypothesis and all previously explored hypotheses in the entire exploration map. The diversity score $d(h)$ is then computed as the inverse of the maximum similarity score, thus giving higher scores to more novel hypotheses.
    \item \textbf{Overall Score}: The overall score of a hypothesis is a weighted average of the relevance, impact, and diversity scores: $s(h) = w_r r(h) + w_i i(h) + w_d d(h)$, where $w_r$, $w_i$, and $w_d$ are hyperparameters that can be tuned based on the use case and desired balance between these criteria. \footnote{In our implementation, we set $w_r = 0.5$, $w_i = 0.2$, and $w_d = 0.3$ based on preliminary experiments, but these can be adjusted as needed.}
\end{itemize}

Finally, the hypothesis $h = \arg\max_{h' \in H} s(h')$ with the highest overall score is selected for exploration, and added to the exploration map as a child of the topic node. The remaining hypotheses in the batch are also added as children of the topic node, but they remain unexplored for now. This allows the system to have a pool of hypotheses to explore in future iterations, while still prioritizing the most promising one for immediate exploration.

Table \ref{tab:hypothesis-generation-example} shows a concrete example from a Web-based run. The upper half lists a subset of the retrieved seeds that plausibly motivated the generated batch. The lower half shows the scorecard used to select the next hypothesis.

\begin{table*}[t]
\centering
\caption{Hypothesis-generation example for a sampled leaf topic in exploration. The upper half lists the subset of relevant Web searches which contributed to the generated hypotheses. The lower half shows the resulting hypotheses and their respective scores on all 3 dimensions. 
The selected hypothesis is boldfaced.}
\label{tab:hypothesis-generation-example}
\begingroup
\scriptsize
\setlength{\tabcolsep}{3pt}
\renewcommand{\arraystretch}{1.05}
\begin{tabularx}{\textwidth}{@{}lXl@{}}
\toprule
\multicolumn{3}{@{}l@{}}{\textbf{Tree Path}: \texttt{Methods \& Data > FEW Nexus Modeling > Integrated Assessment}} \\
\midrule
\multicolumn{3}{@{}l@{}}{\textbf{Relevant Retrieved Seeds}} (Web Search results based on Tree Path) \\
\midrule
\textbf{S10} & \textbf{Title}: \emph{Towards an Inclusive Water-Energy-Food Nexus Framework - UNU}. \textbf{Snippet}: The framework enhances access to WEF resources by all, gender equality and social \ldots \newline \textbf{URL}: \href{https://unu.edu/inweh/event/science-talk-towards-inclusive-water-energy-food-nexus-framework}{unu.edu/inweh/event/science-talk-towards-inclusive-water-energy-food-nexus-framework} & Jul~13,~2024 \\
\textbf{S14} & \textbf{Title}: \emph{4 global risks to look out for in the post-pandemic era}. \textbf{Snippet}: The post-pandemic era is being shaped by heightened global risk and unpredictable shock events. \ldots \newline \textbf{URL}: \href{https://www.weforum.org/stories/2024/08/4-global-risks-to-look-out-for-in-the-post-pandemic-era/}{weforum.org/stories/2024/08/4-global-risks-to-look-out-for-in-the-post-pandemic-era} & Aug~19,~2024 \\
\textbf{S15} & \textbf{Title}: \emph{After Shock to After Strategy---a Resilient System or New \ldots}. \textbf{Snippet}: The COVID-19 pandemic produced a foundational jolt to health care. \ldots \newline \textbf{URL}: \href{https://jamanetwork.com/journals/jamanetworkopen/fullarticle/2841286}{jamanetwork.com/journals/jamanetworkopen/fullarticle/2841286} & Nov~12,~2025 \\
\textbf{S23} & \textbf{Title}: \emph{Integrating UN data and AI for Accelerated Recovery - UN-GGIM}. \textbf{Snippet}: Integrating UN data and AI for accelerated recovery. GIS provides solutions for hazards, crops and planning. \ldots \newline \textbf{URL}: \href{https://ggim.un.org/meetings/GGIM-committee/12th-Session/side_events/UNGN_UNDP.pdf}{ggim.un.org/meetings/GGIM-committee/12th-Session/side\_events/UNGN\_UNDP.pdf} & \\
\textbf{S5} & \textbf{Title}: \emph{Nexus approach to enhance water-energy-food security \ldots - Nature}. \textbf{Snippet}: Model-based nexus assessments can compare climate, socio-economic and demographic scenarios. \ldots \newline \textbf{URL}: \href{https://www.nature.com/articles/s44168-025-00308-4}{nature.com/articles/s44168-025-00308-4} & Dec~16,~2025 \\
\textbf{S24} & \textbf{Title}: \emph{Integrating Agriculture in National Adaptation Plans}. \textbf{Snippet}: Climate change adaptation, food security and agriculture are brought together in national planning. \ldots \newline \textbf{URL}: \href{https://openknowledge.fao.org/handle/20.500.14283/bc143e}{openknowledge.fao.org/handle/20.500.14283/bc143e} & \\
\textbf{S12} & \textbf{Title}: \emph{What Next for the Post Covid Global Economy: Could Negative \ldots}. \textbf{Snippet}: Recurrent negative supply shocks imply a dangerous future for other fragile systems. \ldots \newline \textbf{URL}: \href{https://www.ineteconomics.org/research/research-papers/what-next-for-the-post-covid-global-economy-could-negative-supply-shocks-disrupt-other-fragile-systems}{ineteconomics.org/research/research-papers/what-next-for-the-post-covid-global-economy-...} & \\
\textbf{S27} & \textbf{Title}: \emph{Uncovering Systemic Dynamics through an Integrated WEFE Nexus \ldots}. \textbf{Snippet}: The WEFE nexus is operationalized by embedding ecosystems as a quantified pillar. \ldots \newline \textbf{URL}: \href{https://pmc.ncbi.nlm.nih.gov/articles/PMC12930377/}{pmc.ncbi.nlm.nih.gov/articles/PMC12930377/} & Feb~5,~2026 \\
\bottomrule
\end{tabularx}

\vspace{0.5em}

\begin{tabularx}{\textwidth}{@{}>{\raggedright\arraybackslash}X
>{\centering\arraybackslash}p{0.08\textwidth}
>{\centering\arraybackslash}p{0.08\textwidth}
>{\centering\arraybackslash}p{0.08\textwidth}
>{\centering\arraybackslash}p{0.10\textwidth}@{}}
\textbf{Hypothesis Scorecard} & \textbf{Relevance} & \textbf{Impact} & \textbf{Diversity} & \textbf{Overall} \\
\midrule
Post-pandemic expansion of digital health and education services is magnifying gender and rural-urban inequalities: households without reliable broadband face compounded disadvantages that translate into poorer maternal health outcomes and long-term female labor productivity losses. & 5.00 & 4.00 & 4.16 & 4.55 \\
\specialrule{0.25pt}{0pt}{4pt}
Rapid growth in climate-induced South--South migration is shifting labor away from climate-sensitive rural sectors to informal urban services, depressing remittance flows to agrarian regions and increasing nutritional insecurity among those left behind. & 7.00 & 1.00 & 3.35 & 4.70 \\
\specialrule{0.25pt}{0pt}{4pt}
Global fertilizer trade disruptions stemming from geopolitical conflicts and energy price spikes are incentivizing smallholders in tropical forest frontiers to compensate for lower soil fertility by expanding cropland, accelerating deforestation and heightening zoonotic spill-over risks. & 8.00 & 9.00 & 4.41 & 7.12 \\
\specialrule{0.25pt}{0pt}{4pt}
The deployment of AI-driven ``just-in-time'' logistics in global food and medical supply chains is increasing systemic vulnerability: extreme weather events at a few major trans-shipment ports could cause cascading shortages in Small Island Developing States within weeks. & 1.00 & 7.00 & 3.77 & 3.03 \\
\specialrule{0.25pt}{0pt}{4pt}
\textbf{Escalating climatic droughts are simultaneously shrinking hydropower output and straining agricultural water supplies, triggering energy shortages that undermine cold-chain logistics for vaccines and perishable foods in low-income regions, thereby reversing recent gains in immunization coverage and nutrition.} & \textbf{10.00} & \textbf{10.00} & \textbf{3.80} & \textbf{8.14} \\
\bottomrule
\end{tabularx}
\endgroup
\end{table*}

\subsection{Explorer}

Once a hypothesis is selected, the explorer agent takes over. The explorer is a ReAct-style agent that turns the current hypothesis into a candidate insight by iteratively gathering evidence and revising the hypothesis when needed \citep{yao2023react}. 

The explorer agent follows a loop with each iteration following a fixed \textbf{observe--reason--act} pattern. In \emph{observe}, the explorer reviews the evidence obtained so far and how it relates to the current hypothesis. In \emph{reason}, it explains its next steps and what further evidence would potentially clarify the hypothesis. In \emph{act}, it issues tool calls to its various data sources to continue exploring the current working hypothesis. However, if it deems that a sufficiently novel insight has been reached, it may submit a final insight. This structure makes the trace auditable and supports later report generation.

The loop runs for a bounded number of turns, denoted \texttt{max\_explorer\_turns}. If this limit is reached without a submitted insight, the system stops exploration for that hypothesis, and no insight is produced. However, with a sufficiently high \texttt{max\_explorer\_turns}, this becomes unlikely in practice.

\subsubsection{Verbalization}

At each step, the explorer is required to verbalize the full observe-reason-act cycle into a pre-defined format. The system prompt used to enforce this structure is given in Appendix \ref{app:explorer-verifier-loop-system-prompts}, under the Explorer System Prompt subsection.
\begin{itemize}
\item The \emph{observe} section contains the current working hypothesis, a detailed description of new evidence from all tool calls executed so far, and a classification of the hypothesis action. It also produces a list of additional hypotheses that can be added to the hypothesis pool for future exploration. 
\item The \emph{reason} section captures the rationale for the next queries or for ending exploration. This explicit verbalization creates a stable exploration trace that is later used for reporting and for human inspection.
\item The \emph{act} section contains the tool calls for the next step, or a final insight submission. 
\end{itemize}

Verbalization has multiple benefits. First, it creates an auditable trace. Second, it allows the explorer to generate additional hypotheses (which are branched research tracks). Note that when a tool call is made to validate or refute a given hypothesis, the response may contain information that prompts another research track. However, to not detract the current exploration direction, we store the additional hypotheses generated into the pool. This allows \nomad{} to increase the breadth of exploration as it goes deeper, without losing focus on the current track. Finally, verbalization also allows the system to generate a more coherent report at the end of exploration, since it has access to the explicit reasoning at each step.

\subsubsection{Hypothesis Actions}

In the \emph{act} section, the explorer classifies its next action on the hypothesis into one of the following categories:
\begin{itemize}
\item \textbf{Refine}: Maintain the current hypothesis' core focus, with small tweaks or additions in order to explore different angles to find an interesting insight. There is tangential evidence to the current hypothesis.
\item \textbf{Revise}: Significantly change aspects of the current hypothesis, while still maintaining some connection to it. There is direct counter evidence to the current hypothesis.
\item \textbf{Keep}: Continue exploring the current hypothesis without any changes. There is no evidence yet either to support or refute the hypothesis.
\item \textbf{Validate Further}: The current evidence is leaning towards supporting the hypothesis, but more validation is needed before submitting an insight.
\item \textbf{User Response}: The current evidence is sufficient to submit an insight, and the explorer is ending its loop. The submitted insight is included in this action.
\end{itemize}

Hypothesis action classification is based entirely on the evidence collected so far.

\subsubsection{Tool Use}

In the \emph{act} section, the explorer outputs a list of tool calls. The tool list is configurable and can include Web search, document retrieval, or database queries. The explorer may call multiple tools in a step, and  all tool calls are validated before execution. If a tool call is malformed, the system returns an error message to the explorer and continues the loop without executing tools. Otherwise, tool results are appended to the conversation history and the explorer proceeds.

When a tool call completes, a single natural language response is returned by the tool to the explorer. Most responses contain citations - references into underlying documents, URLs, or database entries. The underlying data forms part of the \emph{citation database} - a unified database which maintains citations across all tools. Each citation points to a specific piece of information in the underlying data source, and contains metadata about it (for example, a citation could point to a specific document and include metadata such as the filename, page number, and text span). The explorer can use these citations in its reasoning and final insight submission.

\subsubsection{Tool SubAgents}
\label{sec:tool-subagents}

Tools in \nomad{} are implemented as subagents. Each presents a single natural language interface to the explorer while internally running its own multi-step loop of LLM calls, query refinement, and result aggregation. This design keeps the explorer's context window clean and lets each subagent pursue whatever depth of search a complex question requires. The three primary subagents are the \emph{SQL subagent} for structured databases, the \emph{document search subagent} for unstructured document corpora, and the \emph{Web search subagent} for retrieving information from the open Web.

\paragraph{SQL Subagent:}
The SQL subagent translates a natural language question into an executable SQL query and returns the results with citations (Algorithm~\ref{alg:sql-subagent}). It receives the question along with the full database schema and uses an LLM to produce a structured response containing a query description and the SQL statement. The output is validated before execution; if malformed, a structured error message is fed back to the LLM and the step retries, up to a configurable limit.

A practical challenge is that databases often store values in forms that differ from natural phrasing: abbreviated codes, inconsistent capitalization, or slightly varying names. To address this, the subagent applies a \emph{semantic entity resolution} step on empty results. For each string predicate, it first identifies the relevant text-matching patterns in the query, then embeds the literal value, retrieves column matches by cosine similarity, and uses a secondary LLM call to verify candidates before rewriting the query with exact matches. The final result is returned as a natural language summary with row-level citations. The system prompts used for SQL generation, pattern extraction, semantic match verification, and result analysis are given in Appendix \ref{app:explorer-verifier-loop-system-prompts}, under the SQL subagent system prompts subsection.

\begin{algorithm}[t]
\caption{SQL Subagent}
\label{alg:sql-subagent}
\begin{algorithmic}[1]
\Require Query $q$, Database Schema $S$
\Ensure Response with citations
\For{$\text{attempt} = 1$ \textbf{to} $\texttt{max\_tries}$}
    \State $(\textit{desc},\, \textit{sql}) \leftarrow \text{LLM-Generate-SQL}(q, S)$ \AlgCommentInLine{generate query description and SQL}
    \If{$\textit{sql}$ is well-formed} \textbf{break} \EndIf
    \State feed structured error back to LLM
\EndFor
\State execute $\textit{sql}$
\If{results are empty}
    \State $P \leftarrow \text{LLM-Extract-Patterns}(\textit{sql})$ \AlgCommentInLine{extract text-matching predicates}
    \For{each string predicate $p \in P$}
        \State $\text{candidates} \leftarrow \text{cosine\_search}(\text{embed}(p),\; \text{embed}(column\_values))$
        \State $\text{verified} \leftarrow \text{LLM-Verify-Matches}(\text{candidates},\, p)$ \AlgCommentInLine{semantic entity resolution}
    \EndFor
    \State rewrite $\textit{sql}$ with verified values; execute
\EndIf
\State add results to citation database
\State \Return $\text{LLM-Analyze}(q, \textit{sql}, \text{results})$ with citations
\end{algorithmic}
\end{algorithm}

\paragraph{Document Search Subagent:}
The document search subagent retrieves passages from an unstructured document corpus using one of two strategies, summarized in Algorithm~\ref{alg:doc-search-subagent}. The default mode is \emph{self-reflection}: after an initial vector similarity search, an LLM analyzes the retrieved passages and identifies any gaps. If gaps exist, a query-enhancement step generates targeted follow-up queries and the subagent issues a new retrieval round. This loop repeats for a bounded number of turns before a grounded answer-generation step produces the final answer with citations, following the same general retrieval-critique-revision pattern explored in Self-RAG \citep{asai2024selfrag}.

For queries spanning multiple independent sub-topics, the subagent instead applies \emph{query decomposition}: it first decides whether decomposition is needed, then breaks the question into focused sub-queries, answers each independently, and synthesizes a combined response. In both modes, an optional Hypothetical Document Embeddings(HyDE) step can be enabled: a short hypothetical answer document is generated first, and its embedding is used for retrieval instead of the original query \cite{gao-etal-2023-precise}. All retrieved passages are formatted with source metadata and added to the shared citation database. The system prompts used for decomposition decision, query decomposition, HyDE, self-reflection, query enhancement, answer generation, and synthesis are given in Appendix \ref{app:explorer-verifier-loop-system-prompts}, under the document search subagent system prompts subsection.

\begin{algorithm}[t]
\caption{Document Search Subagent}
\label{alg:doc-search-subagent}
\begin{algorithmic}[1]
\Require Query $q$
\Ensure Response with citations
\If{$\text{LLM-Decide-Decompose}(q)$}
    \State $Q \leftarrow \text{LLM-Decompose}(q)$
    \For{each $q_i \in Q$}
        \State $A_i \leftarrow \text{Retrieve-and-Answer}(q_i)$ \AlgCommentInLine{independent retrieval and grounded answer per sub-query}
    \EndFor
    \State \Return $\text{LLM-Synthesize}(q, A_1, \ldots, A_{|Q|})$
\EndIf
\If{HyDE is enabled}
    \State $h \leftarrow \text{LLM-HyDE}(q)$
    \State $R \leftarrow \text{VectorSearch}(h)$
\Else
    \State $R \leftarrow \text{VectorSearch}(q)$
\EndIf
\For{$\text{turn} = 1$ \textbf{to} $\texttt{max\_turns}$}
    \State $(\textit{analysis}, \textit{done}, \textit{gaps}) \leftarrow \text{LLM-Reflect}(q, R)$
    \If{$\textit{done}$} \textbf{break} \EndIf
    \State $Q_f \leftarrow \text{LLM-Enhance}(q, R, \textit{gaps})$
    \State $R \leftarrow R \cup \text{VectorSearch}(Q_f)$
\EndFor
\State \Return $\text{LLM-Answer}(q, R)$ with citations
\end{algorithmic}
\end{algorithm}

\paragraph{Web Search Subagent:}
This subagent retrieves information from the Web in response to a natural language query. Rather than forwarding the query directly to a search API, it first uses an LLM to expand the query into a set of diverse, focused search queries, each targeting a different angle of the topic. A separate LLM call determines whether the query implies a recency requirement, which is translated into a time-range filter (past day, week, month, or year) applied to each query. The queries are issued in parallel and their results are merged into a globally ranked result set.

In its more capable configuration, the subagent goes beyond snippets: it scrapes the full content of each result page and summarizes each page with a dedicated LLM call instructed to quote directly from the source. A final synthesis step combines these summaries into a coherent answer. This extractive approach grounds the response in verbatim source material, reducing the risk of hallucination. Results are returned to the explorer as a natural language summary with citations into the source URLs, stored in the shared citation database. The system prompts used for search-strategy generation, page summarization, and final aggregation are given in Appendix \ref{app:explorer-verifier-loop-system-prompts}, under the web search subagent system prompts subsection.

All three subagents expose a uniform interface to the explorer: one call in, one response out. Internal loops (retries, entity resolution, reflection turns, decomposition steps, multi-query expansion) are fully encapsulated and never surface in the explorer's conversation history, keeping its context focused while the subagents perform deeper retrieval.

\subsubsection{Insight Submission}

When the explorer decides it has sufficient evidence, it ends its loop by emitting a \texttt{USER\_RESPONSE} tool call. This call contains a concise insight statement and supporting evidence. The system records this as the current insight for verification. If no valid \texttt{USER\_RESPONSE} is produced (for example, due to repeated formatting errors), the current insight remains empty and verification is skipped.

\subsection{Verifier}

Hallucinations are a commonly observed issue in LLM-based agents, especially when they are used for complex reasoning tasks that require integrating information from multiple sources over several turns. If there exists a hallucination in any of the explorer turns, it increases the likelihood of the final report being flawed or incorrect. 

To mitigate this risk, we introduce the verifier agent that evaluates the explorer's insight before it is finalized. This follows recent work on explicit verification and claim decomposition in long-form generation \citep{dhuliawala2024cov,wanner2025dndscore}. The verifier operates independently from the explorer, and has access to the same data sources and tools as the explorer. However, it does not have access to the explorer's internal reasoning or tool calls, and only receives the final insight statement (with citations stripped) for evaluation. The rationale for this design is to ensure that the verifier's evaluation is based solely on the content of the insight and the underlying data, without being influenced by the presence of citation tags.

The verifier is required to award a final faithfulness score to the insight, based on how much of its content could be directly validated. Since it issues independent tool calls to validate or invalidate claims made by the explorer, the likelihood of an explorer phase hallucination passing through the verifier is low. This ends up providing a strong guardrail against hallucinations in the final insight. The verifier also provides feedback on which parts of the insight were well supported and which parts were not, thus providing useful feedback to the explorer for future iterations.

We need to make the verifier score interpretable, reliable, and actionable for the explorer. To remove ambiguity and subjectiveness in the verifier's scoring, we first make it break the insight down into atomic sub-claims it can award boolean scores to. This reduces the allowed variability in the verifier's scoring and makes it more interpretable. The verifier is also prompted to provide feedback on which sub-claims were supported and which were not, along with the reasoning for each. This makes the feedback more actionable for the explorer, as it can directly see which parts of its insight were weak and need improvement. The sub-claim generation and verifier prompts are given in Appendix \ref{app:explorer-verifier-loop-system-prompts}, under the Verifier System Prompts subsection.
The final score is the average of the sub-claim scores, and the feedback is a concatenated summary of the verifier's feedback on all sub-claims.

To verify individual sub-claims, the verifier follows an identical loop as the explorer (without verbalization because we want the verifier to be fast and concise). At each step, it makes tool calls to validate or invalidate the current sub-claim, and based on the tool results, it either continues gathering evidence or assigns a final boolean score to the sub-claim. The verifier is allowed to assign a final score at any point in the loop if it deems that sufficient evidence has been gathered. Just like the explorer, the verifier also has a bounded number of turns, denoted \texttt{max\_verifier\_turns}, to prevent it from running indefinitely. If this limit is reached without a final score being assigned, the verifier assigns a default score of $0$ to the sub-claim, indicating that it could not be validated.

To further reduce the likelihood of hallucinations in the verifier, we configure it to use models that we know to be less prone to hallucinations, even if they are weaker at reasoning or other tasks. This is because the verifier's main role is to provide a reliable check for existing claims and not to synthesize new ones, so we prioritize faithfulness over reasoning ability in its model choice. We evaluate this design choice in Section~\ref{sec:ablation-studies}: report quality and diversity remain broadly stable across verifier models of different sizes and families, although trustworthiness is more sensitive to verifier choice in the top-down setting.

\subsection{Explorer-Verifier Loop}

\begin{figure}[t]
    \centering
    \includegraphics[width=0.9\linewidth]{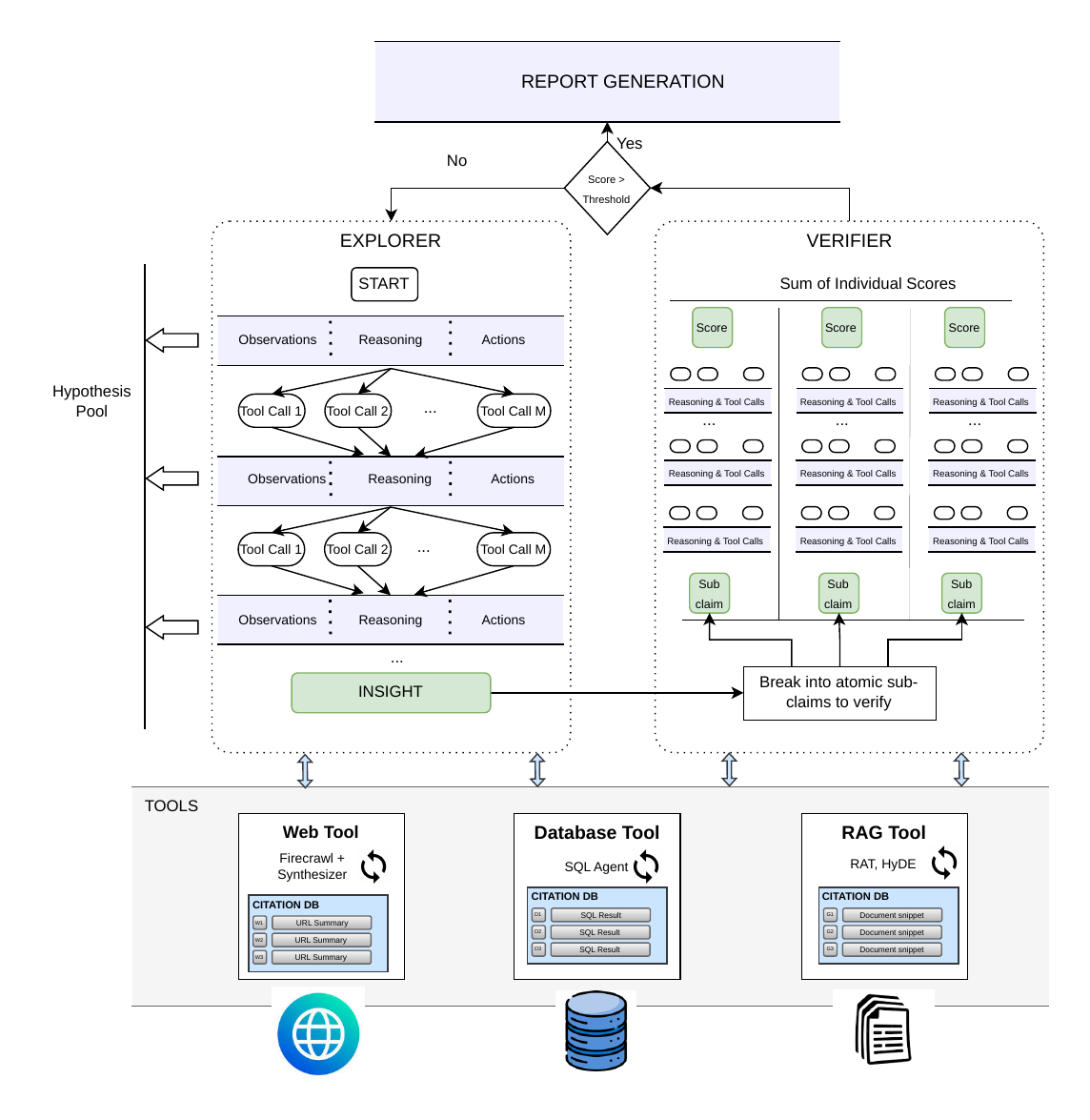}
    \caption{Explorer--verifier loop. The Explorer loops through multiple rounds of \emph{observe-reason-act} workflow making multiple tool calls at each step. Once an insight is generated, the Verifier evaluates it by splitting it into sub-claims and validating each. If the validation passes, the system moves on to report generation. Otherwise, the feedback is used by the explorer to refine the insight. Sources referenced at each step of explorer/verifier are collected into a citation database, and utilized by the report generation.}
    \label{fig:explorer-verifier-loop}
\end{figure}

The explorer and verifier run in a bounded outer loop. Each round calls the explorer to produce a candidate insight, then calls the verifier to score it. The verifier feedback is appended to the explorer conversation history as a user message so that the explorer can refine its next attempt. The loop exits early if the verification score exceeds a minimum threshold. It also stops after a maximum number of rounds \texttt{max\_rounds}. If the score never exceeds the threshold, this \nomad{} run returns no insight. Otherwise, it finalizes the insight and moves on to the Report Generation phase.

\begin{figure*}[p]
    \centering
        \includegraphics[width=\textwidth,height=0.92\textheight,keepaspectratio]{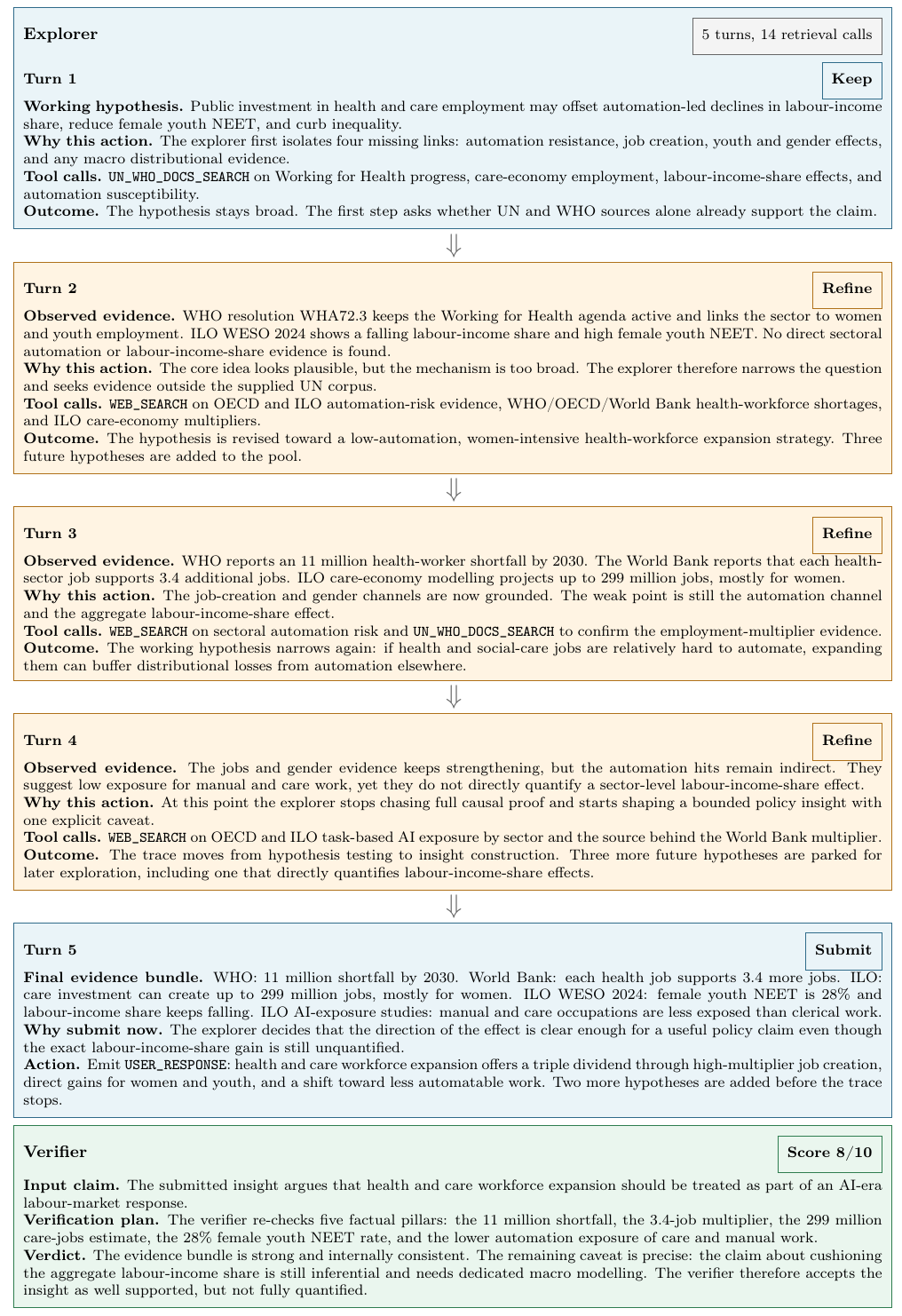}
    \caption{AI-generated summary of a real explorer--verifier trace from the WHO Research Analyst example (initial hypothesis from Table \ref{tab:concept-layer-hypothesis-2}).}
    \label{fig:explorer-trace-example}
\end{figure*}

Figure \ref{fig:explorer-verifier-loop} illustrates both agents and their workflow. Figure \ref{fig:explorer-trace-example} demonstrates an actual \nomad{} run (with AI-generated summaries for readability) for the hypothesis generated in Table \ref{tab:concept-layer-hypothesis-2}. 

\FloatBarrier

\section{Report Generation}

Once exploration is complete and an insight, along with its supporting evidence, is finalized, \nomad{} moves on to the report generation phase. Reports are built to be readable with a clear narrative flow, while also being auditable and traceable to the underlying evidence. The report generation module uses the citation database to provide references and evidence support within the report.

The goal is to transform a short, context-dependent insight into a self-contained artifact. The report must explain the insight, justify it with evidence, and be interpretable without the exploration trace. We therefore treat report generation as a controlled summarization of the full explorer--verifier conversation and the evidence gathered during tool use, in the same broad spirit as prior work on grounded long-form writing and planning \citep{shao2024storm,gu2025rapid}.

Two constraints guide the design. First, the report must follow a coherent narrative with an explicit throughline, so that readers can understand why each section exists and how it advances the argument. Second, every factual claim must be grounded in a traceable source so that the report can be audited. The workflow below makes these constraints explicit at each stage of generation.

\subsection{Report Generation Workflow}
\label{sec:report-generation}

AI-generated reports can be information-heavy and often overwhelming for humans to consume or follow. To maintain readability and coherent train-of-thought, we make several design choices in the report generation workflow.

\begin{figure}[t]
    \centering
    \includegraphics[width=0.95\linewidth]{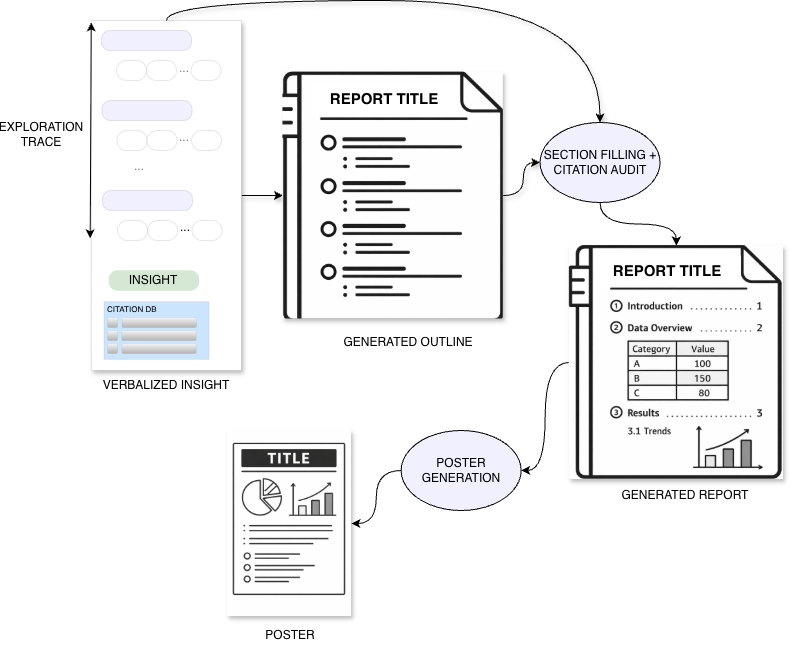}
    \caption{Overview of the report generation workflow. The finalized insight and citation database are first used to draft a report structure. Section filling and citation auditing then iteratively refine this draft into a structured, evidence-grounded report. The audited report is finally transformed into a compact poster that highlights the main insight and key supporting evidence.}
    \label{fig:report-generation}
\end{figure}

Report generation receives the following inputs: the final insight, the conversation history of the explorer (which includes the verifier feedback), and the citation database. It then uses the following workflow to generate the final report (Figure~\ref{fig:report-generation}):

\begin{itemize}
    \item \textbf{Outline Generation}: In the first LLM call, the entire \emph{verbalized insight}, which is the entire explorer-verifier conversation trace with the contents of all cited documents and the final insight, is used to prompt a long context LLM to generate a detailed outline. In this step, the LLM is expected to produce a list of section types and subsections, along with a brief description of what each section will cover.
    \item \textbf{Title and Summary Generation}: Using the outline and the report-level throughline, the model produces a concise title and a summary that highlights the insight, the core evidence, and the main limitations.
    \item \textbf{Section Generation}: The sections planned during the outline generation step are then expanded into full sections. Each section is structured to include the relevant evidence and citations from the citation database, and to align with the section-level throughline.
    \item \textbf{Citation Auditing}: A second pass on each section, with access to the citation database, add any citations that were missed, and checks that all claims are fully supported by the cited evidence. This step also has freedom to minimally edit content - especially to remove unsupported claims - without substantially altering the narrative.
    \item \textbf{Poster Generation}: In addition to the written report, the module also generates a visual poster that summarizes the key insights and evidence in a more digestible format.
\end{itemize}

The system prompts used for outline generation, title and summary generation, section generation, chart generation, and citation auditing are given in Appendix \ref{app:report-generation-system-prompts}, under the Core Report Generation Prompts subsection.

\subsection{Outline Generation}
Outline generation is the highest-leverage step for report quality. We prompt a long-context model with the full verbalized insight and ask it to produce a structured outline. The output is a list of sections and optional subsections, each with a title, a section type, and a short description. 

Each section or subsection in the outline must use one of the following section types:
\begin{itemize}
    \item \textbf{Text}: Paragraph-based section used for detailed explanations, narrative context, and descriptive analysis.
    \item \textbf{Table}: Structured tabular section used to present comparisons, attributes, or dense factual data.
    \item \textbf{Chart}: Visual data section used for standard charts (for example, bar, line, or pie) to highlight patterns and trends. The data for such charts is drawn from the verbalized insight or citation content.
    \item \textbf{Bullet Points}: Concise unordered list used to summarize key points or highlights.
    \item \textbf{Numbered List}: Ordered list used for sequential steps, procedures, or prioritized actions.
    \item \textbf{Categorization}: Grouping section used to organize items into named categories.
    \item \textbf{Measurement}: Metric-focused section used to report judgments or scores (for example, low/medium/high) across categories.
    \item \textbf{Recommendation}: Action section used to provide stakeholder-specific recommendations with justification, impact, and priority.
    \item \textbf{Timeline}: Chronological section used to present events or actions in date order.
\end{itemize}

The outline must also output a \emph{throughline} at the report level and at each section level. The report-level throughline is the main narrative or argument that the report will present. The section-level throughline states how the section advances the report-level throughline and how it connects to neighboring sections. Making the throughline explicit mitigates a common failure mode of long-form generation: a report that is locally coherent but globally meandering. 

The generated outline serves as a scaffold for the subsequent generation steps. Table~\ref{tab:report-outline-example} shows the outline generated for the running WHO example. 

\begin{table}[ht]
    \centering
    \setlength{\tabcolsep}{5pt}
    \renewcommand{\arraystretch}{1.08}
    \small
    \begin{tabularx}{\linewidth}{>{\raggedright\arraybackslash}p{0.09\linewidth}>{\raggedright\arraybackslash}p{0.16\linewidth}>{\raggedright\arraybackslash}X}
        \toprule
        \textbf{Section \#} & \textbf{Section Type} & \textbf{Section Plan} \\
        \midrule
        \multicolumn{3}{c}{Report Title: \textbf{Care Economy’s Triple Dividend}}\\
        \midrule
        1 & Text &
        \begin{minipage}[t]{\linewidth}
        \vspace{0pt}
        \colorbox{myblue!10}{\parbox{\dimexpr\linewidth-2\fboxsep\relax}{\raggedright\textcolor{myblue}{\textbf{The Automation--Inequality Nexus and the Care-Economy Solution}}}}\\[2pt]
        \colorbox{orange!6}{\parbox{\dimexpr\linewidth-2\fboxsep\relax}{\raggedright Summarises ILO WESO evidence on automation pressure and youth NEET stagnation, recaps the WHO 11 million health-worker shortfall, and frames the care economy as a lever for inclusive growth.}}\\[2pt]
        \colorbox{green!6}{\parbox{\dimexpr\linewidth-2\fboxsep\relax}{\raggedright\textit{Introduces the twin challenges of falling labour-income shares and persistent youth NEET rates, then positions health- and care-workforce expansion as a cross-sectoral response that addresses both trends.}}}
        \end{minipage}
        \\
        \addlinespace[3pt]
        2 & Table &
        \begin{minipage}[t]{\linewidth}
        \vspace{0pt}
        \colorbox{myblue!10}{\parbox{\dimexpr\linewidth-2\fboxsep\relax}{\raggedright\textcolor{myblue}{\textbf{Consolidated Evidence for a Triple Dividend}}}}\\[2pt]
        \colorbox{orange!6}{\parbox{\dimexpr\linewidth-2\fboxsep\relax}{\raggedright Plans one compact evidence table with the 11 million shortfall, the 3.4 employment multiplier, the 299 million care-jobs estimate, the female youth-NEET gap, and low automation exposure for care work.}}\\[2pt]
        \colorbox{green!6}{\parbox{\dimexpr\linewidth-2\fboxsep\relax}{\raggedright\textit{Aggregates headline numbers from WHO, World Bank, and ILO to ground the report's argument in concrete, cross-agency evidence.}}}
        \end{minipage}
        \\
        \addlinespace[3pt]
        3 & Chart &
        \begin{minipage}[t]{\linewidth}
        \vspace{0pt}
        \colorbox{myblue!10}{\parbox{\dimexpr\linewidth-2\fboxsep\relax}{\raggedright\textcolor{myblue}{\textbf{Automation Exposure by Sector: Where Health and Care Stand}}}}\\[2pt]
        \colorbox{orange!6}{\parbox{\dimexpr\linewidth-2\fboxsep\relax}{\raggedright Plans a sector comparison chart that places human health and social work near the low-risk end of automation exposure using task-based ILO and OECD evidence.}}\\[2pt]
        \colorbox{green!6}{\parbox{\dimexpr\linewidth-2\fboxsep\relax}{\raggedright\textit{Visualises why shifting employment toward health and care can hedge against AI-driven job displacement, reinforcing the third element of the triple dividend.}}}
        \end{minipage}
        \\
        \addlinespace[3pt]
        4 & Measurement &
        \begin{minipage}[t]{\linewidth}
        \vspace{0pt}
        \colorbox{myblue!10}{\parbox{\dimexpr\linewidth-2\fboxsep\relax}{\raggedright\textcolor{myblue}{\textbf{Quantified Impact Scenario: Meeting the 11 Million Gap by 2030}}}}\\[2pt]
        \colorbox{orange!6}{\parbox{\dimexpr\linewidth-2\fboxsep\relax}{\raggedright Turns the evidence into scenario metrics: direct jobs, indirect jobs from the multiplier, a projected female youth-NEET reduction, and a labour-income-share buffer.}}\\[2pt]
        \colorbox{green!6}{\parbox{\dimexpr\linewidth-2\fboxsep\relax}{\raggedright\textit{Translates evidence into forward-looking metrics to show the scale of possible macro benefits, bridging the gap between descriptive data and policy relevance.}}}
        \end{minipage}
        \\
        \addlinespace[3pt]
        5 & Recommendation &
        \begin{minipage}[t]{\linewidth}
        \vspace{0pt}
        \colorbox{myblue!10}{\parbox{\dimexpr\linewidth-2\fboxsep\relax}{\raggedright\textcolor{myblue}{\textbf{Stakeholder-Specific Policy Recommendations}}}}\\[2pt]
        \colorbox{orange!6}{\parbox{\dimexpr\linewidth-2\fboxsep\relax}{\raggedright Plans stakeholder-specific recommendations for governments, multilateral development banks, education and training systems, and private employers.}}\\[2pt]
        \colorbox{green!6}{\parbox{\dimexpr\linewidth-2\fboxsep\relax}{\raggedright\textit{Provides actionable guidance that operationalises the report's insight for the main decision-makers.}}}
        \end{minipage}
        \\
        \addlinespace[3pt]
        6 & Timeline &
        \begin{minipage}[t]{\linewidth}
        \vspace{0pt}
        \colorbox{myblue!10}{\parbox{\dimexpr\linewidth-2\fboxsep\relax}{\raggedright\textcolor{myblue}{\textbf{Timeline to 2030: Sequencing Investment and Reforms}}}}\\[2pt]
        \colorbox{orange!6}{\parbox{\dimexpr\linewidth-2\fboxsep\relax}{\raggedright Plans dated milestones for workforce audits, financing, training scale-up, service expansion, and a final impact review by 2030.}}\\[2pt]
        \colorbox{green!6}{\parbox{\dimexpr\linewidth-2\fboxsep\relax}{\raggedright\textit{Shows how recommended actions can be phased to realise the triple dividend within the SDG horizon, giving policymakers a coherent road map.}}}
        \end{minipage}
        \\
        \bottomrule
    \end{tabularx}
    \caption{Generated outline, excluding executive summary, for the WHO Research Analyst example in Table \ref{tab:concept-layer-hypothesis-2}. Blue bands show section titles, amber bands show section descriptions, and green bands show section-level throughlines. The report generator fills each section with content one at a time, yet remains tied to the overall narrative using this explicit outline as a scaffold.}
    \label{tab:report-outline-example}
\end{table}

For each section and subsection, the title, description, and throughline serve as system instructions during section generation. The outline in the example of Table \ref{tab:report-outline-example} makes the sequencing explicit: problem framing comes first, then evidence consolidation, then mechanism, then a quantified scenario, then recommendations, and finally a dated implementation path. In this run, the outline also planned a chart section. That section was later declined during section generation because the trace did not provide enough explicit plot-ready values. This illustrates that the scaffold guides the report's structure without forcing unsupported content.

This design is a deliberate tradeoff between flexibility and control. Flexibility comes from dynamic planning, where the model can choose the number, order, and mix of sections based on the evidence, instead of filling a rigid template. The rich variation in section types enables the report to be less of a wall of text, and allows it to be formatted into visually distinct blocks that can be scanned and digested more easily. Control comes from guardrails: each section must use an allowed section type with a fixed schema. These constraints reduce rendering issues in Web and PDF outputs and avoid unusual structures that users may find unfamiliar or hard to follow.  The result is a report format that is adaptive but still predictable and comfortable to consume. 

\subsection{Title and Summary Generation}
The title and summary are generated from the outline and throughline rather than from raw evidence. This ensures that the high-level narrative is consistent with the final report structure. The summary is short and self-contained, and it names the insight, its supporting evidence, and the most important qualification. The summary is used both for user display and for later synthesis into meta-reports.

\subsection{Section Generation}
Sections are generated sequentially. To generate a section's content, we provide the outline, the report-level throughline, and the content produced so far. For the specific section to be generated, the title, description, and section-level throughline are also provided. The generation must follow the specific schema associated with the section type. 
This structure preserves global coherence while allowing local evidence to be inserted where it is most relevant. Within each section, claims are paired with citations drawn from the citation database. 

For chart sections, the model is asked either to decline if the trace does not provide enough explicit evidence to justify a plot, or to generate self-contained plotting code and a caption directly from data points present in the trace.

We treat the citation database as the single source of truth for evidence. Claims that are supported by retrieved evidence are explicitly cited, while unsupported claims are left uncited rather than linked to irrelevant sources.

\subsection{Citation Auditing}
Citation auditing runs after section generation. Although section generation is instructed to add relevant citations, missing or hallucinated citations still occur. During report generation, the model is operating at extremely large contexts, and so instruction following often degrades.

Fine-grained citation attachment and atomic factuality checking have both been shown useful for long-form generation \citep{zhang2025longcite,min2023factscore}. We therefore introduce the citation auditor as a post-processor: it reads current section content, then updates it by adding relevant but missed citations, or removing hallucinated citations. 

The auditor builds a vector index over citation records, chunks report text by sentence, retrieves top-$k$ relevant citations per chunk, and calls the writer model to repair citations. The prompt asks the model to verify citations, and add missing citations close to the supported phrase. If the supported phrase must be minimally updated to align with the additional citation, that is also allowed. If no relevant citation exists for an existing claim, the claim is left uncited rather than attaching an irrelevant source.

\FloatBarrier
\subsection{Poster Generation}
Poster generation converts the full report into a compact visual artifact for rapid consumption. The goal is not to restate every detail. The goal is to surface the report throughline, the core insight, and the strongest supporting evidence in a form that can be scanned quickly.

Conceptually, the poster is a structured projection of the audited report, not a separate reasoning pipeline. High-signal elements are selected from the report and organized into visual blocks such as key findings, evidence highlights, trends, recommendations, and limitations. This keeps the poster aligned with the report narrative while preserving traceability to sources.

We also keep a constrained visual structure for poster content (unlike report content, which is far more flexible). The poster pipeline first prompts a model to extract structured poster content from the audited report: a headline, context, key metrics, insights, actions, risks, and tags, together with an optional selection of at most one chart and at most one table as central evidence. Depending on whether a chart, or table, or both are selected, this structured content is then rendered into one of four fixed visual templates. The system prompts used for poster-content extraction are given in Appendix \ref{app:report-generation-system-prompts}, under the Poster Generation System Prompts subsection.

\subsection{Meta Reports}

\begin{figure}[t]
    \centering
    \includegraphics[width=0.75\linewidth]{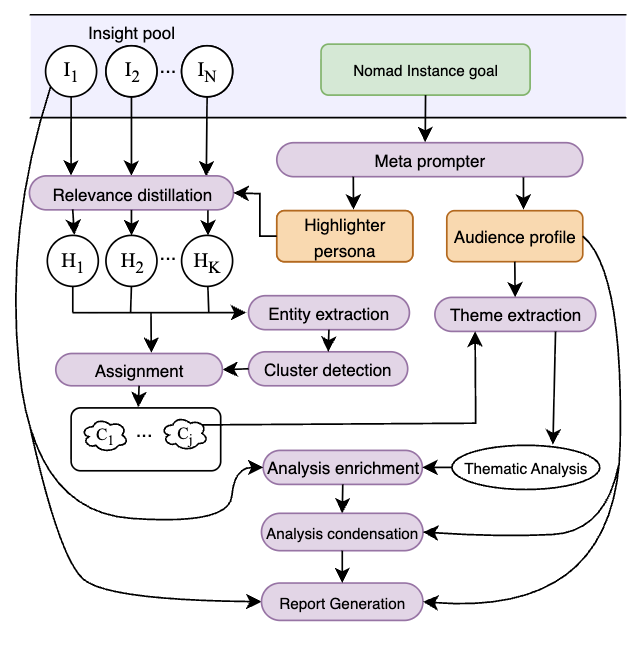}
    \caption{An overview of the meta-report generation pipeline. Inputs from the insight pool ($I_1, \ldots, I_N$), goal from this \nomad{}'s instance (green), feed into the processing stages below. Purple boxes denote pipeline stages: relevance distillation produces $K$ highlights ($H_1, \ldots, H_K$), which undergo entity-based clustering into thematic groups ($C_1, \ldots, C_j$). Each cluster is then refined through thematic analysis, analysis enrichment, and analysis condensation before final report generation.}
    \label{fig:meta-report}
\end{figure}

\begin{figure}[t]
    \centering
    \includegraphics[width=0.95\linewidth]{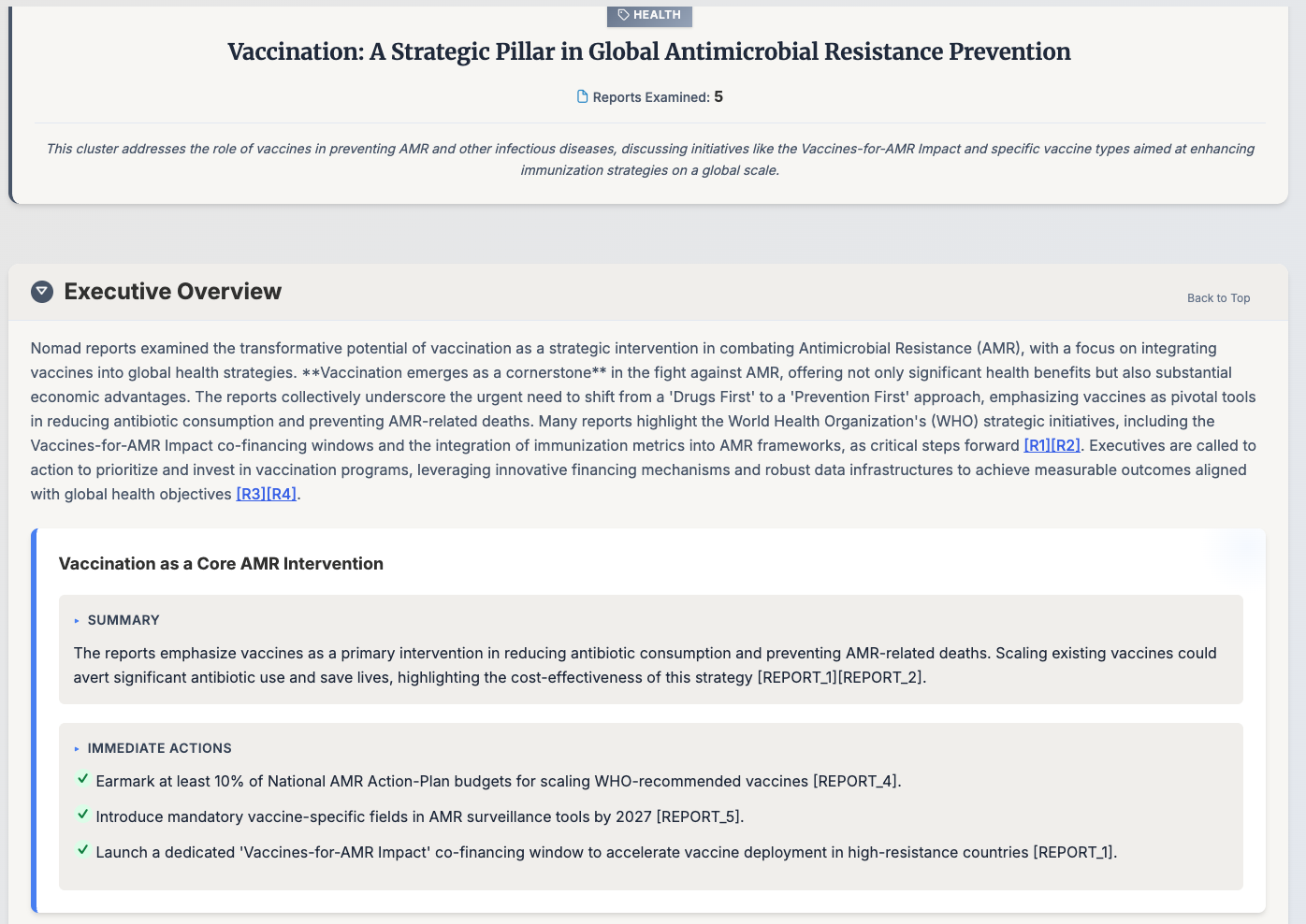}
    \caption{Example generated meta-report for the running WHO analyst example.}
    \label{fig:meta-report-who-example}
\end{figure}

As \nomad{} accumulates insights, a natural opportunity arises to identify recurring themes and cross-cutting patterns across individual reports. Meta-reports synthesize multiple reports into a higher-level narrative, generated periodically from a set of completed reports, their summaries, and their citations. The goal is to surface themes, trends, and open questions that cut across individual insights. To this end, \nomad{} produces \emph{meta-reports} (Figure~\ref{fig:meta-report}, Algorithm~\ref{alg:meta-report}) through three mechanisms: (i)~\emph{relevance distillation}, which applies an audience-aware lens derived from the instance's exploration context to extract relevant highlights; (ii)~\emph{entity-based thematic clustering}, which groups highlights by semantic entities rather than surface similarity; and (iii)~\emph{progressive refinement synthesis}, a three-pass strategy that incrementally condenses heterogeneous evidence through structure, depth, and focus stages. Citations are preserved at the report level, and each meta-report links claims back to the originating reports and their sources, maintaining provenance while enabling cross-report reasoning.
The system prompts used for persona generation, highlight extraction, clustering, synthesis, and report rendering are given in Appendix \ref{app:report-generation-system-prompts}, under the Meta-Report System Prompts subsection.
Figure~\ref{fig:meta-report-who-example} shows one generated meta-report for the running WHO analyst example. It illustrates the final artifact that the user sees after the synthesis pipeline has combined multiple report-level insights into a single higher-level narrative.

\subsubsection{Relevance Distillation}

Relevance is audience-dependent: different stakeholders prioritize different aspects of the same insight. Rather than applying a generic summarizer, which compresses content regardless of the audience, the pipeline uses a \emph{metaprompter} to derive two audience-specific artifacts from the \nomad{} instance's goal:

\begin{itemize}
    \item \textbf{Highlighter persona} $\Phi$: A persona prompt that encodes the target reader's relevance criteria. Different from a generic summarization prompt, the highlighter is an extractive summarizer focused on the goal, which retains only the information pertinent to the reader's interest.
    \item \textbf{Audience profile} $A$: A natural-language description of the target reader's role, concerns, and information needs, kept for use in downstream stages to ensure relevance to the user.
\end{itemize}

The highlighter persona $\Phi$ is applied to each insight in parallel, producing a concise highlight $H_i$ from each insight $I_i$. By adopting the reader's perspective, $\Phi$ retains only the information pertinent to the target audience. This lets us fit the highlights within the clustering step's context window (Section~\ref{sec:meta-clustering}) while preserving relevance to the reader's interest. The audience profile $A$ is carried forward to later stages, where it guides how the analysis and reports are generated and polished.

\subsubsection{Entity-Based Thematic Clustering}
\label{sec:meta-clustering}

The highlights must be organized into coherent thematic groups before synthesis. Because \nomad{}'s breadth-first exploration strategy produces insights spanning heterogeneous subtopics, embedding-based similarity conflates topically distinct insights that share surface vocabulary. The pipeline instead uses an \emph{entity-based} clustering approach: entities capture semantic relationships rather than lexical co-occurrence, yielding more meaningful thematic groupings. The approach proceeds in three steps:

\begin{enumerate}
    \item \textbf{Entity extraction}: For each highlight $H_i$, an LLM extracts named entities (people, organizations, topics, events, locations) along with their types, producing an entity set $\mathcal{E}_i$. This lifts the representation from free text to structured semantic anchors that capture the actual subjects of each insight.
    \item \textbf{Cluster detection}: An LLM receives all entities $\{\mathcal{E}_1,\ldots,\mathcal{E}_N\}$ and identifies $k$ thematic clusters, each with a descriptive label. Optionally, the user may supply seed keywords to guide the LLM toward particular areas of interest.
    \item \textbf{Assignment}: Each highlight is assigned to one or more clusters by an LLM that receives the cluster definitions alongside the highlight content. Unmatched highlights are dropped; empty clusters are pruned.
\end{enumerate}

Steps 1 and 3 are parallelized across all $N$ insights; only step 2 (Cluster detection) requires a single sequential call that sees the full entity landscape. As an alternative, the pipeline also supports manual mode, where the user directly specifies clusters via configuration, bypassing the entity-based steps entirely.

\subsubsection{Progressive Refinement Synthesis}

For each cluster $C_j$, the pipeline produces a condensed summary through three progressive refinement stages. A single-shot summarization approach is insufficient here because the inputs are heterogeneous: highlights are concise but stripped of supporting context, while full insight reports are detailed but lengthy. A single call either loses detail (if working from highlights alone) or exceeds context limits (if working from full reports). The three-pass design decomposes the task so that each pass serves a distinct function---structure, depth, and focus---that a single call cannot balance.

\paragraph{Theme extraction.}
All highlights assigned to $C_j$ are concatenated and provided to an LLM along with the audience profile $A$. The model produces a thematic analysis (500--1000 words) identifying the key themes, patterns, and connections within the cluster. This pass establishes the structural skeleton of the summary from the concise highlights.

\paragraph{Analysis enrichment.}
The thematic analysis is augmented with additional detail from the full (non-highlighted) insight reports in the cluster. Reports are ranked by length (shorter first) and greedily included up to a token budget to remain within the model's context window. The LLM adds supporting evidence, specific data points, and deeper context from these full reports to strengthen the analysis. If no reports fit within the budget, this step is skipped. This pass adds depth that was lost during highlight extraction.

\paragraph{Analysis condensation.}
The enriched analysis is compressed into a concise summary (300--500 words) that retains only the most important findings, again using the audience profile $A$ to guide what is preserved versus discarded. This pass sharpens focus for the target audience.

The three sub-stages run sequentially within each cluster, but clusters are independent and can be processed in parallel.

\subsubsection{Report Rendering with Context Accumulation}

The concise summary from the above analysis condensation stage is expanded into a structured multi-section report through iterative LLM refinement. We use a similar context accumulation approach as in (Section~\ref{sec:report-generation}). Each section is generated using the full conversation history of all previously generated sections in LLM context as well as the insight's text. This expansion from a condensed analysis allows the meta-report to follow a cohesive narrative throughline, while at the same time grounding the report in the original insights. The final meta-report is then rendered with citations that link back to the original insight reports and their sources, preserving traceability while enabling cross-report reasoning.

\begin{algorithm}[t]
\caption{Meta-Report Generation}
\label{alg:meta-report}
\begin{algorithmic}[1]
\Require Insight pool $\mathcal{I} = \{I_1,\ldots,I_N\}$, cluster count $k$
\Ensure Meta-reports $\mathcal{R}$
\State Extract report text $t_i$ from each $I_i \in \mathcal{I}$
\State Sample $\mathcal{S} \subset \mathcal{I}$; generate audience profile $A$ and highlighter persona $\Phi$ via meta prompter from goal of $\mathcal{S}$
\ForAll{$I_i \in \mathcal{I}$} \Comment{parallel $\times N$}
    \State Extract highlight $H_i$ from $t_i$ using persona $\Phi$
\EndFor
\ForAll{$H_i$} \Comment{parallel $\times N$}
    \State Extract entity set $\mathcal{E}_i$ from $H_i$
\EndFor
\State Identify $k$ clusters $\{C_1,\ldots,C_k\}$ from entity mappings $\{\mathcal{E}_i\}$
\ForAll{$H_i$} \Comment{parallel $\times N$}
    \State Assign $H_i$ to best-matching cluster(s) in $\{C_1,\ldots,C_k\}$
\EndFor
\State $\mathcal{R} \gets \emptyset$
\ForAll{non-empty cluster $C_j$} \Comment{parallel $\times k$}
    \State Extract themes $\theta_j$ from $\{H_i : I_i \in C_j\}$ using audience profile $A$
    \State Enrich $\theta_j$ with full reports $\{t_i : I_i \in C_j\}$ up to token budget
    \State Condense $\theta_j$ into summary $\sigma_j$ using audience profile $A$
    \State Generate report sections sequentially with context accumulation
    \State Render report $R_j$; $\mathcal{R} \gets \mathcal{R} \cup \{R_j\}$
\EndFor
\State \Return $\mathcal{R}$
\end{algorithmic}
\end{algorithm}

\FloatBarrier

\section{Evaluations}
\label{sec:evaluations}

\subsection{Evaluation Methods}
Evaluation of long-form, deep-research reports is a challenging problem with some prior work (see Section \ref{sec:related-work}) but no established best practices. We propose a multi-pronged evaluation framework that combines multiple dimensions of evaluations inspired by other methods in the literature.

\subsubsection{Trustworthiness}
The first dimension is \textbf{trustworthiness}, which measures the factual accuracy and reliability of the claims in the reports. Hallucination is a common failure mode for LLM-based report generation, and it can be especially problematic in long-form reports that integrate information from multiple sources. This becomes a more serious problem when quantities are hallucinated or misrepresented. Trustworthiness of a report is higher if its claims are well-supported, both semantically and factually (numeric groundedness).

We use two complementary techniques to measure trustworthiness of a report. 
\paragraph{Numeric Grounding:} To address the issue of numeric hallucination, such as made up dollar values, years and percentages, we define the concept of `numeric groundedness' and use a non-LLM driven, rule-based evaluation approach to measure it. The numeric groundedness of a report is high if the set of numeric claims it makes are well supported by the content of the citations made \emph{near} the claim in the report. The numeric groundedness of a report is an aggregate of the numeric groundedness scores of its sections. The numeric groundedness of a section, in turn, is computed by aggregating the scores of individual \emph{units} within that section. A unit is defined as follows: For tabular sections, a unit is a single row of the table. For itemized or numbered lists, a unit is a single item. For free-form text or other types of sections, a unit is a single sentence. In the following, we describe the numeric groundedness evaluation process for a single unit.

For each unit, we first extract all numbers within the unit. A number is defined as a sequence of digits, possibly containing a leading negative sign or currency symbol, or commas and decimals and an optional multiplier suffix (for example, \texttt{K}, \texttt{M}, \texttt{million}, or \texttt{\%}). Years are tagged separately with a set of similarly defined rules based on context and patterns. The extracted numeric values are normalized to a standard format (for example, \texttt{\$1.5M} would be normalized to \texttt{1500000}). Similarly, we extract all numbers from the content of all cited sources within the report. 

For each extracted numeric in a unit, the system searches for it in source content through a tiered strategy:
\begin{enumerate}
    \item Unit Citations: The numeric value is searched in the content of citations within the same unit. If found as an exact match or as normalized value, this numeric claim is considered grounded, and marked as \texttt{ref}. This indicates strongest evidence.
    \item Section Citations: If the numeric value is not found in unit citations, the system searches for it in citations within the same section. If found in the same section, and the original unit had no citations, then this numeric claim is considered grounded, and marked as \texttt{sec\_ref}. This indicates the claim was cited, but not in the same unit. However, if the original unit had a citation, then this means this numeric claim was misattributed (i.e., its nearest citation doesn't support it but a further citation does). In this case the claim is marked as \texttt{misattributed\_section}.
    \item Report-wide Citations: If still not found, the system searches for the numeric value in any citation within the report. If found in another part of the report, this numeric claim is marked as \texttt{report\_ref} or \texttt{misattributed\_report} depending on whether the original unit had a citation or not. This indicates the claim was cited, but not in the same section or unit.
    \item Previous Sections: If not found in any citations at all, the claim is searched for in the text of previous sections of the report. If found, this numeric claim is marked as \texttt{prev\_section}. This indicates the claim is not directly supported by a citation, but is supported by earlier content within the report. This avoids penalizing a report for repeating a numeric claim across multiple sections without re-citing it.
    \item Explorer History: If still not found, the system searches for the numeric value in the explorer conversation history. If found, this numeric claim is marked as \texttt{explorer}. This indicates the claim is not supported by a citation or previous section, but is supported by the explorer's working hypothesis and evidence.
\end{enumerate}

For numeric claims that are not found in any of these sources, these are considered ungrounded and get one of the following tags:
\begin{enumerate}
    \item \texttt{incorrect\_ref}: citations exist and have retrievable content, but the numeric claim is not found in any of the cited content. This indicates a failure of the citations to support the claim.
    \item \texttt{no\_ref}: no citations exist anywhere for this claim.
    \item \texttt{unverified}: citations exist but their content is not retrievable (for example, due to a broken link or hallucinated links). This indicates an inability to verify the claim's support rather than a direct failure of support.
\end{enumerate}

Each numeric claim thus gets a score based on its tag, with stronger evidence tags getting higher scores. Table \ref{tab:numeric-grounding-tags} gives the tags and their corresponding scores $w(.)$. 

The numeric groundedness of a section $S$ with verification tags $T(S)$ is defined as $$NG(S) = \sum_{t \in T(S)} w(t) \cdot \frac{1}{|T(S)|}$$
The numeric groundedness of a report $R$ with sections $S(R)$ is defined as $$NG(R) = \frac{\sum_{S \in S(R)} NG(S)*|T(S)|}{\sum_{S \in S(R)} |T(S)|}$$
In our implementation, numeric grounding is computed on its native $[0,1]$ scale. In the summary tables, we report the linearly normalized $0$--$100$ version for comparability with the other metrics, i.e., a raw score $x$ is reported as $100x$.

\begin{table}[ht!]
\centering
\caption{Numeric grounding evaluation tags and their corresponding scores.}
\label{tab:numeric-grounding-tags}
\begin{tabular}{lc}
\toprule
\textbf{Tag $t$} & \textbf{Score $w(t)$} \\
\midrule
\texttt{ref} & $1.0$ \\
\texttt{sec\_ref} & $0.8$ \\
\texttt{report\_ref} & $0.7$ \\
\texttt{prev\_section} & $0.5$ \\
\texttt{explorer} & $0.5$ \\
\texttt{misattributed\_section} & $0.4$ \\
\texttt{misattributed\_report} & $0.2$ \\
\texttt{incorrect\_ref} & $0.0$ \\
\texttt{no\_ref} & $0.0$ \\
\texttt{unverified} & $0.1$ \\
\bottomrule
\end{tabular}
\end{table}

\paragraph{Factuality:} The objective of factuality evaluation is to measure whether the factual claims in the report are semantically supported by the cited sources. A report could have perfect numeric groundedness but still be untrustworthy if its claims are not semantically supported by the cited content. For example, if a sentence says ``revenue grew 15\% \texttt{[3]}'' and the citation content of \texttt{[3]} mentions a 15\% growth in users but not revenue, then this is a factuality failure.

We adapt the factuality evaluation protocol of Fan et al. \citep{fan2025understanding}. We first extract all verifiable claims from each section. A verifiable claim is a sentence or clause containing a specific fact that can be checked against a source, such as a quantity, date, named entity, or event. Claims are extracted with an LLM that returns the claim text, the verifiable entities, and the attached citation keys. The factuality of a report is an aggregate of the factuality scores of its sections. The factuality of a section is computed by aggregating the scores of the claims extracted from that section. The verifiable-claim extraction prompt and the factuality judge prompts are given in Appendix \ref{app:evaluation-prompt}, under the Factuality Evaluation Prompts subsection. In the following, we describe the factuality evaluation process for a single claim.

For each cited claim, we retrieve the content of each cited source. We first use content already stored in the report artifacts. If such content is unavailable but a URL is present, we fetch the page content and truncate it to a fixed length. Each claim--source pair is then passed to an LLM judge using the prompt of Fan et al. \citep{fan2025understanding}. The judge receives only the claim and the retrieved source content. It returns a ternary support score: $1$ if the claim is fully supported, $0$ if it is partially supported, and $-1$ if it is not supported. It also returns the source phrase that supports the decision.

For a claim $c$ with pairwise support scores $P(c)$, the final claim verdict is obtained through the following rules:
\begin{enumerate}
    \item If $c$ has no citations, then it is marked as \texttt{UNCITED}.
    \item If citations exist but no cited content can be retrieved, then it is marked as \texttt{UNVERIFIABLE}.
    \item Otherwise, if $\max P(c)=1$, then it is marked as \texttt{SUPPORTED}.
    \item Otherwise, if $\max P(c)=0$, then it is marked as \texttt{PARTIAL}.
    \item Otherwise, it is marked as \texttt{UNSUPPORTED}.
\end{enumerate}

We use the maximum over cited sources because a claim is semantically grounded if at least one of its citations supports it. Each claim verdict is then mapped to a score in $[0,1]$. Table \ref{tab:factuality-verdicts} gives the verdicts and their corresponding scores $Q(.)$. We assign non-zero scores to \texttt{UNCITED} and \texttt{UNVERIFIABLE} because these indicate missing attribution or unavailable source content rather than direct contradiction.

The factuality of a section $S$ with claims $C(S)$ is defined as 
$$F(S) = \sum_{c \in C(S)} Q(c) \cdot \frac{1}{|C(S)|}$$
The factuality of a report $R$ with sections $S(R)$ is defined as 
$$F(R) = \frac{\sum_{S \in S(R)} F(S)\cdot|C(S)|}{\sum_{S \in S(R)} |C(S)|}$$
As with numeric grounding, factuality is computed on its native $[0,1]$ scale and linearly normalized to $0$--$100$ in the summary tables, i.e., a raw score $x$ is reported as $100x$.

\begin{table}[ht!]
\centering
\caption{Factuality evaluation verdicts and their corresponding scores.}
\label{tab:factuality-verdicts}
\begin{tabular}{lc}
\toprule
\textbf{Verdict $v$} & \textbf{Score $Q(v)$} \\
\midrule
\texttt{SUPPORTED} & $1.0$ \\
\texttt{PARTIAL} & $0.5$ \\
\texttt{UNCITED} & $0.3$ \\
\texttt{UNVERIFIABLE} & $0.25$ \\
\texttt{UNSUPPORTED} & $0.0$ \\
\bottomrule
\end{tabular}
\end{table}

\subsubsection{Quality}
Quality is intended to measure how useful and well-written a report section is beyond factual support alone. A section may be factually correct but still be unhelpful if it lacks analysis, omits important aspects of the topic, or fails to communicate its findings clearly. We therefore evaluate each report section with an LLM judge across five dimensions: \emph{analytical quality}, \emph{originality}, \emph{coverage}, \emph{actionability}, and \emph{presentation}.

One approach to using an LLM judge for quality evaluation is to prompt it to output a score for each dimension based on a rubric. For example, the judge could be asked to rate analytical quality on a scale of 1 to 5 by first emitting its reasoning over the section content, then classifying the quality into one of five classes: score $1$ if the section is entirely descriptive and shows no analytical depth, to score $5$ if it shows deep, multi-layered analysis of underlying drivers and far-reaching consequences. While this approach is straightforward, we observe it to suffer from high variance and consistency issues. The reasoning emitted by the judge ends up focusing on different aspects of the report across different runs, leading to inconsistent scores. One way to avoid this is to use a non-sampling generation method for the judge (e.g., greedy sampling or temperature $=0$). However, this is a symptomatic fix and does not address the underlying issue of the judge's reasoning being unfocused and inconsistent across runs. Moreover, when the input report is too lengthy, it becomes even more unpredictable which aspects the judge will focus on in its reasoning, leading to even higher variance.

We use two techniques to address the above. First, we judge each section individually. Yet, in grading a section, we still provide the full report first, and then repeat the to-be-graded section as the `marked section' to focus the judge's attention. This is to ensure that when a section content draws logical conclusions based on other sections, or when it points the reader to other sections, it is not penalized for logical incompleteness or inconsistency. Second, we construct - for each of the above 5 dimensions - a large set of detailed \emph{attributes}. These attributes are binary, that is the LLM judge must award either a yes or no for each attribute. As an example, an attribute for analytical quality is `Demonstrates a clear and logical causal chain', while another is `Leaves key assumptions implicit'. Thus attributes can be positive or negative, and the LLM judge is now free to activate both positive and negative attributes for a given section, and as many or as few attributes as suitable. Each attribute, depending on its desirability or lack thereof, is mapped to a score in $(-2, -1, 1, 2)$, and the final score for each dimension is obtained by summing the scores of the activated attributes for that dimension. This approach allows the judge to be more consistent (binary scores are more stable), and for the final aggregated scores to be more interpretable. We obtain the attribute sets for each dimension by first allowing LLM judges to generate free-form reasoning on text sections, then extracting attributes from these reasoning traces with a separate LLM. The extracted sets were then manually reviewed, refined, and scored by human evaluators. The runtime section-level quality-evaluation prompt is given in Appendix \ref{app:evaluation-prompt}, under the Quality Evaluation Prompts subsection. Tables \ref{tab:quality-attributes-1}, \ref{tab:quality-attributes-2}, and \ref{tab:quality-attributes-3} list the quality-evaluation attributes and their associated scores.
In the summary tables, we report the final aggregated quality scores after linear normalization to a $0$--$100$ scale.

\begin{table*}[t]
\centering
\caption{Quality-evaluation attributes and scores for analytical quality and originality.}
\label{tab:quality-attributes-1}
{\small
\renewcommand{\arraystretch}{1.1}
\noindent\textbf{Analytical Quality}
\par
\begin{tabular}{@{}p{0.72\textwidth}lc@{}}
\toprule
Keyphrase & Class & Score \\
\midrule
Demonstrates a clear and logical causal chain. & Very Good & +2 \\
Defines limitations and boundary conditions explicitly. & Very Good & +2 \\
Provides strong and comprehensive citation coverage. & Very Good & +2 \\
Shows mostly sound reasoning throughout. & Good & +1 \\
Briefly acknowledges limitations. & Good & +1 \\
Supports most claims with evidence. & Good & +1 \\
Contains noticeable reasoning gaps. & Bad & -1 \\
Leaves key assumptions implicit. & Bad & -1 \\
Presents findings as established facts without justification. & Bad & -1 \\
Displays major reasoning gaps. & Very Bad & -2 \\
Fails to discuss limitations. & Very Bad & -2 \\
Makes unsupported quantitative claims. & Very Bad & -2 \\
\bottomrule
\end{tabular}

\vspace{0.75em}
\noindent\textbf{Originality}
\par
\begin{tabular}{@{}p{0.72\textwidth}lc@{}}
\toprule
Keyphrase & Class & Score \\
\midrule
Offers a distinctive and imaginative perspective. & Very Good & +2 \\
Presents clearly non-obvious ideas. & Very Good & +2 \\
Demonstrates strong creative reframing. & Very Good & +2 \\
Provides a fresh and engaging perspective. & Good & +1 \\
Introduces somewhat original ideas. & Good & +1 \\
Incorporates novel framing. & Good & +1 \\
Shows only minor originality. & Bad & -1 \\
Presents fairly conventional ideas. & Bad & -1 \\
Includes minor hints of newness. & Bad & -1 \\
Entirely unoriginal or derivative. & Very Bad & -2 \\
Offers nothing truly surprising. & Very Bad & -2 \\
Delivers already familiar insights. & Very Bad & -2 \\
\bottomrule
\end{tabular}
}
\end{table*}

\begin{table*}[t]
\centering
\caption{Quality-evaluation attributes and scores for coverage and actionability.}
\label{tab:quality-attributes-2}
{\small
\renewcommand{\arraystretch}{1.1}
\noindent\textbf{Coverage}
\par
\begin{tabular}{@{}p{0.72\textwidth}lc@{}}
\toprule
Keyphrase & Class & Score \\
\midrule
Provides broad, multi-angle coverage. & Very Good & +2 \\
Thoroughly explores the central hypothesis. & Very Good & +2 \\
Aligns strongly with the topic and stated goal. & Very Good & +2 \\
Covers most of the relevant dimensions. & Good & +1 \\
Delivers strong topic coverage. & Good & +1 \\
Presents generally relevant and focused content. & Good & +1 \\
Exhibits noticeable gaps in coverage. & Bad & -1 \\
Includes minor irrelevant details. & Bad & -1 \\
Leaves several dimensions under-explored. & Bad & -1 \\
Omits major dimensions of the topic. & Very Bad & -2 \\
Provides narrow or overly selective coverage. & Very Bad & -2 \\
Contains largely off-topic discussions. & Very Bad & -2 \\
\bottomrule
\end{tabular}

\vspace{0.75em}
\noindent\textbf{Actionability}
\par
\begin{tabular}{@{}p{0.72\textwidth}lc@{}}
\toprule
Keyphrase & Class & Score \\
\midrule
Offers directly implementable recommendations. & Very Good & +2 \\
Clearly advances the user's goal. & Very Good & +2 \\
Demonstrates high practical relevance for the target audience. & Very Good & +2 \\
Provides actionable and practical insights. & Good & +1 \\
Moderately supports goal achievement. & Good & +1 \\
Gives clear direction with limited operational detail. & Good & +1 \\
Contains vague or weakly actionable suggestions. & Bad & -1 \\
Provides minimal operational guidance. & Bad & -1 \\
Suggests sound but incremental actions. & Bad & -1 \\
Lacks a clear path to implementation. & Very Bad & -2 \\
Fails to meaningfully advance goals. & Very Bad & -2 \\
Remains entirely theoretical or descriptive. & Very Bad & -2 \\
\bottomrule
\end{tabular}
}
\end{table*}

\begin{table*}[t]
\centering
\caption{Quality-evaluation attributes and scores for presentation.}
\label{tab:quality-attributes-3}
{\small
\renewcommand{\arraystretch}{1.1}
\begin{tabular}{@{}p{0.72\textwidth}lc@{}}
\toprule
Keyphrase & Class & Score \\
\midrule
Maintains an engaging and professional tone. & Very Good & +2 \\
Features a clear and well-structured structure. & Very Good & +2 \\
Uses highly readable and professional language. & Very Good & +2 \\
Adopts a generally engaging tone. & Good & +1 \\
Demonstrates mostly well-structured organization. & Good & +1 \\
Communicates clearly and effectively. & Good & +1 \\
Shows uneven tone or inconsistent style. & Bad & -1 \\
Contains somewhat disjointed sections. & Bad & -1 \\
Occasionally confusing in expression. & Bad & -1 \\
Exhibits a flat or disengaging tone. & Very Bad & -2 \\
Displays poor overall structure. & Very Bad & -2 \\
Is difficult to read or follow. & Very Bad & -2 \\
\bottomrule
\end{tabular}
}
\end{table*}

\FloatBarrier

\subsubsection{Distinctness}
Distinctness measures the extent to which different sections of a report contribute non-overlapping information, analysis, or viewpoints. It is the complementary notion to redundancy: high distinctness means low repetition across sections. We adapt the redundancy evaluation protocol of Fan et al. \citep{fan2025understanding} to detect within-report repetition across sections, but report the resulting metric as distinctness because its score is interpreted in the higher-is-better direction.

The objective is to identify pairs of sections that convey essentially the same information, even when the wording differs. This includes repeated viewpoints, repeated examples or references, and implicit repetition where the same conclusion or argument is restated in paraphrased form. It does not include sections that address related topics but contribute different content, or cases where one section extends another with genuinely new information.

\paragraph{Section Filtering:} Not all sections should be compared. Some overlap is expected by construction and should not be penalized. We therefore first remove sections that are likely to repeat content by design or that do not contain enough substantive text for a meaningful comparison. Specifically, we exclude the title and executive summary, conclusion-style sections detected by title matching (for example, ``conclusion'' or ``key takeaways''), sections shorter than $200$ characters, and URL-heavy sections in which more than $30\%$ of words are URLs. The remaining sections form the candidate set for distinctness evaluation.

\paragraph{Pair Sampling:} If a report has $n$ retained sections, then there are $\frac{n(n-1)}{2}$ possible section pairs. Evaluating all pairs is feasible for short reports, but it becomes unnecessarily expensive for long reports. We therefore define a maximum number of evaluated pairs $K$, with default value $K=30$. If the total number of valid pairs is at most $K$, we evaluate all of them. Otherwise, we sample $K$ unique section pairs uniformly at random without replacement using a fixed seed for reproducibility.

\paragraph{Pairwise Judging:} Each sampled pair of sections is flattened to plain text and passed to an LLM judge using a prompt adapted from Fan et al. \citep{fan2025understanding}. The judge is instructed to focus on repetition at the information level rather than at the surface-form level. In particular, it checks for repeated viewpoints, repeated examples, repeated facts, repeated references, and similar logical flow. It is also explicitly instructed not to rely on external knowledge, and not to count a shared example as repetition if the two sections use that example to explain different concepts. The distinctness judge system and user prompts are given in Appendix \ref{app:evaluation-prompt}, under the Distinctness Evaluation Prompts subsection.

For each section pair $(S_i,S_j)$, the judge returns a distinctness score $DS(S_i,S_j)\in\{0,1,2,3,4\}$ together with a natural-language explanation and the specific repeated content it found. Lower scores indicate more redundancy and therefore lower distinctness, while higher scores indicate that the two sections are more distinct. Invalid judge outputs are retried up to a fixed retry budget. Table \ref{tab:distinctness-scores} gives the scoring scale.

\begin{table}[ht!]
\centering
\caption{Distinctness evaluation scores and their interpretation.}
\label{tab:distinctness-scores}
\begin{tabular}{lp{9.5cm}}
\toprule
\textbf{Score} & \textbf{Meaning} \\
\midrule
$4$ & Almost no repetition. The sections are effectively independent. \\
$3$ & Slight repetition. There are one or two minor overlaps, but the sections remain clearly distinct. \\
$2$ & Some repetition. There are multiple overlapping points, and the sections are only moderately distinct. \\
$1$ & Severe repetition. A large amount of content is repeated, and distinctness is low. \\
$0$ & Excessive repetition. The two sections are nearly entirely repetitive. \\
\bottomrule
\end{tabular}
\end{table}

\paragraph{Report-Level Aggregation:} Let $\Pi(R)$ be the set of evaluated section pairs for report $R$. We define the report distinctness score as the mean pairwise score
$$DS(R) = \sum_{(S_i,S_j)\in \Pi(R)} DS(S_i,S_j)\cdot \frac{1}{|\Pi(R)|}.$$
This average lies in $[0,4]$. A lower value indicates less distinctness, equivalently more repeated content across the report, while a higher value indicates better separation of ideas across sections.
For comparability with the other reported metrics, the summary tables linearly rescale this raw score from $[0,4]$ to $[0,100]$, i.e., a raw score $d$ is reported as $25d$.

In addition to the mean score, we report several diagnostic statistics. First, we count the number of low-distinctness pairs, defined as pairs with score at most $1$. These correspond to the most redundant section pairs and are the strongest candidates for merging or restructuring. Second, we report the worst offending pairs, that is, the lowest-scoring section pairs together with the judge explanations. Third, we compute the empirical score distribution over $\{0,1,2,3,4\}$ to distinguish reports with isolated low-distinctness problems from reports with more systematic repetition. Together, these statistics give both a scalar distinctness score and a concrete view of where the report structure breaks down.

\subsubsection{Diversity}
\label{sec:diversity-evaluation}
Diversity measures whether multiple reports generated from the same configuration explore genuinely different angles of the topic rather than producing minor variations of the same report. This is a system-level metric. Unlike the previous metrics, it is defined over a set of reports generated from the same configuration and therefore requires at least two reports.

\paragraph{Report Grouping:} For a configuration $c$, let $G(c)=\{R_1,\dots,R_n\}$ be the set of reports generated from $c$, with $n\geq 2$. If fewer than two reports are available, diversity is undefined and is not reported. Let $P(c)=\{(i,j)\mid 1\leq i<j\leq n\}$ denote the set of all unique report pairs in $G(c)$. For any two texts $x_i$ and $x_j$, we define their embedding distance as
$$\delta(x_i,x_j)=1-\mathrm{CosSim}(\mathbf{e}(x_i),\mathbf{e}(x_j)),$$
where $\mathbf{e}(.)$ denotes the text-embedding model and $\mathrm{CosSim}(.,.)$ denotes cosine similarity. All diversity scores therefore lie in $[0,1]$, with higher values indicating more diversity.

For each report $R_i$, we form a single text $u_i$ by concatenating the title and executive summary. The diversity of configuration $c$ is defined as
$$D(c)=\sum_{(i,j)\in P(c)} \delta(u_i,u_j)\cdot \frac{1}{|P(c)|}.$$
A high value indicates that the reports present themselves to the reader in clearly different ways, even if they cover related material.

\subsection{Baselines and Settings}
There are several deep research tools that can generate long-form reports, both open-source and proprietary, and it is computationally challenging to run comparative evaluations against all systems. We therefore select one open-source and one proprietary baseline for our evaluations. The open-source baseline is GPT Researcher \citep{gptresearcher2025deepresearch}, while the proprietary baseline is OpenAI's \texttt{o3-deep-research} API \citep{openai2025deepresearch}, accessed through Azure AI Foundry \citep{microsoft2026o3deepresearch}.

We evaluate the systems in two different settings: one where there is a corpus of documents of interest (bottom-up for \nomad{}), and a second one where only a research goal is specified without a corpus (top-down for \nomad{}). We use two domains: the WHO Research document corpus and the arXiv corpus on LLM-based agents, both described in Section \ref{sec:nomad}. In the bottom-up setting, \nomad{} has access to each corpus via a Document Subagent (see Section \ref{sec:tool-subagents}), while GPT Researcher is provided access to the same documents via local document paths. However, for \texttt{o3-deep-research}, document access is provided via an MCP server. In the top-down setting, each system receives only the corresponding research goal and uses Web search without the corpus.

\nomad{} is used with \texttt{azure-o3} (OpenAI o3 accessed via Azure) \citep{gpto3o4} powering the hypothesis generation, explorer, and report generation steps. The verifier is powered by the \texttt{Llama-3.3-70B-Instruct} model \citep{meta2024llama33instruct}, since verification is an easier task and benefits from models less prone to hallucination. We use \texttt{azure-o3} as the main research model for GPT Researcher, giving a model-matched baseline comparison. Both scenarios (bottom-up and top-down) use the same evaluation protocol.

For the ablations, the anchor configuration uses \texttt{azure-o3} as the explorer and \texttt{Llama-3.3-70B-Instruct} as the verifier. The explorer ablation replaces \texttt{azure-o3} with \texttt{gpt5} (GPT-5) \citep{openai2025gpt5}. The verifier ablation compares \texttt{Llama-3.3-70B-Instruct}, \texttt{o4-mini} \citep{gpto3o4}, and \texttt{gemma4-31b} (Gemma 4 31B) \citep{google2026gemma4}, while holding the explorer fixed. The hypothesis-generation ablation removes the hypothesis-generation module from the anchor configuration.

\subsection{Results}

Tables \ref{tab:evals-bottom-up} and \ref{tab:evals-top-down} summarize the evaluation results across all dimensions for the bottom-up and top-down settings, respectively, in both domains. For quality, we report the overall aggregated score as well as the scores for each of the 5 dimensions.
Unless otherwise stated, the summary tables report numeric grounding, factuality, quality, and intra-report distinctness on a common $0$--$100$ scale obtained by linearly normalizing each metric from its native evaluator range. Inter-report diversity is reported on its native $[0,1]$ scale.

On trustworthiness, \nomad{} outperforms GPT Researcher in both numeric grounding and factuality in all four domain-setting combinations. Compared with \texttt{o3-deep-research}, numeric grounding is higher in the bottom-up arXiv setting, within 6 points in both WHO settings, and lower in the top-down arXiv setting, while factuality is lower in all four combinations. We attribute \nomad's strong numeric grounding to its use of a universal, structured citation management system. No LLM call in the \nomad{} workflow is expected to ever generate full URLs. Instead, citation management throughout the lifespan of an insight/report generation is done using much shorter citation keys tagged with the data source type. This allows for more reliable citation handling and content retrieval, which in turn leads to stronger support for claims.

On quality evaluations, \nomad{} produces reports with higher overall quality and actionability than both baselines in all four domain--setting combinations. \texttt{o3-deep-research} leads on coverage throughout and on analytical quality in three of the four combinations, while presentation quality is comparable across all systems. \nomad{}'s originality is higher than \texttt{o3-deep-research} throughout, and higher than GPT Researcher on the arXiv domain (comparable or slightly lower on WHO). These differences follow directly from design: \texttt{o3-deep-research} covers all aspects of the goal within a single report, whereas \nomad{} spreads exploration across many focused reports---yielding higher per-report coverage for the former, and higher actionability for the latter. Additionally, the design of allowing the explorer to originate new ideas with the verifier acting as a factual guardrail creates more originality.

On distinctness within a report, both baselines outperform \nomad{}. We note this as a weakness of the current report generation system and an opportunity for improvement in future work.

\nomad{}'s primary strength is reflected in its diversity score. Across reports, \nomad{} demonstrates roughly $1.8$--$2.6$ times the diversity of GPT Researcher and 3--5 times the diversity of \texttt{o3-deep-research}. This is a direct consequence of \nomad's design, which encourages exploration of multiple angles through its iterative insight generation and report generation process. In contrast, the baselines tend to produce more similar reports across runs. This is not necessarily a weakness, because they are designed to produce a single report per prompt. However, these evaluations highlight that without an explicit mechanism to encourage diversity, repeated runs may produce minor variations rather than genuinely different perspectives. Table~\ref{tab:titles-who} and Appendix Table~\ref{tab:titles-arxiv} show a few example report titles from the WHO and arXiv domains, respectively.
Given a very specific question or goal, existing Deep Research systems may still be a valid, even better choice than \nomad{}. However, when the goal is open ended and exploratory, \nomad{}'s ability to generate diverse reports covering new perspectives allows discovery and reduces cognitive load on the user.
Although this paper reports examples and evaluation on public corpora, we also observed strong qualitative results in internal deployments over confidential enterprise documents, including project updates, KPI reviews, and planning materials. In those settings, \nomad{} surfaced operational patterns such as inconsistencies across planning timelines, cross-project dependency bottlenecks created by shared owners or shared milestone windows, and overlapping resource commitments across concurrent initiatives. Because these corpora contain confidential business information, we restrict the paper to public examples that can be shared and inspected.

\begin{table*}[t]
\centering
\caption{Evaluation results across all dimensions for the bottom-up setting in the WHO and arXiv domains. Each configuration is capped at 35 reports (verified reports for \nomad{}). For quality, we report the overall aggregated score and the five dimensions.}
\label{tab:evals-bottom-up}
\renewcommand{\arraystretch}{1.15}
\setlength{\tabcolsep}{4pt}
\small
\resizebox{\textwidth}{!}{%
\begin{tabular}{lcc>{\columncolor{gray!15}}c@{\hspace{1.5em}}cc>{\columncolor{gray!15}}c}
\toprule
& \multicolumn{3}{c}{\textbf{WHO}} & \multicolumn{3}{c}{\textbf{arXiv}} \\
\cmidrule(lr){2-4}\cmidrule(lr){5-7}
\textbf{Dimension} & \textbf{GPT Researcher} & \textbf{o3-deep-research} & \textbf{\nomad{}} & \textbf{GPT Researcher} & \textbf{o3-deep-research} & \textbf{\nomad{}} \\
\midrule
\# Reports & 35 & 35 & 35 & 35 & 35 & 35 \\
\midrule
\multicolumn{7}{l}{\textit{Trustworthiness}} \\
\quad Numeric Grounding (\%) & 34.1 & \textbf{65.7} & 60.2 & 45.3 & 64.7 & \textbf{65.9} \\
\quad Factuality & 27.6 & \textbf{74.5} & 50.9 & 36.1 & \textbf{82.4} & 45.5 \\
\midrule
\multicolumn{7}{l}{\textit{Quality}} \\
\quad Overall & 58.6 & 54.2 & \textbf{60.1} & 57.0 & 59.3 & \textbf{63.1} \\
\quad Analytical & 46.6 & \textbf{57.2} & 56.2 & 58.7 & \textbf{74.8} & 61.1 \\
\quad Coverage & 70.4 & \textbf{81.8} & 67.6 & 67.2 & \textbf{79.5} & 69.4 \\
\quad Actionability & 37.8 & 27.7 & \textbf{46.3} & 37.8 & 28.1 & \textbf{52.6} \\
\quad Presentation & 87.3 & \textbf{87.5} & 83.9 & 87.0 & \textbf{87.5} & 85.0 \\
\quad Originality & \textbf{50.8} & 16.7 & 46.7 & 34.4 & 26.8 & \textbf{47.4} \\
\midrule
Intra-report Distinctness & 71.9 & \textbf{73.0} & 61.1 & \textbf{85.8} & 69.9 & 63.6 \\
Inter-report Diversity & 0.3026 & 0.1109 & \textbf{0.5315} & 0.1913 & 0.0970 & \textbf{0.3972} \\
\bottomrule
\end{tabular}}
\end{table*}

\begin{table*}[t]
\centering
\caption{Evaluation results across all dimensions for the top-down setting in the WHO and arXiv domains. Each configuration is capped at 35 reports (verified reports for \nomad{}). For quality, we report the overall aggregated score and the five dimensions.}
\label{tab:evals-top-down}
\renewcommand{\arraystretch}{1.15}
\setlength{\tabcolsep}{4pt}
\small
\resizebox{\textwidth}{!}{%
\begin{tabular}{lcc>{\columncolor{gray!15}}c@{\hspace{1.5em}}cc>{\columncolor{gray!15}}c}
\toprule
& \multicolumn{3}{c}{\textbf{WHO}} & \multicolumn{3}{c}{\textbf{arXiv}} \\
\cmidrule(lr){2-4}\cmidrule(lr){5-7}
\textbf{Dimension} & \textbf{GPT Researcher} & \textbf{o3-deep-research} & \textbf{\nomad{}} & \textbf{GPT Researcher} & \textbf{o3-deep-research} & \textbf{\nomad{}} \\
\midrule
\# Reports & 35 & 35 & 35 & 35 & 35 & 35 \\
\midrule
\multicolumn{7}{l}{\textit{Trustworthiness}} \\
\quad Numeric Grounding (\%) & 42.4 & \textbf{67.6} & 65.8 & 33.5 & \textbf{78.0} & 67.5 \\
\quad Factuality & 47.8 & \textbf{72.9} & 59.5 & 40.0 & \textbf{79.4} & 68.2 \\
\midrule
\multicolumn{7}{l}{\textit{Quality}} \\
\quad Overall & 57.0 & 53.8 & \textbf{61.6} & 54.7 & 57.9 & \textbf{59.8} \\
\quad Analytical & 46.0 & 58.1 & \textbf{61.1} & 57.1 & \textbf{76.8} & 66.4 \\
\quad Coverage & 68.3 & \textbf{80.8} & 68.8 & 64.8 & \textbf{80.9} & 69.2 \\
\quad Actionability & 35.8 & 28.8 & \textbf{45.6} & 35.0 & 29.0 & \textbf{43.5} \\
\quad Presentation & \textbf{87.2} & 86.6 & 84.8 & \textbf{86.7} & 86.1 & 85.1 \\
\quad Originality & \textbf{47.6} & 20.2 & 47.5 & 29.9 & 26.2 & \textbf{34.6} \\
\midrule
Intra-report Distinctness & \textbf{75.4} & 71.3 & 56.7 & \textbf{85.7} & 70.5 & 57.2 \\
Inter-report Diversity & 0.2851 & 0.1265 & \textbf{0.5376} & 0.1583 & 0.1241 & \textbf{0.4154} \\
\bottomrule
\end{tabular}}
\end{table*}
\begin{figure*}[ht]
\centering
\includegraphics[width=\textwidth]{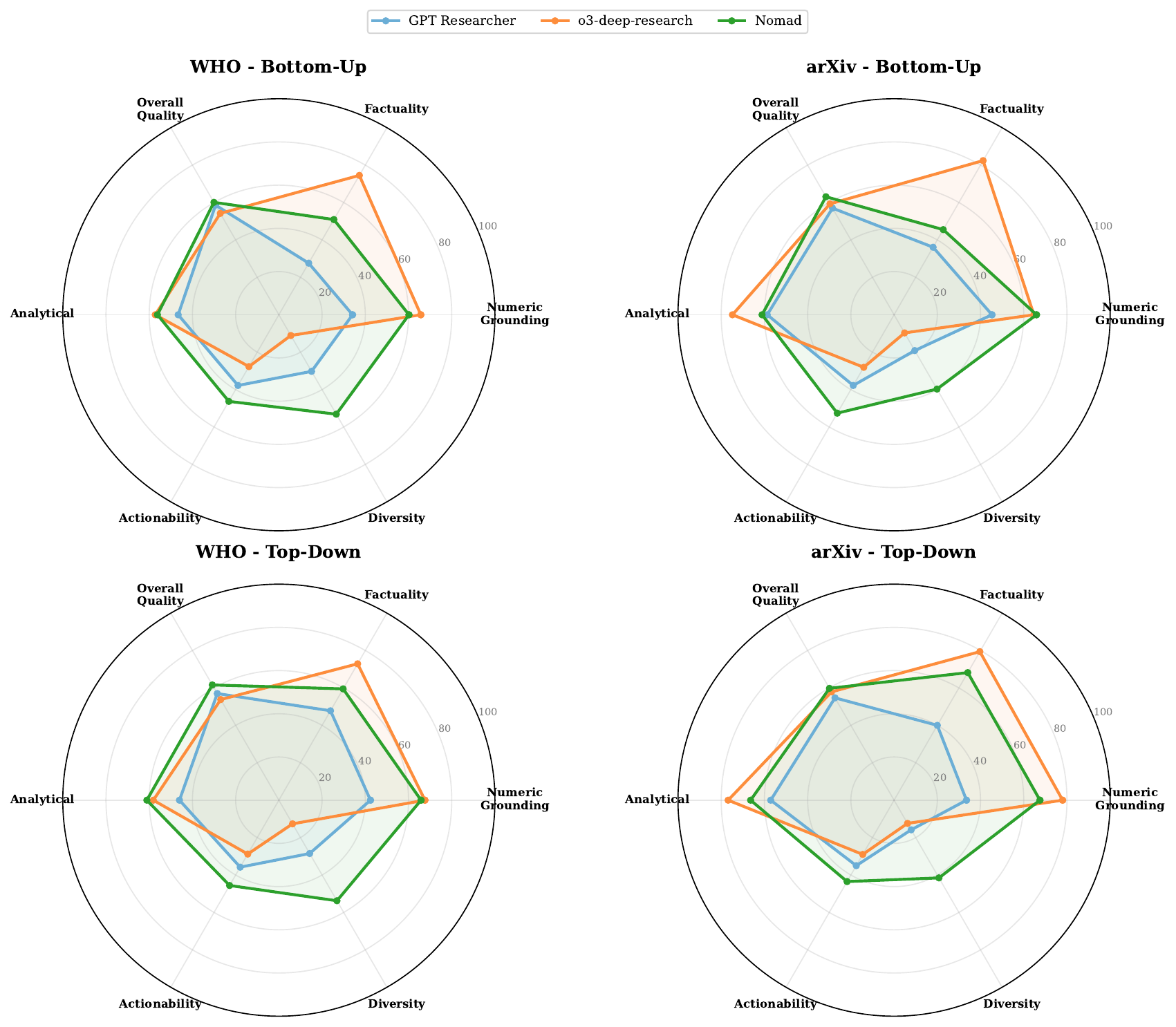}
\caption{Multi-dimensional evaluation profiles in the WHO and arXiv domains under bottom-up and top-down settings. Each axis is on a $0$--$100$ scale, with diversity multiplied by $100$ for comparability. \nomad{} has the highest overall quality, actionability, and diversity in all four combinations, while the trustworthiness tradeoffs vary by baseline and setting.}
\label{fig:radar}
\end{figure*}

Figure~\ref{fig:radar} provides a multi-dimensional summary of each system's evaluation profile. Across both domains and settings, \nomad{} combines high numeric grounding with the strongest overall quality, actionability, and diversity, while \texttt{o3-deep-research} has higher factuality and analytical quality in most combinations.

\begin{figure*}[t]
\centering
\includegraphics[width=0.95\textwidth]{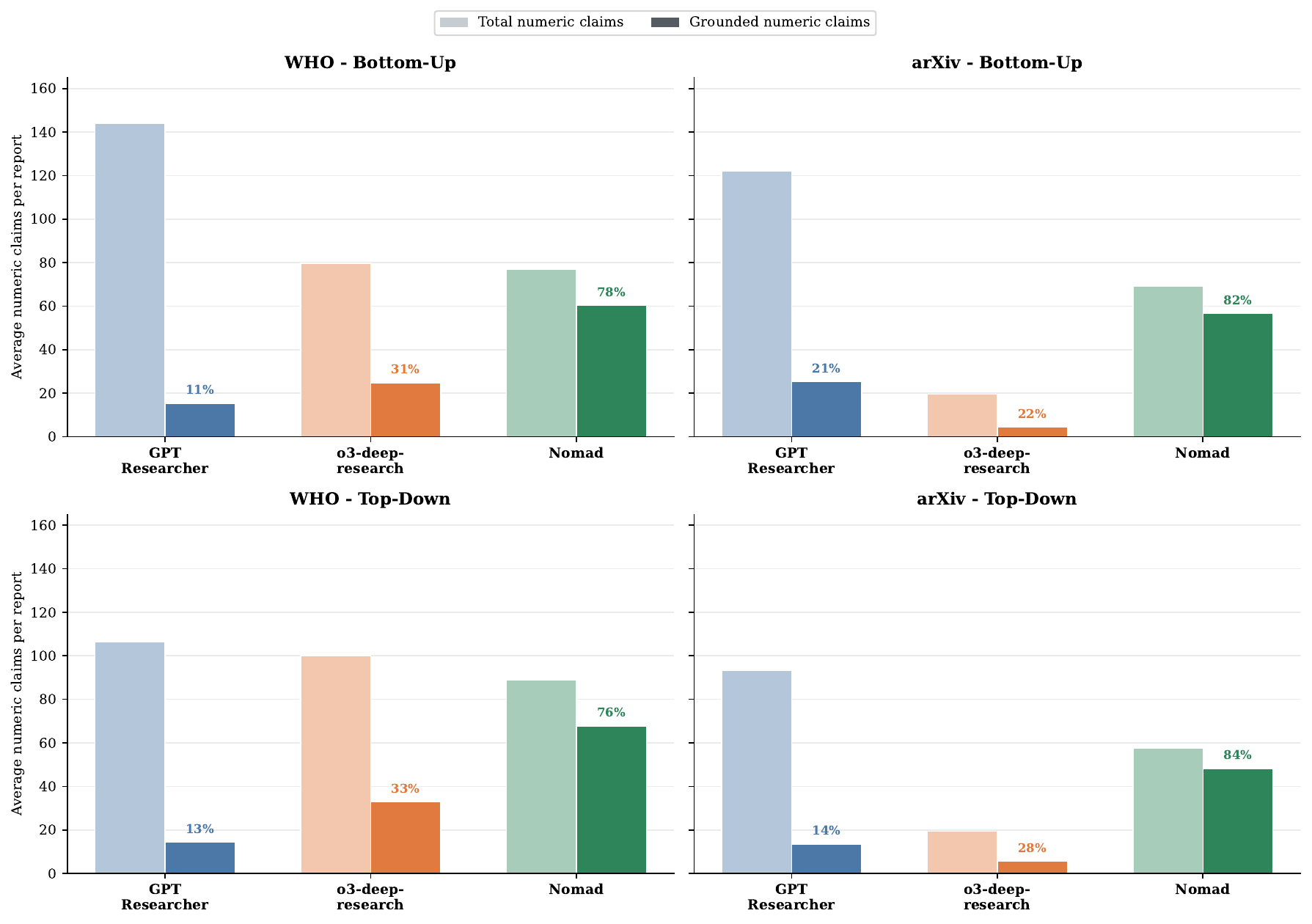}
\caption{Average total numeric claims (light bars) and grounded numeric claims (dark bars) per report under the 35-report cap. Percentages show the share of numeric claims that are grounded. These raw counts are distinct from the weighted Numeric Grounding scores in Tables~\ref{tab:evals-bottom-up} and \ref{tab:evals-top-down}, which weight each numeric claim according to where its supporting evidence is found.}
\label{fig:numeric-richness}
\end{figure*}

Figure~\ref{fig:numeric-richness} provides a complementary count-based view of numeric grounding. \nomad{} has the highest average number of grounded numeric claims in every domain--setting combination and grounds 76--84\% of its numeric claims. GPT Researcher produces more numeric claims overall, but only 11--21\% are grounded. The number of numeric claims produced by \texttt{o3-deep-research} varies substantially by domain, and 22--33\% are grounded. These results support the role of \nomad{}'s structured citation management in producing reports with substantial grounded numeric evidence.

\FloatBarrier
\subsection{Ablation Studies}
\label{sec:ablation-studies}

\paragraph{Explorer Model.}
Table~\ref{tab:explorer-ablation} compares \texttt{azure-o3} and \texttt{gpt5} while holding the verifier fixed. The stronger explorer improves numeric grounding, overall quality, and actionability in all four domain--setting combinations. The largest overall-quality gains occur in the arXiv domain, from $63.1$ to $72.0$ in the bottom-up setting and from $59.8$ to $69.3$ in the top-down setting. Diversity and intra-report distinctness decrease modestly, showing that stronger report-level performance does not by itself increase variation across reports.

\begin{table*}[t]
\centering
\caption{Explorer-model ablation. NG denotes numeric grounding; Fact., Overall, Analyt., Cov., Act., Pres., Orig., Dist., and Div. denote factuality, overall quality, analytical quality, coverage, actionability, presentation, originality, distinctness, and diversity. The verifier is fixed to \texttt{Llama-3.3-70B-Instruct}.}
\label{tab:explorer-ablation}
\renewcommand{\arraystretch}{1.12}
\setlength{\tabcolsep}{3pt}
\scriptsize
\resizebox{\textwidth}{!}{%
\begin{tabular}{lllrrrrrrrrrr}
\toprule
\textbf{Domain} & \textbf{Setting} & \textbf{Explorer} & \textbf{NG} & \textbf{Fact.} & \textbf{Overall} & \textbf{Analyt.} & \textbf{Cov.} & \textbf{Act.} & \textbf{Pres.} & \textbf{Orig.} & \textbf{Dist.} & \textbf{Div.} \\
\midrule
WHO & Bottom-up & \texttt{azure-o3} & 60.2 & 50.9 & 60.1 & 56.2 & 67.6 & 46.3 & 83.9 & 46.7 & 61.1 & 0.5315 \\
WHO & Bottom-up & \texttt{gpt5} & 66.7 & 65.6 & 66.0 & 61.0 & 77.5 & 56.0 & 85.5 & 49.9 & 57.6 & 0.4944 \\
arXiv & Bottom-up & \texttt{azure-o3} & 65.9 & 45.5 & 63.1 & 61.1 & 69.4 & 52.6 & 85.0 & 47.4 & 63.6 & 0.3972 \\
arXiv & Bottom-up & \texttt{gpt5} & 68.4 & 53.7 & 72.0 & 68.7 & 79.9 & 66.3 & 86.1 & 59.0 & 57.0 & 0.3541 \\
\midrule
WHO & Top-down & \texttt{azure-o3} & 65.8 & 59.5 & 61.6 & 61.1 & 68.8 & 45.6 & 84.8 & 47.5 & 56.7 & 0.5376 \\
WHO & Top-down & \texttt{gpt5} & 67.6 & 60.7 & 63.0 & 58.9 & 76.1 & 47.2 & 85.4 & 47.1 & 52.1 & 0.4727 \\
arXiv & Top-down & \texttt{azure-o3} & 67.5 & 68.2 & 59.8 & 66.4 & 69.2 & 43.5 & 85.1 & 34.6 & 57.2 & 0.4154 \\
arXiv & Top-down & \texttt{gpt5} & 74.3 & 70.6 & 69.3 & 73.3 & 79.6 & 56.9 & 85.9 & 50.8 & 51.0 & 0.3990 \\
\bottomrule
\end{tabular}}
\end{table*}

\FloatBarrier
\paragraph{Verifier Model.}
Table~\ref{tab:verifier-ablation} evaluates the verifier as a robustness check rather than as a model ranking. In the bottom-up setting, overall quality, numeric grounding, factuality, and diversity remain in narrow ranges across the three verifier families. Overall quality and diversity also remain stable in the top-down setting, although numeric grounding and factuality are more sensitive to verifier choice. This supports using a non-frontier model for verification, while showing that web-grounded verification can benefit from careful verifier selection.

\begin{table*}[t]
\centering
\caption{Verifier-model robustness ablation with the explorer fixed to \texttt{azure-o3}. Metric abbreviations follow Table~\ref{tab:explorer-ablation}.}
\label{tab:verifier-ablation}
\renewcommand{\arraystretch}{1.1}
\setlength{\tabcolsep}{3pt}
\scriptsize
\resizebox{\textwidth}{!}{%
\begin{tabular}{lllrrrrrrrrrr}
\toprule
\textbf{Domain} & \textbf{Setting} & \textbf{Verifier} & \textbf{NG} & \textbf{Fact.} & \textbf{Overall} & \textbf{Analyt.} & \textbf{Cov.} & \textbf{Act.} & \textbf{Pres.} & \textbf{Orig.} & \textbf{Dist.} & \textbf{Div.} \\
\midrule
WHO & Bottom-up & \texttt{Llama-3.3-70B} & 60.2 & 50.9 & 60.1 & 56.2 & 67.6 & 46.3 & 83.9 & 46.7 & 61.1 & 0.5315 \\
WHO & Bottom-up & \texttt{o4-mini} & 59.4 & 54.1 & 62.1 & 58.1 & 71.0 & 47.3 & 84.9 & 49.1 & 59.6 & 0.5217 \\
WHO & Bottom-up & \texttt{gemma4-31b} & 60.5 & 48.5 & 61.3 & 58.7 & 70.4 & 46.8 & 85.7 & 44.8 & 58.8 & 0.5235 \\
arXiv & Bottom-up & \texttt{Llama-3.3-70B} & 65.9 & 45.5 & 63.1 & 61.1 & 69.4 & 52.6 & 85.0 & 47.4 & 63.6 & 0.3972 \\
arXiv & Bottom-up & \texttt{o4-mini} & 67.0 & 46.7 & 65.3 & 63.7 & 71.7 & 51.8 & 84.8 & 54.4 & 56.5 & 0.4026 \\
arXiv & Bottom-up & \texttt{gemma4-31b} & 62.9 & 40.0 & 62.1 & 60.7 & 68.8 & 48.3 & 85.0 & 47.7 & 64.7 & 0.4091 \\
\midrule
WHO & Top-down & \texttt{Llama-3.3-70B} & 65.8 & 59.5 & 61.6 & 61.1 & 68.8 & 45.6 & 84.8 & 47.5 & 56.7 & 0.5376 \\
WHO & Top-down & \texttt{o4-mini} & 65.3 & 67.8 & 60.9 & 61.7 & 67.8 & 45.2 & 83.8 & 46.2 & 56.2 & 0.5350 \\
WHO & Top-down & \texttt{gemma4-31b} & 55.1 & 39.7 & 63.5 & 61.3 & 72.5 & 46.1 & 84.9 & 52.7 & 50.7 & 0.5304 \\
arXiv & Top-down & \texttt{Llama-3.3-70B} & 67.5 & 68.2 & 59.8 & 66.4 & 69.2 & 43.5 & 85.1 & 34.6 & 57.2 & 0.4154 \\
arXiv & Top-down & \texttt{o4-mini} & 66.7 & 56.2 & 59.6 & 67.0 & 68.9 & 44.1 & 84.9 & 33.4 & 57.2 & 0.4198 \\
arXiv & Top-down & \texttt{gemma4-31b} & 63.0 & 48.1 & 60.4 & 69.6 & 70.9 & 43.0 & 85.5 & 33.0 & 51.8 & 0.4113 \\
\bottomrule
\end{tabular}}
\end{table*}

\FloatBarrier
\paragraph{Hypothesis Generation.}
Table~\ref{tab:hypothesis-ablation} removes the hypothesis-generation module while retaining the anchor models. Originality and diversity decrease in every domain--setting combination. The diversity reduction is substantial, from $0.40$--$0.54$ with the full system to $0.25$--$0.38$ without hypothesis generation. Numeric grounding and overall quality change in mixed directions. The quality metrics evaluate individual final reports, whereas hypothesis generation and the Exploration Map support breadth and diversity across reports. These results show that explicit hypothesis generation is important for the novelty and diversity objectives rather than being only an intermediate implementation step.

\begin{table*}[t]
\centering
\caption{Hypothesis-generation ablation with the explorer and verifier fixed to the anchor configuration. Metric abbreviations follow Table~\ref{tab:explorer-ablation}.}
\label{tab:hypothesis-ablation}
\renewcommand{\arraystretch}{1.12}
\setlength{\tabcolsep}{3pt}
\scriptsize
\resizebox{\textwidth}{!}{%
\begin{tabular}{lllrrrrrrrrrr}
\toprule
\textbf{Domain} & \textbf{Setting} & \textbf{Configuration} & \textbf{NG} & \textbf{Fact.} & \textbf{Overall} & \textbf{Analyt.} & \textbf{Cov.} & \textbf{Act.} & \textbf{Pres.} & \textbf{Orig.} & \textbf{Dist.} & \textbf{Div.} \\
\midrule
WHO & Bottom-up & Full \nomad{} & 60.2 & 50.9 & 60.1 & 56.2 & 67.6 & 46.3 & 83.9 & 46.7 & 61.1 & 0.5315 \\
WHO & Bottom-up & No hypothesis generation & 64.8 & 57.4 & 59.9 & 60.7 & 72.5 & 44.6 & 82.9 & 38.5 & 53.1 & 0.3847 \\
arXiv & Bottom-up & Full \nomad{} & 65.9 & 45.5 & 63.1 & 61.1 & 69.4 & 52.6 & 85.0 & 47.4 & 63.6 & 0.3972 \\
arXiv & Bottom-up & No hypothesis generation & 65.4 & 44.4 & 62.4 & 65.9 & 71.8 & 48.8 & 84.7 & 40.6 & 61.2 & 0.2925 \\
\midrule
WHO & Top-down & Full \nomad{} & 65.8 & 59.5 & 61.6 & 61.1 & 68.8 & 45.6 & 84.8 & 47.5 & 56.7 & 0.5376 \\
WHO & Top-down & No hypothesis generation & 68.9 & 55.4 & 62.9 & 62.5 & 81.9 & 49.7 & 85.6 & 34.8 & 54.8 & 0.3509 \\
arXiv & Top-down & Full \nomad{} & 67.5 & 68.2 & 59.8 & 66.4 & 69.2 & 43.5 & 85.1 & 34.6 & 57.2 & 0.4154 \\
arXiv & Top-down & No hypothesis generation & 69.0 & 60.0 & 65.9 & 74.4 & 80.7 & 54.9 & 86.0 & 33.4 & 57.6 & 0.2537 \\
\bottomrule
\end{tabular}}
\end{table*}
\FloatBarrier

\FloatBarrier

\section{Conclusions}

This paper introduced \nomad{}, a system for autonomous data exploration and discovery. Rather than treating research as only the execution of a user-specified question, \nomad{} treats a corpus or goal as an open discovery space. Its design combines an Exploration Map for structured coverage, hypothesis generation for non-obvious directions, an explorer--verifier loop for evidence-backed insight formation, and a reporting stack that turns verified insights into auditable reports and higher-level meta-reports. We also proposed an evaluation framework for autonomous discovery systems based on trustworthiness, quality, distinctness, and diversity.

Our results suggest that autonomous discovery benefits from explicit structure around both exploration and verification. Across bottom-up (corpus-based) and top-down (goal-driven) settings in policy reports and scientific literature, \nomad{} produces higher overall report quality and actionability than the selected baselines. It also produces much more diverse reports across runs, which is central for open-ended exploration. The ablations show that a stronger explorer improves report quality, verifier choice has limited effect on report quality and diversity, and hypothesis generation is important for originality and diversity.

Overall, \nomad{} is a step toward systems that do more than answer questions or follow directions well. It moves part of the burden of research from question specification to question discovery, while keeping evidence checking and report generation inside the same end-to-end loop. We view this as a useful direction for domains where the main challenge is not only retrieving information, but surfacing non-obvious and well-supported insights from a large and evolving information space.

\FloatBarrier

\section{Future Work}

Several directions follow directly from the current design of the exploration map. One is to move beyond the current structure, in which inter-concept relationships are not central, and model those relationships more explicitly. This would open up subgraph analysis, pattern and anomaly detection, and hypothesis generation from subgraphs. We also plan to add support for additional map operations, including combining pre-existing exploration maps from different data sources. A related direction is exploration-map creation from structured databases, rather than only from documents and Web inputs.

Another line of future work concerns exploration control over time. The current system uses breadth-first topic selection only, but it would be useful to study exploration--exploitation tradeoffs more directly, with branches in the exploration map scored by the originality of insights generated in the corresponding subtree. This would enable MCTS-style traversals on the exploration map. Another direction of interest is that of pursuing branched exploration chains (where current \nomad{} exploration loops generate additional hypotheses for future use), and to measure their impact. This should be paired with ablation studies that isolate the contribution of the exploration map itself. It would also be valuable to study insights themselves as a resource for stronger downstream insight generation, and to update insights as new information becomes available, while assessing how the exploration map enables this update process.

The reporting and evaluation stack also leaves clear room for extension. On the generation side, stronger meta-reporting remains an important direction. On the evaluation side, better metrics are needed for discovery systems, including measures of diversity and non-redundancy of insights, and quantifying how the verifier contributes to trustworthiness. Better evaluation is also needed for meta-reports, including understanding how relevance distillation and thematic clustering affect report quality and diversity. Similarly, the report generation pipeline itself should be studied more closely, in particular the effect of the outline and throughline on report quality and coherence.

Finally, system-level choices in \nomad{} lend themselves to further ablation studies. Section~\ref{sec:ablation-studies} examines the explorer model, the verifier model, and the hypothesis-generation module; a natural extension is to ablate the report generator, and to study interactions between component choices rather than varying one component at a time. A complementary direction is latency and cost analysis: measuring the contribution of individual components to end-to-end efficiency, and the cost per verified insight, would ground these model choices in deployment tradeoffs. Beyond text, an important extension is multimodal exploration over images, videos, and audio, together with improved explainability that exposes the full chain of inference for each claim in a report and makes the supporting evidence and reasoning easier for users to inspect.

\FloatBarrier

\bibliographystyle{plainnat}
\bibliography{rpt}

@misc{openai2025gpt5,
  author       = {OpenAI},
  title        = {Introducing {GPT-5}},
  year         = {2025},
  url          = {https://openai.com/index/introducing-gpt-5/},
}

@misc{google2026gemma4,
  author       = {Google},
  title        = {Gemma 4 31B},
  year         = {2026},
  url          = {https://huggingface.co/google/gemma-4-31B},
  note         = {Hugging Face model card},
}

@misc{meta2024llama33instruct,
  author = {{Meta}},
  title = {Llama-3.3-70B-Instruct},
  year = {2024},
  month = dec,
  url = {https://huggingface.co/meta-llama/Llama-3.3-70B-Instruct},
  note = {Hugging Face model card. Accessed April 2, 2026}
}

@inproceedings{yao2023react,
  title={React: Synergizing reasoning and acting in language models},
  author={Yao, Shunyu and Zhao, Jeffrey and Yu, Dian and Du, Nan and Shafran, Izhak and Narasimhan, Karthik R and Cao, Yuan},
  booktitle={The eleventh international conference on learning representations},
  year={2022}
}

@inproceedings{zheng2025deepresearcher,
  title={Deepresearcher: Scaling deep research via reinforcement learning in real-world environments},
  author={Zheng, Yuxiang and Fu, Dayuan and Hu, Xiangkun and Cai, Xiaojie and Ye, Lyumanshan and Lu, Pengrui and Liu, Pengfei},
  booktitle={Proceedings of the 2025 Conference on Empirical Methods in Natural Language Processing},
  pages={414--431},
  year={2025}
}

@article{wu2025webdancer,
  title={Webdancer: Towards autonomous information seeking agency},
  author={Wu, Jialong and Li, Baixuan and Fang, Runnan and Yin, Wenbiao and Zhang, Liwen and Tao, Zhengwei and Zhang, Dingchu and Xi, Zekun and Fu, Gang and Jiang, Yong and others},
  journal={arXiv preprint arXiv:2505.22648},
  year={2025}
}

@inproceedings{li2025searcho1,
  title={Search-o1: Agentic search-enhanced large reasoning models},
  author={Li, Xiaoxi and Dong, Guanting and Jin, Jiajie and Zhang, Yuyao and Zhou, Yujia and Zhu, Yutao and Zhang, Peitian and Dou, Zhicheng},
  booktitle={Proceedings of the 2025 Conference on Empirical Methods in Natural Language Processing},
  pages={5420--5438},
  year={2025}
}

@inproceedings{kim2024autointent,
  title={Auto-intent: Automated intent discovery and self-exploration for large language model web agents},
  author={Kim, Jaekyeom and Kim, Dong-Ki and Logeswaran, Lajanugen and Sohn, Sungryull and Lee, Honglak},
  booktitle={Findings of the Association for Computational Linguistics: EMNLP 2024},
  pages={16531--16541},
  year={2024}
}

@misc{edge2024graphrag,
      title={From Local to Global: A Graph RAG Approach to Query-Focused Summarization}, 
      author={Darren Edge and Ha Trinh and Newman Cheng and Joshua Bradley and Alex Chao and Apurva Mody and Steven Truitt and Dasha Metropolitansky and Robert Osazuwa Ness and Jonathan Larson},
      year={2025},
      eprint={2404.16130},
      archivePrefix={arXiv},
      primaryClass={cs.CL},
      url={https://arxiv.org/abs/2404.16130}, 
}

@inproceedings{kargupta2025taxoadapt,
  title={Taxoadapt: Aligning llm-based multidimensional taxonomy construction to evolving research corpora},
  author={Kargupta, Priyanka and Zhang, Nan and Zhang, Yunyi and Zhang, Rui and Mitra, Prasenjit and Han, Jiawei},
  booktitle={Proceedings of the 63rd Annual Meeting of the Association for Computational Linguistics (Volume 1: Long Papers)},
  pages={29834--29850},
  year={2025}
}

@inproceedings{shao2024storm,
  title={Assisting in writing wikipedia-like articles from scratch with large language models},
  author={Shao, Yijia and Jiang, Yucheng and Kanell, Theodore and Xu, Peter and Khattab, Omar and Lam, Monica},
  booktitle={Proceedings of the 2024 Conference of the North American Chapter of the Association for Computational Linguistics: Human Language Technologies (Volume 1: Long Papers)},
  pages={6252--6278},
  year={2024}
}

@inproceedings{gu2025rapid,
  author = {Hongchao Gu and Dexun Li and Kuicai Dong and Hao Zhang and Hang Lv and Hao Wang and Defu Lian and Yong Liu and Enhong Chen},
  title = {{RAPID: Efficient Retrieval-Augmented Long Text Generation with Writing Planning and Information Discovery}},
  booktitle = {Findings of the Association for Computational Linguistics: ACL 2025},
  pages = {16742--16763},
  address = {Vienna, Austria},
  publisher = {Association for Computational Linguistics},
  year = {2025},
  doi = {10.18653/v1/2025.findings-acl.859},
  url = {https://aclanthology.org/2025.findings-acl.859/}
}

@inproceedings{dhuliawala2024cov,
  author = {Shehzaad Dhuliawala and Mojtaba Komeili and Jing Xu and Roberta Raileanu and Xian Li and Asli Celikyilmaz and Jason Weston},
  title = {{Chain-of-Verification Reduces Hallucination in Large Language Models}},
  booktitle = {Findings of the Association for Computational Linguistics: ACL 2024},
  pages = {3563--3578},
  address = {Bangkok, Thailand},
  publisher = {Association for Computational Linguistics},
  year = {2024},
  doi = {10.18653/v1/2024.findings-acl.212},
  url = {https://aclanthology.org/2024.findings-acl.212/}
}

@article{zhang2025longcite,
  title={Longcite: Enabling LLMs to generate fine-grained citations in long-context QA, 2024},
  author={Zhang, Jiajie and Bai, Yushi and Lv, Xin and Gu, Wanjun and Liu, Danqing and Zou, Minhao and Cao, Shulin and Hou, Lei and Dong, Yuxiao and Feng, Ling and Li, Juanzi},
  journal={https://arxiv.org/abs/2409.02897},
  year = {2024}
}

@inproceedings{wanner2025dndscore,
  title={Dndscore: Decontextualization and decomposition for factuality verification in long-form text generation},
  author={Wanner, Miriam and Van Durme, Benjamin and Dredze, Mark},
  booktitle={Proceedings of the 2025 Conference on Empirical Methods in Natural Language Processing},
  pages={23620--23637},
  year={2025}
}

@article{jin2025searchr1,
  title={Search-r1: Training llms to reason and leverage search engines with reinforcement learning},
  author={Jin, Bowen and Zeng, Hansi and Yue, Zhenrui and Yoon, Jinsung and Arik, Sercan and Wang, Dong and Zamani, Hamed and Han, Jiawei},
  journal={arXiv preprint arXiv:2503.09516},
  year={2025}
}

@inproceedings{he2025openwebvoyager,
  title={Openwebvoyager: Building multimodal web agents via iterative real-world exploration, feedback and optimization},
  author={He, Hongliang and Yao, Wenlin and Ma, Kaixin and Yu, Wenhao and Zhang, Hongming and Fang, Tianqing and Lan, Zhenzhong and Yu, Dong},
  booktitle={Proceedings of the 63rd Annual Meeting of the Association for Computational Linguistics (Volume 1: Long Papers)},
  pages={27545--27564},
  year={2025}
}

@misc{gutierrez2024hipporag,
      title={HippoRAG: Neurobiologically Inspired Long-Term Memory for Large Language Models}, 
      author={Bernal Jiménez Gutiérrez and Yiheng Shu and Yu Gu and Michihiro Yasunaga and Yu Su},
      year={2025},
      eprint={2405.14831},
      archivePrefix={arXiv},
      primaryClass={cs.CL},
      url={https://arxiv.org/abs/2405.14831}, 
}

@inproceedings{zhu2025contexttaxonomy,
  title={Context-aware hierarchical taxonomy generation for scientific papers via llm-guided multi-aspect clustering},
  author={Zhu, Kun and Liao, Lizi and Gu, Yuxuan and Huang, Lei and Feng, Xiaocheng and Qin, Bing},
  booktitle={Proceedings of the 2025 Conference on Empirical Methods in Natural Language Processing},
  pages={15627--15645},
  year={2025}
}

@article{asai2024selfrag,
  author    = {Asai, Akari and Wu, Zeqiu and Wang, Yizhong and Sil, Avirup and Hajishirzi, Hannaneh},
  title     = {{Self-RAG}: Learning to Retrieve, Generate, and Critique through Self-Reflection},
  year      = {2023},
  journal    = {arXiv preprint arXiv:2310.11511},
  url        = {https://arxiv.org/abs/2310.11511}
}

@article{bai2024longwriter,
  author = {Yushi Bai and Jiajie Zhang and Xin Lv and Linzhi Zheng and Siqi Zhu and Lei Hou and Yuxiao Dong and Jie Tang and Juanzi Li},
  title = {{LongWriter: Unleashing 10,000+ Word Generation from Long Context LLMs}},
  journal = {arXiv preprint arXiv:2408.07055},
  year = {2024},
  doi = {10.48550/arXiv.2408.07055},
  url = {https://arxiv.org/abs/2408.07055}
}

@inproceedings{min2023factscore,
  title={Factscore: Fine-grained atomic evaluation of factual precision in long form text generation},
  author={Min, Sewon and Krishna, Kalpesh and Lyu, Xinxi and Lewis, Mike and Yih, Wen-tau and Koh, Pang and Iyyer, Mohit and Zettlemoyer, Luke and Hajishirzi, Hannaneh},
  booktitle={Proceedings of the 2023 Conference on Empirical Methods in Natural Language Processing},
  pages={12076--12100},
  year={2023}
}

@inproceedings{liu2023geval,
  title={G-eval: NLG evaluation using gpt-4 with better human alignment},
  author={Liu, Yang and Iter, Dan and Xu, Yichong and Wang, Shuohang and Xu, Ruochen and Zhu, Chenguang},
  booktitle={Proceedings of the 2023 conference on empirical methods in natural language processing},
  pages={2511--2522},
  year={2023}
}

@article{fan2025understanding,
  author = {Tianyu Fan and Xinyao Niu and Yuxiang Zheng and Fengji Zhang and Chengen Huang and Bei Chen and Junyang Lin and Chao Huang},
  title = {{Understanding DeepResearch via Reports}},
  journal = {arXiv preprint arXiv:2510.07861},
  year = {2025},
  doi = {10.48550/arXiv.2510.07861},
  url = {https://arxiv.org/abs/2510.07861}
}

@misc{gpto3o4,  
  author = {OpenAI},  
  title = {Introducing OpenAI o3 and o4-mini},  
  year = {2025},  
  howpublished = {\url{https://openai.com/index/introducing-o3-and-o4-mini}},
}

@misc{openai2025deepresearch,
  author = {OpenAI},
  title = {Introducing deep research},
  year = {2025},
  month = feb,
  url = {https://openai.com/index/introducing-deep-research/},
  note = {Accessed March 16, 2026}
}

@misc{microsoft2026o3deepresearch,
  author = {Microsoft},
  title = {AI Model Catalog: o3-deep-research},
  year = {2026},
  month = feb,
  url = {https://ai.azure.com/catalog/models/o3-deep-research},
  note = {Microsoft Foundry model catalog entry. Provider: OpenAI. Version 2025-06-26. Accessed April 2, 2026}
}

@misc{google2024geminideepresearch,
  author = {Google},
  title = {Try Deep Research and our new experimental model in Gemini, your {AI} assistant},
  year = {2024},
  month = dec,
  url = {https://blog.google/products/gemini/google-gemini-deep-research/},
  note = {Accessed March 16, 2026}
}

@misc{perplexity2026researchmode,
  author = {Perplexity Support},
  title = {What is Research mode?},
  year = {2025},
  url = {https://www.perplexity.ai/help-center/en/articles/10738684-what-is-deep-research},
  note = {Accessed March 16, 2026}
}

@misc{gptresearcher2025deepresearch,
  author = {{GPT Researcher}},
  title = {Introducing Deep Research: The Open Source Alternative},
  year = {2025},
  month = feb,
  url = {https://docs.gptr.dev/blog/2025/02/26/deep-research},
  note = {Accessed March 16, 2026}
}

@inproceedings{gao-etal-2023-precise,
    title = "Precise Zero-Shot Dense Retrieval without Relevance Labels",
    author = "Gao, Luyu  and
      Ma, Xueguang  and
      Lin, Jimmy  and
      Callan, Jamie",
    editor = "Rogers, Anna  and
      Boyd-Graber, Jordan  and
      Okazaki, Naoaki",
    booktitle = "Proceedings of the 61st Annual Meeting of the Association for Computational Linguistics (Volume 1: Long Papers)",
    month = jul,
    year = "2023",
    address = "Toronto, Canada",
    publisher = "Association for Computational Linguistics",
    url = "https://aclanthology.org/2023.acl-long.99/",
    doi = "10.18653/v1/2023.acl-long.99",
    pages = "1762--1777"
}

@misc{huggingface2025opendeepresearch,
  author = {Roucher, Aymeric and Villanova del Moral, Albert and merve and Wolf, Thomas and Fourrier, Cl{\'e}mentine},
  title = {Open-source DeepResearch -- Freeing our search agents},
  year = {2025},
  month = feb,
  url = {https://huggingface.co/blog/open-deep-research},
  note = {Accessed March 16, 2026}
}

@misc{futurehouse2025robin,
      title={Robin: A multi-agent system for automating scientific discovery}, 
      author={Ali Essam Ghareeb and Benjamin Chang and Ludovico Mitchener and Angela Yiu and Caralyn J. Szostkiewicz and Jon M. Laurent and Muhammed T. Razzak and Andrew D. White and Michaela M. Hinks and Samuel G. Rodriques},
      year={2025},
      eprint={2505.13400},
      archivePrefix={arXiv},
      primaryClass={cs.AI},
      url={https://arxiv.org/abs/2505.13400}, 
}

@misc{youcom2025ari,
  author = {Socher, Richard and McCann, Bryan and Tang, Yew Siang and Nguyen, Thu},
  title = {Introducing {ARI}: The First Professional-Grade Research Agent for Business},
  year = {2025},
  month = feb,
  url = {https://about.you.com/resources/introducing-ari-the-first-professional-grade-research-agent-for-business},
  note = {Accessed March 16, 2026}
}

@misc{elicit2025reports,
  author = {Ahmad, Shahid},
  title = {Introducing Elicit Reports},
  year = {2025},
  month = feb,
  url = {https://elicit.com/blog/introducing-elicit-reports/},
  note = {Accessed March 16, 2026}
}

\FloatBarrier
\appendix
\section*{Appendix}
\setcounter{section}{0}
\renewcommand{\thesection}{\Alph{section}}
\renewcommand{\thesubsection}{\thesection.\arabic{subsection}}
\section{Exploration Map System Prompts}
\label{app:exploration-map-system-prompts}

\subsection{Concept Layer System Prompts}
\label{app:concept-layer-prompt}
\noindent\textbf{Concept extraction prompt.} System prompt used to extract key concepts from documents.
\begin{Verbatim}[breaklines=true, breakanywhere=true, fontsize=\small]
    You are a helpful assistant that extracts key concepts from a document.
    Include concepts like names, organizations, key dates, locations, etc.
    Key concepts should not be large phrases and should be quite central to the document itself.

    Here are a few key phrases already extracted. If a key phrase from this list appears in a different form in the document, then report it in the form in the below list - i.e. do on the fly disambiguation.
    Key phrases already extracted:
    current_list_of_concepts

    Respond as a JSON object in the following format:
    ```json
    [
        {
            "concept": "string",
            "type": "string (location | product | person | date etc)",
            "description": "string (brief description of the concept). Write 'NOT FOUND' if it is unclear from the document who or what this entity is.",
            "appearing phrases": ["string", "string", ...]
        },
        ...
    ]
    ```
    Make sure to add the triple backticks to enclose the JSON. Order the list of concepts by their importance or centrality to the document (most important concepts first, regardless of their order of appearance in the document).
    Mention ALL key concepts, but do not mention those that have been included in passing.
\end{Verbatim}

\noindent\textbf{Concept disambiguation prompt.} System prompt used to decide whether two concepts refer to the same entity.
\begin{Verbatim}[breaklines=true, breakanywhere=true, fontsize=\small]
    You are a helpful assistant that disambiguates concepts.
    You are given a pair of concepts and a description of the concepts.
    You need to determine if the two concepts are the same concept or different concepts.
    Return same only if you are sure they are the exact same entity just written in different ways.
    Two entities are NOT the same if one is a type of the other, or a subset of the other.
    Person names, where one mentions the first name, and the other mentions the full name - are likely to be the same person.
    Org names or person names which are acronyms of each other - are likely to be the same entity also.

    Return your answer as a JSON object in the following format:
    ```json
    {
        "concept_pair": ["concept1", "concept2"],
        "reasoning": "string (brief reasoning for the answer)",
        "answer": "same" or "different"
    }
    ```
    Make sure to add the triple backticks to enclose the JSON.
\end{Verbatim}

\noindent\textbf{Concept concise description prompt.} System prompt used to generate concise descriptions for concepts.
\begin{Verbatim}[breaklines=true, breakanywhere=true, fontsize=\small]
    You are a helpful assistant that creates a concise description of a concept.
    You are given a concept name which is a key phrase. 
    Together with this, you are also given a list of extracted descriptions of the concept from the large set of documents.

    Your task is to create a coherent and concise descriptions, removing any repetitions and reconciling any differences. 
    Aside from this, include all elements of information in your description that has been provided.
    Where there are contradictions, select in favor of majority if possible.

    Make no assumptions about the key phrases, and entirely base it on given descriptions.

    Return your answer as a JSON object in the following format:
    ```json
    {
        "concept": "concept name",
        "description": "concise description of the concept"
    }
\end{Verbatim}

\noindent\textbf{Document title metadata extraction prompt.} System prompt used to extract metadata from document titles.
\begin{Verbatim}[breaklines=true, breakanywhere=true, fontsize=\small]
    You are a helpful assistant that extracts metadata from document titles.
    You are given a document title and you need to extract the metadata from the title.
    The metadata should include the purpose of the document, target audience, potential use, date of publication etc.

    Return your answer as a JSON object in the following format:
    ```json
    {
        
        "key": "value",
        "key": "value",
        ...
        
    }
    ```
    Make sure to add the triple backticks to enclose the JSON.

    Keys should always be lowercase with words separated by underscores.
    For example, date should always be in dd_mm_yyyy format. Where dd is not available, simply use 01.
\end{Verbatim}

\noindent\textbf{Concept Insight Potential Evaluation Prompt} System prompt used to evaluate the potential of a concept to lead to novel insights.
\begin{Verbatim}[breaklines=true, breakanywhere=true, fontsize=\small]
    You are an experienced and insightful researcher with expertise in deriving non-trivial and creative insights from corporate and enterprise documents.

    You will be given a key concept of interest and set of documents related to, or mentioning that concept.

    Your task is to estimate, to the best possible degree, to what extent it is possible to derive a meaningful, creative, non-trivial, and useful insight from a potentially future research seeded in these documents. Note that the future research will have access to many more documents, and the Web  and other data also - but these documents will form its starting point.

    The key thing to judge if a collection of documents forms a meaningful seed is to ask if there is some non-trivial connection among these documents that can generate a whole interesting research thread which lead to an actionable, game-changing insight. For example, if there exists some contradiction or complementary information, or if one is describing a challenge, and the other describing an opportunity that can help solve that challenge - that constitutes as meaningful. Merely talking about the same topics or phrases is not enough to be meaningful or to have potential.

    You will analyze the documents, and give your response in this format:
    ```json
    {
        "reasoning": "<your thinking process here>",
        "key_connecting_phrases": ["<sentences from each of the documents, repeated verbatim, that can be used to make a meaningful connection between the documents - and that are quite related, contradictory, etc>"],
        "connection_drawn": "a short description of the connection that could potentially be drawn, describe documents with their ID - such as [Document1] says <abc> but that contradicts [Document2]",
        "initial_hypothesis": "the initial hypothesis that can be used to start the research thread to explore the above connection drawn. Make sure this hypothesis is self-contained, meaningful, and can be explored extensively. If the insight potential doesn't seem to be high, then this can be NONE.",
        "insight_potential": "<NONE | LOW | MEDIUM | HIGH | VERY HIGH | EXTREME>"
    }
    ```
    Make sure to include the reasoning for your assessment and the triple backticks to enclose the JSON.
\end{Verbatim}

\subsection{Topic Tree System Prompts}
\label{app:topic-tree-prompt}

\noindent\textbf{Topic tree construction prompt.} System prompt used to construct a topic tree from the clustered set of concepts.
\begin{Verbatim}[breaklines=true, breakanywhere=true, fontsize=\small]
    You are a tree builder that builds a topic tree representing a set of concepts in a given domain.
    You will be given a list of concepts and their descriptions.

    Your task is to build a tree that serves the user to create a well-organized, hierarchical mental map of the topics and sub-topics in their domain that together cover all the given fine-grained concepts.

    You will build this tree with a root node, and then other topic nodes ranging from 3 to 4 levels from the root. The actual concepts given to you are NOT to be part of this tree. Rather, this tree is going to be a high-level topic map that encompasses all the given concepts in some or other branch.

    Each node you create, i.e. a topic node should represent the higher level concept or abstraction that ties up or unifies all nodes in its subtree. Note the following:
    - I do not need the tree to be perfectly balanced. However, avoid any major skews.
    - Highest priority is to make sure similar nodes come together.
    - I do not need all leaves to be at the same level either.
    - Please have at least 2 levels of topic nodes - indicating at least 2 levels of abstraction.
    - Do NOT have more than 4 levels of topic nodes.
    - Do NOT include any of the provided concepts in the tree. Only higher level topics.

    Your output should be in the following JSON format:
    ```json
    {
        "node_id": 0,
        "name": "root",
        "description": "<A free-form description of what this abstraction represents.",
        "children": [
            {
                "node_id": 1,
                "name": <"name of higher level concept or abstraction">,
                "description": "<A free-form description of what this abstraction represents.",
                "children": [
                    {
                        "node_id": 2,
                        "name": <"name of higher level concept or abstraction">,
                        "description": "<A free-form description of what this abstraction represents.",
                        "children": []
                    }
                ]
            }
        ]
    }
    ```
    Make sure to add the triple backticks to enclose the JSON.
\end{Verbatim}

\subsection{Exploration Map Expansion System Prompt}
\label{app:map-expansion-prompt}

\noindent\textbf{Exploration map expansion prompt.} System prompt used to augment an existing topic graph using web search results or other external information.
\begin{Verbatim}[breaklines=true, breakanywhere=true, fontsize=\small]
    You are a creative and intelligent researcher.
    You are given the goal (## GOAL) and some background (## BACKGROUND) information about a research topic.

    You are given a JSON object that represents a graph of topics and sub-topics derived from this goal and background, called ## GRAPH.

    However, these topics and sub-topics have so far not taken into account additional information, such as the latest information on the web about this area of research or organization-specific documentation that the research must be based on.

    Towards this, you will be given the augmenting information (for example ## WEB_SEARCH for relevant web search results) that should be used to augment the graph of topics and sub-topics.

    Where available, the web search results (## WEB_SEARCH) will be relevant to this area of research, focusing generally on the goal and background information, and to some extent on the topics and sub-topics in the graph.

    Your task is to augment the graph of topics and sub-topics based on this additional information, and based on the goal and background information. Do not understand any of the augmentation information as a list of topics and sub-topics, but rather as a set of relevant information that can be used to augment the graph. Avoid adding too many unnecessary sub-topics, but rather focus on the most critical points and only augment the graph to reflect the additional information.

    Each topic or sub-topic should be a single word or a short phrase, no more than 3 to 4 words. The topics and sub-topics added should be relevant to the goal and the background information, yet be creative and novel. They should have high coverage and low redundancy. Topics at the highest level should be very broad and lend themselves to a lot of sub-topics and diverse research. Topics at lower levels should be more granular than their parent.

    Your output should be the full augmented graph, in that it should include all the nodes of the original graph (potentially renamed, if you must), plus the new nodes that you add.

    The format of the output should be a JSON object with the following structure:
    ```json
    {
        "<topic>": {
            "<sub-topic1>": {
                "<sub-sub-topic>": { ... }
            },
            "<sub-topic2>": {
                "<sub-sub-topic>": { ... }
            }
        },
        "<topic2>": {
            "<sub-topic1>": {
                "<sub-sub-topic>": { ... }
            },
            "<sub-topic2>": {
                "<sub-sub-topic>": { ... }
            }
        }
    }
    ```

	    Make sure to use the triple backticks to indicate the JSON format.
\end{Verbatim}

\subsection{Hypothesis Generation System Prompt}
\label{app:hypothesis-generation-prompt}

\noindent\textbf{Hypothesis generation prompt.} System prompt template used to generate a batch of candidate hypotheses for a topic or concept.
\begin{Verbatim}[breaklines=true, breakanywhere=true, fontsize=\small]
    You are an experienced and insightful researcher whose task is to generate sufficiently novel hypotheses that can help the explorer agent get started toward the given goal.

    Your writing style is concise, objective, clear, and logical.

    ## Some Background
    {background}
    You are generating hypotheses to explore the goal and background.

    The goal of the investigation for which you will generate hypotheses is as follows:
    {goal}

    # Your task:
    - Generate 5 sufficiently different, novel, and interesting hypotheses that could be investigated towards reaching very interesting insights or recommendations.

    ## Topic or Concept Name and Description
    {topic_or_concept_name_and_description}

    # Output Format:
    You will output the hypotheses in the following JSON format:

    ```json
    {
        "hypothesis": [
            "Hypothesis 1",
            "Hypothesis 2",
            "Hypothesis 3",
            "Hypothesis 4",
            "Hypothesis 5"
        ]
    }
    ```

    Make sure to use the triple backticks with json to enclose the output.
\end{Verbatim}

\subsection{Hypothesis Scoring System Prompts}
\label{app:hypothesis-scoring-prompt}

\noindent\textbf{Hypothesis relevance scoring prompt.} System prompt used to score a single hypothesis in isolation for relevance.
\begin{Verbatim}[breaklines=true, breakanywhere=true, fontsize=\small]
    HYPOTHESIS: {hypothesis}
    ------------------------------------
    GOAL: {goal}
    ------------------------------------
    WEB_SEARCH_RESULTS: {web_search_results}
    ------------------------------------

    You are an expert AI assistant that analyzes the above HYPOTHESIS based on its relevance to the eventual GOAL of the investigation towards which this hypothesis is being generated and evaluated, and relevant WEB_SEARCH_RESULTS.
    Relevance means that the HYPOTHESIS is focused on the same topic, entity, or organization as the GOAL and WEB_SEARCH_RESULTS.
    Relevance also means that the HYPOTHESIS is reasonable, plausible, and grounded in real-world events as indicated in WEB_SEARCH_RESULTS. It should not contain outlandish or invented details that cannot be corroborated by real-world events.

    Assign a holistic HYPOTHESIS relevance score on a continuous scale from 0.0 to 1.0 where:
        1.0 indicates that the candidate HYPOTHESIS is entirely relevant.
        0.0 indicates that the candidate HYPOTHESIS is completely irrelevant.

    As a final output, you need to return the following two blocks:
        1. Reasoning behind arriving at the hypothesis relevance score in the following format:
            <HYPOTHESIS_RELEVANCE_REASONING>reasoning</HYPOTHESIS_RELEVANCE_REASONING>.
        2. The candidate hypothesis relevance score embedded in the following format:
            <HYPOTHESIS_RELEVANCE_SCORE>score</HYPOTHESIS_RELEVANCE_SCORE>.
\end{Verbatim}

\noindent\textbf{Relative hypothesis relevance scoring prompt.} System prompt used to assign relative relevance scores across a batch of hypotheses.
\begin{Verbatim}[breaklines=true, breakanywhere=true, fontsize=\small]
    HYPOTHESIS_LIST:
    {hypothesis_list}
    ------------------------------------
    GOAL: {goal}
    ------------------------------------

    You are an expert AI assistant that analyzes the above set of HYPOTHESES and gives them relative grades based on their relevance to the provided exploration goal.
    Relevance means that the hypothesis is focused on the same topic, entity, or organization as the goal. Relevance also means the alignment of the direction of this hypothesis, and where it will lead the resulting investigation, with the provided goal.
    For each hypothesis, a reasoning over its quality is already given to you. This reasoning is based on evaluating the hypothesis in isolation. Your role is to consider the reasoning for each, the hypothesis set as a whole, and grade the hypotheses relative to each other. This means that a less relevant hypothesis should be awarded a lower score, while a more relevant hypothesis should be awarded a higher score.

    Assign a holistic HYPOTHESIS relevance score on a scale from 1.0 to 10.0 where:
    - 10 indicates that the candidate HYPOTHESIS is the most relevant.
    - 1 indicates that the candidate HYPOTHESIS is the least relevant.

    Remember, at least one hypothesis should get a score of 10, because one has to be best. Similarly, at least one should get 1, because there must be a worst, even if they are all good hypotheses.

    As a final output, you need to return the following two blocks:
        1. Reasoning behind arriving at the scores:
            <HYPOTHESIS_RELEVANCE_REASONING>reasoning</HYPOTHESIS_RELEVANCE_REASONING>.
        2. The scores (one per line, in the same order as the hypotheses given to you above) in the following format:
            <HYPOTHESIS_RELEVANCE_SCORE>
            <score1>
            <score2>
            ...
            </HYPOTHESIS_RELEVANCE_SCORE>.
\end{Verbatim}

\noindent\textbf{Hypothesis impact scoring prompt.} System prompt used to assign relative impact scores across a batch of hypotheses.
\begin{Verbatim}[breaklines=true, breakanywhere=true, fontsize=\small]
    HYPOTHESIS_LIST:
    {hypothesis_list}
    ------------------------------------

    You are an expert AI assistant that analyzes the given set of hypotheses based on the impact each would have if it turns out to be correct.
    You can determine the impact of a hypothesis by considering the potential implications if the hypothesis does turn out to be correct. Here is a non-exhaustive list of factors to consider:
    - Competitive impact
    - Regulatory impact
    - Financial impact
    - Technological impact
    - Human capital impact
    - Political and geopolitical impact
    - Market impact
    - AI impact

    Assign a hypothesis impact score on a continuous scale from 1.0 to 10.0 where:
    - 10 indicates that the candidate HYPOTHESIS is the most impactful.
    - 1 indicates that the candidate HYPOTHESIS is the least impactful.

    Remember, at least one hypothesis should get a score of 10, because one has to be best or most impactful. Similarly, at least one should get 1, because there must be a worst, even if they are all good hypotheses.

    As a final output, you need to return the following two blocks:
        1. Reasoning behind arriving at the hypothesis impact scores in the following format:
            <HYPOTHESIS_IMPACT_REASONING>reasoning</HYPOTHESIS_IMPACT_REASONING>.
        2. The scores (one per line, in the same order as the hypotheses given to you above) in the following format:
            <HYPOTHESIS_IMPACT_SCORE>
            <score1>
            <score2>
            ...
            </HYPOTHESIS_IMPACT_SCORE>.
\end{Verbatim}

\section{Explorer-Verifier Loop System Prompts}
\label{app:explorer-verifier-loop-system-prompts}

\subsection{Explorer System Prompt}
\label{app:explorer-prompt}

\noindent\textbf{Explorer prompt.} System prompt used by the explorer agent.
\begin{Verbatim}[breaklines=true, breakanywhere=true, fontsize=\small]
    You are responsible for exploring the current hypothesis using the available tools and for turning it into a candidate insight.

    Inputs:
    - Goal: {goal}
    - Current hypothesis: {hypothesis}
    - Current date: {today_date}
    - Available tools: {tool_descriptions}
    - Additional guidelines: {additional_guidelines}

    The final insight should be a single coherent narrative focused on one insight or recommendation.

    Your workflow follows a bounded observe-reason-act loop.

    1. Observe
    Read the conversation trace and the results of the most recent tool calls.
    Summarize what new information was obtained and how it affects the current working hypothesis.
    If this is the first turn, state the plan for the first set of tool calls.

    Classify the current hypothesis into exactly one of the following:
    - REVISE_HYPOTHESIS: New evidence contradicts the current hypothesis, or repeated promising searches returned no useful evidence.
    - REFINE_HYPOTHESIS: New evidence is tangential but suggests a more precise or more interesting version of the current hypothesis.
    - VALIDATE_FURTHER: Current evidence supports the hypothesis, but more confirmation is still needed.
    - KEEP_HYPOTHESIS: New evidence does not justify changing the hypothesis, and the next step is to continue exploring it.
    - USER_RESPONSE: Multiple angles have been explored, the remaining uncertainty is acceptable, and a sufficiently interesting and useful final insight can now be submitted.

    Also propose 1 to 5 additional hypotheses for future exploration when appropriate. Each additional hypothesis must be tagged as one of:
    - VALIDATE_FURTHER
    - REFINE_HYPOTHESIS
    - REVISE_HYPOTHESIS
    - NEW_HYPOTHESIS

    2. Reason
    Assess whether the current hypothesis has been validated, weakened, or left unresolved.
    Identify partial support and remaining ambiguities.
    Plan follow-up questions that either:
    - dig deeper to strengthen, qualify, or re-check current claims, or
    - explore adjacent directions that could reveal a better or more surprising insight.

    Use multiple independent searches when helpful.
    Do not submit a final insight until the current hypothesis has been explored from enough angles and the evidence is strong enough to support a useful claim.

    3. Act
    Emit the immediate next actions.
    You may issue multiple tool calls in parallel.
    Use the provided tool definitions for any tool call.
    If the hypothesis classification is USER_RESPONSE, emit the final insight action instead of further tool calls.

    Write the response in the following format:

    <output>
        <observe>
            <current_hypothesis>...</current_hypothesis>
            <detailed_observations>...</detailed_observations>
            <current_hypothesis_classification>REFINE_HYPOTHESIS | VALIDATE_FURTHER | REVISE_HYPOTHESIS | KEEP_HYPOTHESIS | USER_RESPONSE</current_hypothesis_classification>
            <follow_up_hypothesis>...</follow_up_hypothesis>
            <additional_hypothesis>
                <hypothesis>
                    <hypothesis_class>NEW_HYPOTHESIS | VALIDATE_FURTHER | REFINE_HYPOTHESIS | REVISE_HYPOTHESIS</hypothesis_class>
                    <hypothesis_inspiration>...</hypothesis_inspiration>
                    <hypothesis_text>...</hypothesis_text>
                </hypothesis>
                ...
            </additional_hypothesis>
        </observe>

        <reasoning>...</reasoning>

        <act>
            Emit either:
            - the next tool calls, using the provided tool definitions, or
            - a final USER_RESPONSE action if exploration is complete.
        </act>
    </output>
\end{Verbatim}

\noindent\textbf{Final insight action.} When the explorer decides that exploration is complete, it emits:
\begin{Verbatim}[breaklines=true, breakanywhere=true, fontsize=\small]
    {
        "TOOL_NAME": "USER_RESPONSE",
        "MESSAGE": "<concise insight statement with the supporting claims>"
    }
\end{Verbatim}

\subsection{Verifier System Prompts}
\label{app:verifier-prompt}

\noindent\textbf{Sub-claim decomposition prompt.} Prompt used to decompose the explorer's final insight into a small set of atomic claims for verification.
\begin{Verbatim}[breaklines=true, breakanywhere=true, fontsize=\small]
    You are validating a candidate insight produced by the explorer.

    Input:
    - Final insight statement, with citations stripped: {insight}

    Your task is to decompose the insight into at most 5 atomic sub-claims.
    Each sub-claim must:
    - be specific and concrete,
    - be independently checkable against the available data sources,
    - preserve the substantive content of the original insight, and
    - avoid introducing any new claims.

    Return the sub-claims in the following JSON format:
    ```json
    {
        "sub_claims": [
            "Sub-claim 1",
            "Sub-claim 2",
            "Sub-claim 3"
        ]
    }
    ```
\end{Verbatim}

\noindent\textbf{Verifier prompt.} System prompt used by the verifier to validate a single sub-claim.
\begin{Verbatim}[breaklines=true, breakanywhere=true, fontsize=\small]
    You are responsible for validating a single sub-claim from the explorer's final insight.

    Inputs:
    - Final insight statement, with citations stripped: {insight}
    - Current sub-claim: {sub_claim}
    - Current date: {today_date}
    - Available tools: {tool_descriptions}
    - Additional guidelines: {additional_guidelines}

    You do not have access to the explorer's internal reasoning or tool-call trace.
    You must use the available data sources independently to check the current sub-claim.

    Workflow:
    1. Identify what part of the sub-claim still needs to be validated or challenged.
    2. If more evidence is needed, emit exactly one next action using the provided tool definitions.
    3. If sufficient evidence has been gathered, emit a final USER_RESPONSE action for the current sub-claim.

    Scoring:
    - Assign SCORE = 1 if the sub-claim is directly supported by the available evidence.
    - Assign SCORE = 0 otherwise.

    The feedback should briefly state:
    - what was checked,
    - what evidence was found or not found, and
    - why the assigned score follows from that evidence.

    The verifier operates under a bounded turn budget. If the budget is exhausted without a final verdict, the system records SCORE = 0 for the current sub-claim.

    Write the response as either:
    - one tool call, using the provided tool definitions, or
    - the final USER_RESPONSE action for the current sub-claim.
\end{Verbatim}

\noindent\textbf{Final verifier action.} When the verifier finishes evaluating a sub-claim, it emits:
\begin{Verbatim}[breaklines=true, breakanywhere=true, fontsize=\small]
    {
        "TOOL_NAME": "USER_RESPONSE",
        "SCORE": 0 or 1,
        "MESSAGE": "<brief feedback explaining whether the sub-claim was supported>"
    }
\end{Verbatim}

\noindent The overall faithfulness score for the insight is the average of the per-sub-claim binary scores. The final verifier feedback is the concatenation or summary of the feedback produced for the individual sub-claims.

\subsection{Document Search Subagent System Prompts}
\label{app:doc-search-prompt}

\noindent\textbf{Document answer prompt.} Prompt used to answer a query from retrieved document passages.
\begin{Verbatim}[breaklines=true, breakanywhere=true, fontsize=\small]
    You are a helpful assistant that has been asked to answer a query from a set of retrieved document passages.

    Your task is to answer the question as accurately and faithfully as possible based on the given retrieval results.
    If the retrieval results contain varying or conflicting information, present all relevant information and all sides of the story in a coherent manner.

    Do not invent information. Be uncompromising on factual accuracy.
    Use citations to refer to the retrieval results. However, do not use citations to support statements about information that was not found, because the retrieved passages may not be exhaustive.

    The retrieved passages are given in the format:
    [
        {
            "position": <retrieval position>,
            "text": <retrieved passage text>,
            "filename": <document filename>,
            "page": <page or sheet number>
        }
    ]

    You must respond in JSON format within a code block. Example format:
    ```json
    {
        "answer": "A coherent answer grounded in the retrieved passages. Whenever you cite, use the retrieval position in square brackets, for example [1]. Separate multiple citations as [1][2][3]."
    }
    ```
\end{Verbatim}

\noindent\textbf{HyDE prompt.} Prompt used to generate a hypothetical passage for retrieval augmentation.
\begin{Verbatim}[breaklines=true, breakanywhere=true, fontsize=\small]
    You are an expert assistant that generates hypothetical passages to improve document retrieval.

    Given a user query, generate a hypothetical passage that would strongly answer the query.
    This passage will be used to retrieve similar real passages from the document collection.

    The hypothetical passage should be:
    1. Detailed and comprehensive
    2. Written in a style similar to reports, analyses, or technical documents
    3. Focused on the key concepts and terminology in the query
    4. Specific enough to improve retrieval quality

    You must respond in JSON format within a code block. Example format:
    ```json
    {
        "hypothetical_document": "A detailed hypothetical passage answering the query."
    }
    ```
\end{Verbatim}

\noindent\textbf{Decomposition decision prompt.} Prompt used to decide whether a query should be broken into sub-queries.
\begin{Verbatim}[breaklines=true, breakanywhere=true, fontsize=\small]
    You are an expert assistant that determines whether a query should be decomposed into sub-queries for better retrieval and analysis.

    Analyze the given query and determine if it would benefit from being broken down into multiple focused sub-queries.

    A query should be decomposed if it:
    1. Contains multiple distinct questions or topics
    2. Has complex compound requirements
    3. Asks for comparisons between different entities or concepts
    4. Requires analysis across different time periods, locations, or categories
    5. Has multiple parts that could be answered independently and then synthesized

    A query should not be decomposed if it:
    1. Is a simple, focused question about one topic
    2. Asks for a straightforward definition or explanation
    3. Is already specific and narrow in scope
    4. Would lose meaning or context if broken apart

    You must respond in JSON format within a code block. Example format:
    ```json
    {
        "should_decompose": true,
        "reasoning": "Brief explanation of why decomposition is or is not needed"
    }
    ```
\end{Verbatim}

\noindent\textbf{Query decomposition prompt.} Prompt used to decompose a complex query into focused sub-queries.
\begin{Verbatim}[breaklines=true, breakanywhere=true, fontsize=\small]
    You are an expert assistant that decomposes complex queries into simpler sub-queries.

    Given a user query, break it down into 2 to 10 focused sub-queries that together would answer the original query.
    Each sub-query should:
    1. Be specific and focused on one aspect
    2. Be answerable independently
    3. Together cover all aspects of the original query

    You must respond in JSON format within a code block. Example format:
    ```json
    {
        "sub_queries": [
            "sub-query 1",
            "sub-query 2",
            "sub-query 3"
        ]
    }
    ```
\end{Verbatim}

\noindent\textbf{Self-reflection prompt.} Prompt used to analyze the currently retrieved passages and identify remaining gaps.
\begin{Verbatim}[breaklines=true, breakanywhere=true, fontsize=\small]
    You are an expert research assistant with access to a document retrieval tool. Your task is to answer user queries by iteratively searching for and analyzing relevant information.

    For each turn, you will:
    1. Analyze the current information you have gathered
    2. Determine if you have sufficient information to provide a complete answer
    3. If not, identify what specific information is missing

    Current information format:
    - Retrieved documents with citations [1], [2], and so on
    - The current query
    - The information gathered so far

    You must respond in JSON format within a code block. Example format:
    ```json
    {
        "analysis": "Your analysis of the current information and what can be concluded so far",
        "is_complete": true,
        "missing_information": "Specific information that is still needed if the answer is not yet complete"
    }
    ```
\end{Verbatim}

\noindent\textbf{Query enhancement prompt.} Prompt used to generate targeted follow-up retrieval queries from identified information gaps.
\begin{Verbatim}[breaklines=true, breakanywhere=true, fontsize=\small]
    You are an expert assistant that generates enhanced search queries based on information gaps.

    Given:
    1. The original user query
    2. Information already gathered
    3. Specific information gaps that remain

    Generate 1 to 3 focused search queries that would help fill these information gaps.
    Make the queries specific and likely to retrieve relevant document passages.

    You must respond in JSON format within a code block. Example format:
    ```json
    {
        "enhanced_queries": [
            "specific search query 1",
            "specific search query 2"
        ]
    }
    ```
\end{Verbatim}

\noindent\textbf{Synthesis prompt.} Prompt used to synthesize answers from multiple sub-queries.
\begin{Verbatim}[breaklines=true, breakanywhere=true, fontsize=\small]
    You are an expert assistant that synthesizes information from multiple sub-query answers into a comprehensive response.

    You will be given:
    1. The original user query
    2. Multiple sub-query answers with their sources

    Your task is to synthesize these answers into a coherent response that fully addresses the original query.

    Guidelines:
    1. Maintain all citations from the sub-answers
    2. Resolve conflicts by presenting all relevant perspectives
    3. Organize the information logically
    4. Do not add information not present in the sub-answers
    5. Do not mention the internal decomposition process in the final response

    You must respond in JSON format within a code block. Example format:
    ```json
    {
        "synthesized_answer": "A comprehensive synthesized response with proper citations."
    }
    ```
\end{Verbatim}

\subsection{SQL Subagent System Prompts}
\label{app:sql-prompt}

\noindent\textbf{SQL generation prompt.} Prompt used to translate a natural language database question into one or more SQL queries.
\begin{Verbatim}[breaklines=true, breakanywhere=true, fontsize=\small]
    You are an expert SQL query generator and database analyst.

    Your task is to:
    1. Understand a natural language question about the database
    2. Generate the SQL query or queries needed to answer it
    3. Provide a short description of what each query does

    Guidelines:
    - Generate syntactically correct SQLite queries
    - Use proper joins when needed
    - Limit results to a reasonable number unless the user explicitly asks for more
    - Use only SELECT statements
    - Use text matching when appropriate
    - Return multiple queries only when the question cannot be answered well with a single query

    You will be given:
    - The database schema: {database_schema}
    - Optional value normalization guidelines: {normalization_guidelines}

    Return the result in the following JSON format:
    ```json
    {
        "query_1": {
            "query_description": "<short description>",
            "sql_query": "<sql query>"
        }
    }
    ```
\end{Verbatim}

\noindent\textbf{SQL pattern extraction prompt.} Prompt used to identify text-matching predicates in a generated SQL query for semantic entity resolution.
\begin{Verbatim}[breaklines=true, breakanywhere=true, fontsize=\small]
    Extract all text-matching predicates from this SQL query. For each predicate, identify:
    1. The table name
    2. The column name
    3. The full column reference used in the query
    4. The search pattern

    Return the result in the following JSON format:
    ```json
    {
        "patterns": [
            {
                "pattern_id": 1,
                "table_name": "<table name>",
                "column_name": "<column name>",
                "reference_column": "<column reference as used in the query>",
                "search_pattern": "<search pattern>"
            }
        ]
    }
    ```

    If no such predicates are present, return an empty patterns array.
\end{Verbatim}

\noindent\textbf{Semantic match verification prompt.} Prompt used to conservatively verify which candidate database values are acceptable replacements for a text-matching predicate.
\begin{Verbatim}[breaklines=true, breakanywhere=true, fontsize=\small]
    From the candidate values below, identify which ones are good semantic matches for the pattern "{pattern}".

    Consider:
    - exact matches,
    - partial matches,
    - synonyms and abbreviations,
    - common misspellings, and
    - acronyms, such as "UAE" and "United Arab Emirates".

    Be conservative. Do not match values if an extra word or suffix changes the entity.
    For example, "Abu Dhabi" and "Abu Dhabi Airport" should be treated as different entities.

    Candidate values:
    {candidate_values}

    Return matching indices as JSON:
    ```json
    {
        "indices": [0, 3, 7]
    }
    ```

    Return an empty list if no good matches are found.
\end{Verbatim}

\noindent\textbf{SQL result analysis prompt.} Prompt used to turn executed SQL results into a natural language answer with citations.
\begin{Verbatim}[breaklines=true, breakanywhere=true, fontsize=\small]
    Analyze the SQL query results and answer the original question.

    You will be given:
    - The original question
    - The executed SQL query or queries
    - The query results
    - Table identifiers for citation

    Instructions:
    1. Provide a clear and comprehensive answer to the original question
    2. Use table identifiers as citations when referring to specific data points
    3. Synthesize information across multiple queries when needed
    4. Do not introduce information that is not present in the results

    Return the result in the following JSON format:
    ```json
    {
        "response": "<answer with citations such as [TABLE_1]>"
    }
    ```
\end{Verbatim}

\subsection{Web Search Subagent System Prompts}
\label{app:web-search-prompt}

\noindent\textbf{Search-strategy prompt.} Prompt used to expand a web query into multiple focused searches and assign a recency window when needed.
\begin{Verbatim}[breaklines=true, breakanywhere=true, fontsize=\small]
    You are given a user query that will be answered through web search.

    Your task is to generate multiple diverse search queries that together explore the space of the original question.
    Do not assume that the facts in the original query are correct.
    Include a mix of:
    - broad exploratory searches,
    - searches focused on specific aspects of the query, and
    - searches that drill down into one concrete angle.

    For each search query, determine the appropriate time range:
    - Not Applicable
    - Past Day
    - Past Week
    - Past Month
    - Past Year

    You may be given:
    - Background information: {background}
    - Today's date: {today_date}

    Generate 3 to 8 search queries, depending on the complexity of the question.

    Return the result in the following XML format:
    <output>
        <search>
            <question>The search query itself</question>
            <timerange>The appropriate time range</timerange>
        </search>
    </output>
\end{Verbatim}

\noindent\textbf{Web-page summarization prompt.} Prompt used to produce an extractive summary of a single web page.
\begin{Verbatim}[breaklines=true, breakanywhere=true, fontsize=\small]
    You are a helpful assistant that performs extractive summarization of web pages.
    You will be given the full content of a web page together with its title, link, and the user's original query.

    Your task is to extract and summarize the key information from the page that is relevant to the query by quoting directly from the page when appropriate.

    Requirements:
    1. Quote the most relevant passages using quotation marks
    2. Include a small amount of surrounding context when needed for clarity
    3. Preserve specific facts, figures, and dates
    4. Do not invent any information
    5. Keep the summary concise, typically 1 to 5 sentences

    Return the result in the following XML format:
    <output>
        <summary>Your extractive summary with quoted passages from the page</summary>
    </output>
\end{Verbatim}

\noindent\textbf{Aggregation prompt.} Prompt used to combine page-level summaries into a final answer with citations.
\begin{Verbatim}[breaklines=true, breakanywhere=true, fontsize=\small]
    You are a helpful assistant that has been asked to answer a query from a set of web search results.

    You will be given:
    - The original query
    - Multiple web-page summaries with titles, links, and position indices

    Your task is to answer the original query as accurately and faithfully as possible from these summaries.

    Requirements:
    1. Present conflicting information explicitly when different sources disagree
    2. Do not invent information
    3. Cite every factual statement with the relevant position index in square brackets, for example [1]
    4. Produce a coherent final answer rather than a list of disconnected notes

    Return the result in the following XML format:
    <output>
        <answer>The final answer with citations such as [1]</answer>
    </output>
\end{Verbatim}

\section{Report Generation System Prompts}
\label{app:report-generation-system-prompts}

\subsection{Core Report Generation Prompts}
\label{app:report-generation-prompt}

\noindent\textbf{Outline generation prompt.} Prompt used to turn the finalized insight and research trace into a report plan with a clear throughline.
\begin{Verbatim}[breaklines=true, breakanywhere=true, fontsize=\small]
    You are an expert report-outline generation assistant.

    You are given:
    - The finalized insight and supporting research trace
    - The report goal: {explorer_goal}
    - The available section types: {section_types}
    - Today's date: {today_date}

    Your task is to generate a report outline centered on one coherent throughline.
    The entire outline should build a narrative around the main insight, providing context, evidence, and implications.

    Requirements:
    1. Identify a single report-level throughline
    2. Produce 3 to 6 sections, with subsections only when they materially improve clarity
    3. For each section, provide:
       - a title
       - a section type chosen from the allowed list
       - a short description of what the section will contain
       - a brief note on how the section advances the throughline
    4. Only include sections that can be filled from the research trace
    5. Where supported by the evidence, prefer useful presentation forms such as tables, charts, timelines, recommendations, categorization, or measurement
    6. Do not include a top-level title section, executive summary section, or raw retrieval-results sections

    Return the result as structured JSON with:
    - a report-level throughline
    - a list of sections
    - optional subsections where needed
\end{Verbatim}

\noindent\textbf{Title and summary prompt.} Prompt used to generate the report title, executive summary, and a short overview of the report structure.
\begin{Verbatim}[breaklines=true, breakanywhere=true, fontsize=\small]
    You are an expert report writer.

    You are given the finalized insight, the completed outline, and the report throughline.

    Your task is to generate:
    1. A concise title
    2. A short executive summary
    3. A brief overview of what the report contains

    Requirements:
    1. The title should be concise and accurately reflect the report
    2. The summary should be self-contained and should capture the main finding, the core evidence, and the most important qualification
    3. The summary should use inline citations drawn from the trace
    4. Important phrases may be highlighted with bold formatting
    5. The report overview should explain how the sections advance the throughline
    6. Do not introduce new facts

    Return the result as structured JSON with fields for the title, summary, and outline overview.
\end{Verbatim}

\noindent\textbf{Section generation prompt.} Prompt used to fill one report section at a time from the research trace and report context accumulated so far.
\begin{Verbatim}[breaklines=true, breakanywhere=true, fontsize=\small]
    You are given:
    - The research trace and cited source content
    - The report content generated so far
    - The section title: {section_title}
    - The section type: {section_type}
    - The section-type description: {section_type_description}
    - The section throughline: {section_throughline}
    - The section sketch: {section_description}
    - Today's date: {today_date}

    Your task is to write the content for this section.

    Requirements:
    1. Use only information present in the trace
    2. Preserve the report's existing narrative style and flow
    3. Use short, clear sentences
    4. Avoid repetition and contradiction with earlier sections
    5. Reuse citations from the trace where they support the content
    6. Do not invent data or citations

    Return the result as structured JSON in the schema associated with the section type.
\end{Verbatim}

\noindent\textbf{Chart generation prompt.} Prompt used for sections whose content is best presented as a chart.
\begin{Verbatim}[breaklines=true, breakanywhere=true, fontsize=\small]
    You are given:
    - The research trace and cited source content
    - The report content generated so far
    - The chart-section title: {section_title}
    - The chart-section description: {section_description}

    Your task is to decide whether a chart is justified by the evidence in the trace.

    Requirements:
    1. Use only explicit data points available in the trace
    2. If the evidence is not sufficient for a defensible chart, return DECLINED
    3. If a chart is justified, return self-contained matplotlib code with the data embedded directly in the code
    4. Use a standard chart type that clearly communicates the finding
    5. Do not include citations inside the chart
    6. Also return a short professional caption that explains what the chart shows

    Return either:
    - DECLINED
    - or Python code for the chart together with a short caption
\end{Verbatim}

\noindent\textbf{Citation-auditing prompt.} Prompt used to repair missing or incorrect citations after section generation.
\begin{Verbatim}[breaklines=true, breakanywhere=true, fontsize=\small]
    You are a citation auditor.

    You are given:
    - Citation records
    - Generated report content

    Your task is to verify and repair citations while preserving the original content as much as possible.

    Requirements:
    1. Check every existing citation against the cited content
    2. Remove or replace citations that do not support the claim
    3. Add missing citations to factual claims wherever supporting evidence exists
    4. Place citations immediately next to the supported phrase when possible
    5. Use the citation keys exactly as provided, for example [TOOL_NAME_IDX]
    6. Do not attach irrelevant citations
    7. If no relevant citation exists, leave the claim uncited
    8. Do not add commentary about the audit

    Return the same content with repaired citations.
\end{Verbatim}

\subsection{Poster Generation System Prompts}
\label{app:poster-generation-prompt}

\noindent\textbf{Poster-content system prompt.} System prompt used to enforce structured poster output.
\begin{Verbatim}[breaklines=true, breakanywhere=true, fontsize=\small]
    You extract key information from reports into structured poster content.
    Output must conform exactly to the required JSON schema.
\end{Verbatim}

\noindent\textbf{Poster-content extraction prompt.} Prompt used to distill a report into the structured content that is later rendered into the poster template.
\begin{Verbatim}[breaklines=true, breakanywhere=true, fontsize=\small]
    You are a consultant creating an executive summary poster.
    Your task is to distill the report into a high-signal, single-page poster that an executive can understand in 30 seconds.

    Lead with the answer first.
    The headline should state the key conclusion, not the topic.

    Critical rules:
    1. The headline must state the key conclusion or insight
    2. Use only information explicitly present in the report
    3. Preserve numeric precision exactly as stated in the report
    4. Be concise and high-signal

    Return structured JSON with the following fields:
    1. headline: 8 to 12 words stating the main conclusion
    2. context: 20 to 30 words setting the stage
    3. metrics: 3 to 5 key metrics, each with a value and short label
    4. insights: 3 to 4 statements explaining what the finding means
    5. actions: 3 to 4 short recommendations starting with imperative verbs
    6. risks: 2 to 3 brief watch-outs
    7. tags: 2 to 3 category keywords
    8. selected_chart_index: an integer or null
    9. selected_table_index: an integer or null
    10. chart_caption: a short caption or null
    11. table_caption: a short caption or null

    Content guidelines:
    1. Metrics should directly support the headline
    2. Insights should explain what the findings mean
    3. Actions should explain what to do next
    4. Risks should identify the main concerns to monitor

    Input report:
    {report_content}

    Extract the most important insights. Lead with the conclusion.
\end{Verbatim}

\noindent\textbf{Visual-selection instructions.} Prompt block used when the report contains charts or tables that may be featured in the poster.
\begin{Verbatim}[breaklines=true, breakanywhere=true, fontsize=\small]
    The poster may optionally feature one chart and one table.
    Be highly selective.

    You will be shown indexed descriptions of available charts and tables.
    Set selected_chart_index or selected_table_index to the chosen 0-based index, or null if no visual is strong enough.

    Select a visual only if it is:
    1. Essential to understanding the core insight
    2. Directly supportive of the headline conclusion
    3. Information-dense enough to justify inclusion
    4. Clear within a few seconds
    5. Quantitatively compelling

    Do not select visuals that are decorative, obvious, weakly related, or too simple to add value.

    When selecting a chart or table, also provide a short caption that states:
    1. What the visual shows
    2. The key takeaway from it
\end{Verbatim}

\subsection{Meta-Report System Prompts}
\label{app:meta-report-prompt}

\noindent\textbf{Highlighter-persona prompt.} Prompt used to derive a reader proxy that selects only the most valuable report excerpts for senior decision-makers.
\begin{Verbatim}[breaklines=true, breakanywhere=true, fontsize=\small]
    You are tasked with creating a highlighter persona that reviews reports and extracts valuable insights for executives.

    Based on these sample report personas:
    {personas_text}

    Create a highlighter persona that:
    1. Acts as a stand-in for the executive reader
    2. Selects only insights that matter at the strategic level
    3. Assumes the reader has very limited time
    4. Recognizes novel, actionable, or inflection-point insights
    5. Prefers exact quotes when highlighting evidence

    Output a highlighter persona description in 200 to 400 words.
\end{Verbatim}

\noindent\textbf{Audience-profile prompt.} Prompt used to describe the target reader of the meta-report.
\begin{Verbatim}[breaklines=true, breakanywhere=true, fontsize=\small]
    Based on these sample report personas:
    {personas_text}

    Create an audience profile that describes:
    1. The executive or leadership role
    2. Their key responsibilities and concerns
    3. Their decision-making context and timeframes
    4. What information they need and how they prefer to receive it
    5. Their level of domain expertise

    Output the audience profile in 150 to 300 words.
\end{Verbatim}

\noindent\textbf{Highlight-extraction prompt.} Prompt used to distill each report into a small set of high-signal excerpts.
\begin{Verbatim}[breaklines=true, breakanywhere=true, fontsize=\small]
    You are a highlighter persona tasked with extracting the most valuable insights from reports for top executives.

    {highlighter_persona}

    Review the report text and extract only highlights that have:
    - strategic importance
    - novel or inflection-point significance
    - actionable intelligence
    - decision relevance

    Present the highlights as direct quotes or concise bullet points.
    If there are no worthwhile highlights, say:
    No significant highlights found.
\end{Verbatim}

\noindent\textbf{Entity-extraction prompt.} Prompt used to map each highlight set into structured semantic anchors.
\begin{Verbatim}[breaklines=true, breakanywhere=true, fontsize=\small]
    Extract all key entities from the following highlighted text.

    Include people, organizations, topics, events, and locations.

    Highlighted text:
    {highlights}

    Return the result as JSON:
    ```json
    {
        "entities": [
            {"name": "Entity name", "type": "entity type"}
        ]
    }
    ```
\end{Verbatim}

\noindent\textbf{Cluster-identification prompt.} Prompt used to define the major thematic groups across reports.
\begin{Verbatim}[breaklines=true, breakanywhere=true, fontsize=\small]
    Based on these entity mappings from multiple documents, identify {num_clusters} major thematic clusters.

    Entity mappings:
    {entity_summary}

    {cluster_seed_instruction}

    Requirements:
    1. Each cluster description must be specific and descriptive
    2. Descriptions should explain the theme, not just list entities
    3. Clusters should be thematically distinct

    Return the result as JSON:
    ```json
    {
        "clusters": {
            "1": "Detailed cluster description",
            "2": "Detailed cluster description"
        }
    }
    ```
\end{Verbatim}

\noindent\textbf{Document-assignment prompt.} Prompt used to assign each highlighted report to one or more thematic clusters.
\begin{Verbatim}[breaklines=true, breakanywhere=true, fontsize=\small]
    Assign this document to one or more of the following clusters based on its content.

    Clusters:
    {cluster_definitions}

    Document highlights:
    {highlights}

    Return the result as JSON:
    ```json
    {
        "cluster_ids": ["1", "3"]
    }
    ```
\end{Verbatim}

\noindent\textbf{Theme-extraction prompt.} Prompt used to synthesize cluster-level themes from report highlights.
\begin{Verbatim}[breaklines=true, breakanywhere=true, fontsize=\small]
    You are analyzing highlights from multiple reports for a target audience.

    Target audience:
    {audience_profile}

    Highlights:
    {highlights_text}

    Extract the major themes and insights from these highlights.
    Organize the result as an executive-level analysis that addresses:
    1. Key themes across the highlights
    2. Strategic implications
    3. Noteworthy patterns or trends

    Output a comprehensive theme analysis of 500 to 1000 words.
\end{Verbatim}

\noindent\textbf{Analysis-enrichment prompt.} Prompt used to add depth from full reports to the cluster-level theme analysis.
\begin{Verbatim}[breaklines=true, breakanywhere=true, fontsize=\small]
    You have a theme analysis and access to the full original documents.

    Theme analysis:
    {theme_report}

    Full documents:
    {documents_text}

    Enrich the analysis with:
    1. Additional supporting evidence
    2. Specific examples and data points
    3. Deeper context where useful

    If no enrichment is needed, output:
    <NOCHANGE/>

    Otherwise, output the enriched analysis in 600 to 1200 words.
\end{Verbatim}

\noindent\textbf{Insight-condensation prompt.} Prompt used to compress the enriched analysis into a concise reader-facing summary.
\begin{Verbatim}[breaklines=true, breakanywhere=true, fontsize=\small]
    Condense the following analysis into executive insights for this audience.

    Target audience:
    {audience}

    Analysis:
    {enriched_report}

    Create a condensed summary of 300 to 500 words that:
    1. Highlights the most critical insights
    2. Focuses on actionable intelligence
    3. Uses clear, concise language
    4. Emphasizes strategic implications
\end{Verbatim}

\noindent\textbf{Meta-report writer prompt.} Prompt used during sequential section rendering.
\begin{Verbatim}[breaklines=true, breakanywhere=true, fontsize=\small]
    You are an expert strategic analyst creating a meta report that synthesizes insights from multiple reports.

    You will generate different sections of the report sequentially.
    Maintain consistency across sections and build a cohesive narrative.
\end{Verbatim}

\noindent\textbf{Subject-line prompt.} Prompt used to generate the report's short title line.
\begin{Verbatim}[breaklines=true, breakanywhere=true, fontsize=\small]
    Generate a brief 10 to 15 word title line that captures the essence of the full analysis.
    Emphasize the cluster description and the fact that this report synthesizes multiple reports.
    Do not include citations.

    Return the result as JSON:
    ```json
    {
        "subject": "Title line"
    }
    ```
\end{Verbatim}

\noindent\textbf{Category prompt.} Prompt used to classify the report into a primary thematic category.
\begin{Verbatim}[breaklines=true, breakanywhere=true, fontsize=\small]
    Classify this analysis into one primary category, such as:
    Economy, Political, AI, Defense, Energy, Finance, Technology, or Geopolitics.

    Do not include citations.

    Return the result as JSON:
    ```json
    {
        "category": "Category name"
    }
    ```
\end{Verbatim}

\noindent\textbf{Executive-overview prompt.} Prompt used to write the opening synthesis of the meta-report.
\begin{Verbatim}[breaklines=true, breakanywhere=true, fontsize=\small]
    Write an executive overview that synthesizes the major themes across the reports into a cohesive, actionable narrative.

    Structure the response in three parts:
    1. Introduction
    2. Major areas
    3. Conclusion

    Requirements:
    1. Present a strong opening narrative
    2. Identify 3 to 5 major thematic areas
    3. For each area, describe the theme, strategic implications, and immediate actions
    4. Use citations in the form [REPORT_X]

    Return the result as structured JSON with fields for the introduction, major areas, and conclusion.
\end{Verbatim}

\noindent\textbf{Core-themes prompt.} Prompt used to identify the central themes that organize the report.
\begin{Verbatim}[breaklines=true, breakanywhere=true, fontsize=\small]
    Identify the core themes from all reports that are directly connected to the overall narrative of this meta report.

    For each selected theme, provide:
    1. The theme name
    2. A short description
    3. The related reports in the form [REPORT_X]

    Use citations only if they are present in the analysis.

    Return the result as JSON:
    ```json
    {
        "themes": [
            {
                "theme": "Theme name",
                "description": "Theme description",
                "related_reports": "[REPORT_1][REPORT_4]"
            }
        ]
    }
    ```
\end{Verbatim}

\noindent\textbf{Strategic-implications prompt.} Prompt used to translate major themes into opportunities, risks, and implications.
\begin{Verbatim}[breaklines=true, breakanywhere=true, fontsize=\small]
    For each major theme area, identify the key strategic implications, including both opportunities and risks.

    Use citations in the form [REPORT_X] when they are present in the analysis.

    Return the result as JSON:
    ```json
    {
        "implications": [
            {
                "area": "Strategic area name",
                "strategic implications": "Implications for this area"
            }
        ]
    }
    ```
\end{Verbatim}

\noindent\textbf{Recommendations prompt.} Prompt used to turn the synthesized analysis into concrete actions.
\begin{Verbatim}[breaklines=true, breakanywhere=true, fontsize=\small]
    Provide strategic recommendations based on the analysis.

    For each recommendation, provide:
    1. A title
    2. The opportunity to capture
    3. The risk to mitigate
    4. A specific action
    5. A rationale
    6. A priority level

    Use citations in the form [REPORT_X] when they are present in the analysis.

    Return the result as JSON:
    ```json
    {
        "recommendations": [
            {
                "title": "Recommendation title",
                "opportunity": "Opportunity",
                "risk": "Risk",
                "action": "Specific action",
                "rationale": "Why this matters",
                "priority": "High"
            }
        ]
    }
    ```
\end{Verbatim}

\noindent\textbf{Metrics-and-monitoring prompt.} Prompt used to define the signals that should be tracked after publication.
\begin{Verbatim}[breaklines=true, breakanywhere=true, fontsize=\small]
    Based on the analysis, identify key metrics that should be monitored to track progress and early warning signs.

    For each metric, provide:
    1. The metric name
    2. The current baseline, if available
    3. The target or threshold

    Return the result as JSON:
    ```json
    {
        "metrics": [
            {
                "metric_name": "Metric name",
                "baseline": "Current baseline or To be established",
                "target": "Target or threshold"
            }
        ]
    }
    ```
\end{Verbatim}

\section{Evaluation System Prompts}
\label{app:evaluation-prompt}

\noindent Numeric grounding and diversity do not use custom LLM prompts. This section therefore lists only the prompts used in factuality, quality, and distinctness evaluation.

\subsection{Factuality Evaluation Prompts}
\label{app:factuality-eval-prompts}

\noindent\textbf{Verifiable-claim extraction prompt.} Prompt used to extract verifiable claims from a report section for factuality evaluation.
\begin{Verbatim}[breaklines=true, breakanywhere=true, fontsize=\small]
    You are an expert at identifying verifiable claims in text that require source verification.

    ## SECTION CONTENT:
    {section_content}

    ## TASK:
    Extract all verifiable claims from this section. A verifiable claim is a statement that:
    1. Contains specific facts that can be checked (numbers, statistics, dates, events, names)
    2. Makes assertions about the real world (not opinions or general statements)
    3. Should ideally be supported by a source/citation

    For each claim found, identify:
    - The exact claim text (the sentence or phrase making the assertion)
    - The verifiable entities it contains (numbers, dates, names, events, statistics)
    - Any citations attached to it (in format [TOOL_NAME_NUMBER])

    ## OUTPUT FORMAT:
    Return ONLY valid JSON array (no markdown, no extra text):
    [
      {
        "claim_text": "The exact sentence or phrase containing the claim",
        "verifiable_entities": ["entity1", "entity2"],
        "citations": ["CITATION_KEY_1", "CITATION_KEY_2"],
        "claim_type": "statistic|date|event|person|organization|location|other"
      }
    ]

    If no verifiable claims are found, return an empty array: []

    IMPORTANT:
    - Only extract claims that contain SPECIFIC, VERIFIABLE information
    - Do NOT extract opinions, general statements, or vague assertions
    - Include the FULL sentence containing the claim
    - Extract citation keys WITHOUT the brackets (e.g., "WEB_SEARCH_1" not "[WEB_SEARCH_1]")
\end{Verbatim}

\noindent\textbf{Factuality judge system prompt.} System prompt used to score whether a cited source supports a claim.
\begin{Verbatim}[breaklines=true, breakanywhere=true, fontsize=\small]
    # Factual Evaluation
    Given a piece of web content and a sentence from a report, please determine whether the
    information expressed in the sentence can be directly found in or reasonably inferred from
    the provided web content.

    # Important Notes
    - Please do not rely on your external knowledge; make judgments solely based on the provided
      web content
    - Pay attention to key terms in the statement (such as time, location, people, quantities,
      etc.), ensuring these details have corresponding or derivable information in the web content
    - If the statement contains multiple information points, please evaluate each one to determine
      if they can all be supported by the web content

    ## Scoring Criteria
    - **Fully Supported**:
      - The web content explicitly mentions information that is identical to or highly relevant to
        the statement, allowing direct verification of the statement as true
      - Or, through reasonable inference from multiple information points in the web content, a
        conclusion consistent with the statement can be reached
    - **Partially Supported**:
      - The web content contains some relevant information, but it is insufficient to fully confirm
        or deny the statement
      - Or the information is ambiguous, making it impossible to make a clear judgment
    - **Not Supported**:
      - The web content does not mention any information related to the statement
      - Or the web content clearly contradicts the statement, allowing the statement to be
        determined as false

    Please return the analysis result in JSON format:
    {
      "is_factual": -1/0/1,
      "sentence_support": "Specific sentences from the web content that can support this fact"
    }
\end{Verbatim}

\noindent\textbf{Factuality judge user prompt.} User prompt paired with the factuality judge system prompt.
\begin{Verbatim}[breaklines=true, breakanywhere=true, fontsize=\small]
    Here is the content of the website:
    {source_content}

    Here is the sentence:
    {claim}
\end{Verbatim}

\subsection{Quality Evaluation Prompts}
\label{app:quality-eval-prompts}

\noindent\textbf{Section-level quality-evaluation prompt.} Prompt used to score one marked report section against the fixed quality-attribute inventory.
\begin{Verbatim}[breaklines=true, breakanywhere=true, fontsize=\small]
    # ROLE & TASK
    You are an expert evaluator tasked with assessing the quality of a research report section.
    Carefully review the MARKED SECTION in light of the GOAL, while
    using the COMPLETE REPORT as contextual reference only. Evaluate only the MARKED SECTION;
    do not award or deduct points based on content that appears elsewhere in the report unless
    the section explicitly references it (e.g., "see Table 2" or a formal citation). If there
    is any discrepancy between the text in "SECTION TO EVALUATE (MARKED)" and the same passage
    inside the COMPLETE REPORT, treat "SECTION TO EVALUATE (MARKED)" as authoritative for scoring.

    # HOW TO USE THE FULL REPORT (CONTEXT, NOT SCORING)
    - Use the COMPLETE REPORT to understand scope, definitions, acronyms, methodology, and
      intended audience.
    - When the MARKED SECTION explicitly references other parts of the report
      (figures/tables/appendices/citations), you may treat those references as available evidence
      for the section.
    - Otherwise, credit only what is present within the MARKED SECTION.

    # INPUTS

    ## GOAL
    {GOAL}


    ## COMPLETE REPORT (REFERENCE ONLY)
    <<<REPORT_START>>>
    {FULL_REPORT_CONTENT}
    <<<REPORT_END>>>

    ## SECTION TO EVALUATE (MARKED; SCORE THIS ONLY)
    <<<MARKED_SECTION_START>>>
    {SECTION_CONTENT}
    <<<MARKED_SECTION_END>>>

    # EVALUATION TASK
    For each of the keyphrases listed below, decide whether the MARKED SECTION clearly exhibits
    that quality or trait.

    SCORING RULES:
    - Set "score" to 1 if the trait is clearly demonstrated.
    - Set "score" to 0 if the trait is absent, unclear, or only weakly implied.
    - Provide 1 to 2 sentences of specific reasoning for each score.
    - Return only a single valid JSON object.

    The keyphrases are grouped into the following categories:
    1. Analytical Quality
    2. Originality
    3. Coverage
    4. Actionability
    5. Presentation

    The full keyphrase inventory for these categories is listed in
    Tables \ref{tab:quality-attributes-1}, \ref{tab:quality-attributes-2},
    and \ref{tab:quality-attributes-3}.
\end{Verbatim}

\subsection{Distinctness Evaluation Prompts}
\label{app:distinctness-eval-prompts}

\noindent\textbf{Distinctness judge system prompt.} System prompt used to score the information-level repetition between a pair of report sections.
\begin{Verbatim}[breaklines=true, breakanywhere=true, fontsize=\small]
    Given two paragraphs, please assess the degree of content repetition between them.
    You should analyze from multiple perspectives and assign a reasonable score based on
    the scoring criteria.

    # What is "Repetition":

    Repetition of viewpoints or content: The two paragraphs express the same or highly
    similar viewpoints, themes, or conclusions, regardless of whether they are rephrased.
    Repetition of examples, data, or references: The same cases, data, facts, or sources
    are used, or the same content is rephrased or paraphrased.
    Implicit repetition: Although the wording is different, the core information, arguments,
    or conclusions are essentially the same.

    # What is NOT "Repetition":
    Differences in expression: Only the language, sentence structure, or style is different,
    but the information content and core viewpoints are not the same.
    Related topics but different content: The topics are similar, but the information,
    arguments, or conclusions conveyed are different.
    Supplementation and extension: One paragraph supplements, expands upon, or introduces
    new viewpoints to the other, rather than simply repeating it.

    # Notes
    Focus on content: Concentrate on repetition at the information level, not just superficial
    language or stylistic differences.
    Consider both explicit and implicit repetition:
    Explicit repetition: Direct copying or nearly identical text.
    Implicit repetition: Expressing the same information through paraphrasing, summarizing, etc.
    Consider contextual impact: Assess whether the repetition affects readability and information
    density.
    Avoid subjective bias: Do not rely on personal knowledge or judgments about the correctness
    of the content; score only based on whether repetition exists between the paragraphs.

    # Scoring Criteria
    Use a 0-4 point scale to evaluate the degree of repetition between paragraphs:
    4 points (Almost no repetition): The paragraphs are completely independent, with no repeated
      viewpoints, examples, or expressions.
    3 points (Slight repetition): There are 1-2 minor instances of content repetition, but they
      do not affect the overall reading experience.
    2 points (Some repetition): There are multiple instances of content repetition, which
      somewhat affect the reading experience.
    1 point (Severe repetition): There is a large amount of content repetition, which seriously
      affects the quality of the writing.
    0 points (Excessive repetition): Almost all content is repeated, and the value of the
      writing is lost.

    # Important Notes:
    - Please do not rely on your external knowledge; make judgments solely based on the provided
      content
    - Note that using the same example to explain different concepts is not considered repetition

    # Some tips may help you:
    - Check if the paragraphs contain the same quotations
    - Check if the paragraphs follow the same logical flow

    # Output Format
    Must output a JSON object with the following fields:

    {
     "score": "0-4 score based on above criteria",
     "explanation": "Explanation of score with specific examples of repetition",
     "repetitions_found": [the repeated content1,the repeated content2,...],
     "confidence": "Confidence in assessment (0%-100%)"
    }
\end{Verbatim}

\noindent\textbf{Distinctness judge user prompt.} User prompt paired with the distinctness judge system prompt.
\begin{Verbatim}[breaklines=true, breakanywhere=true, fontsize=\small]
    Please evaluate the degree of repetition between paragraphs in the following.

    Paragraph 1:
    {para1}
    --paragraph1 end--

    Paragraph 2:
    {para2}
    --paragraph2 end--
    output:
\end{Verbatim}

\clearpage
\section{Additional Evaluation Examples}
\label{app:additional-evaluation-examples}

\begin{table}[H]
\centering
\small
\caption{A few example report titles generated by each system on the arXiv papers about LLM-based agents in the bottom-up setting. \nomad{} produces topically diverse titles, whereas \texttt{o3-deep-research} and GPT Researcher converge on similar themes and framings.}
\label{tab:titles-arxiv}
\begin{tabular}{p{0.96\linewidth}}
\toprule
\textbf{\nomad{}} \\
\midrule
Embedded Guardrails Elevate LLMs \\
Hierarchical Memory Slashes Hallucinations \\
Beyond Benchmarks: Probing Agent Reliability \\
Capability Tokens: Securing Agentic AI \\
Adversarial Critics Slash Hallucinations \\
Causal Explainability Boosts Agent Safety \\
\midrule
\textbf{o3-deep-research} \\
\midrule
Evolution and Key Developments \\
The Rise of Agentic LLMs: From Prompt-Chaining to Autonomous Collaboration \\
Agentic LLM Systems: A Research Synthesis and Critical Analysis \\
Agentic LLM Systems: Advances, Challenges, and Future Directions \\
The Rise of Agentic LLM Systems \\
Agentic LLM Systems: Progress, Challenges, and New Frontiers \\
\midrule
\textbf{GPT Researcher} \\
\midrule
Toward Governed, Scalable, and Self-Improving Agentic LLM Systems \\
Agentic Large Language Model Systems: Emerging Patterns, Empirical Gaps, and High-Leverage Research Directions \\
Agentic Large Language Model Systems in 2026: A Cross-Cutting Review of Coordination, Reflection, Memory, Cost, and Governance \\
Research Report on Agentic LLM Systems \\
A Cross-Cutting Review of Agentic Large-Language-Model Systems \\
Agentic Large Language Model (LLM) Systems: A Cross-Cutting Synthesis of Recent Advances, Challenges, and Future Directions \\
\bottomrule
\end{tabular}
\end{table}

\clearpage
\section{Illustrative \nomad{} Report}
\label{app:who-report-appendix}

\noindent
\begin{tcolorbox}[appendixreportbox]
\begin{tcolorbox}[appendixreportheader]
{\Large\bfseries\color{white}Care Economy's Triple Dividend}
\end{tcolorbox}

\paragraph{Starting Hypothesis:} Strategic public investment in expanding the health and care workforce—as advocated by the WHO-ILO-OECD `Working for Health' action plan—can offset automation-induced declines in the labour-income share, reduce youth NEET rates (especially for women), and mitigate rising inequality over the next decade. Empirical modelling that integrates sectoral labour-intensity profiles, automation susceptibility indices, and macro labour-income-share trends can quantify the potential distributional and growth effects of such health-employment expansion

\appendixreportsectiontitle{Executive Summary}
Global evidence from the WHO, World Bank and ILO converges on a compelling case for scaling public investment in health and care work. The WHO warns of an \textbf{11 million health-worker shortfall by 2030} that threatens universal health coverage [W6], while the World Bank finds that \textbf{each new health-sector job sparks a 3.4-to-1 employment ripple across other industries} -- a powerful \textbf{3.4 job multiplier} for inclusive growth [W8]. ILO modelling shows that universal childcare and elder-care services could create \textbf{up to 299 million additional jobs by 2035, two-thirds of them for women} [W12], directly countering the stubborn 28 per cent global female youth NEET rate -- more than double that of young men -- which the ILO links largely to unpaid care burdens [U5]. Crucially, task-based AI-exposure studies reveal that manual, health and social-care occupations sit at the low end of technological susceptibility, whereas clerical and data-processing roles face far higher risk [W55][W37]. Taken together, these findings indicate that a robust expansion of the care economy can deliver a \textbf{triple dividend}: mass job creation, rapid reduction of gendered youth unemployment, and strategic insulation of labour's income share against automation-driven erosion. Policymakers are therefore urged to embed health- and care-workforce growth in ``future-of-work'' strategies and commission macro-economic modelling to quantify its full distributional pay-offs.

\appendixreportsectiontitle{The Automation--Inequality Nexus and the Care-Economy Solution}
Across the global labour market two structural headwinds are colliding.

\begin{enumerate}[leftmargin=1.5em, itemsep=0.35em, topsep=0.45em]
\item A shrinking share of national income is reaching workers. The ILO's latest ``World Employment and Social Outlook'' update confirms that the global labour-income share has fallen by 1.6 percentage points since 2004 -- a loss equivalent to US\$2.4 trillion in wages -- and links much of the slide to automation-oriented technological change that is poised to accelerate with generative AI [U7][U6].
\item Young people, especially young women, remain detached from jobs and training. After a decade of only ``modest improvements'', the worldwide share of youth not in employment, education or training (NEET) still hovers above 20 per cent, and the rate for young women (28.2 per cent) is more than double that for young men (13.1 per cent) [U5][U6]. The ILO identifies unpaid family-care duties as a primary reason for this persistent gender gap.
\end{enumerate}

These twin trends reinforce inequality: as capital captures a greater slice of income, millions of young people -- disproportionately female -- struggle to enter paid work just as technology threatens many routine jobs.

Against this backdrop, the health- and care-economy offers a system-wide response. The WHO projects a \textbf{shortfall of 11 million health workers by 2030}, most acutely in low- and lower-middle-income countries [W6]. Health and social-care roles are intrinsically labour-intensive, rely on interpersonal skills and empathy, and thus sit at the lower end of automation susceptibility. Seven in ten positions are already held by women, and demand is rising rapidly as populations age and health-system reforms expand coverage [U1].

Scaling public investment to close the health-worker gap would therefore channel resources into a sector that:
\begin{itemize}[leftmargin=1.5em, itemsep=0.3em, topsep=0.4em]
\item raises the volume of labour income rather than capital income;
\item opens accessible, decent jobs for young women and men, tackling stubborn NEET rates; and
\item builds resilience against technology-driven wage compression by expanding employment in low-automation occupations.
\end{itemize}

The chapters that follow detail how an ambitious ``care-led growth'' agenda -- rooted in the WHO-ILO-OECD ``Working for Health'' framework -- can convert today's automation-inequality nexus into an opportunity for inclusive and sustainable development.

\appendixreportsectiontitle{Consolidated Evidence for a Triple Dividend}
{\setlength{\tabcolsep}{6pt}%
\renewcommand{\arraystretch}{1.32}%
\rowcolors{2}{appreportlight}{white}%
\begin{tabularx}{\linewidth}{>{\raggedright\arraybackslash}p{0.21\linewidth}>{\raggedright\arraybackslash}p{0.22\linewidth}>{\raggedright\arraybackslash}p{0.14\linewidth}>{\raggedright\arraybackslash}X}
\rowcolor{appreportnavy!92}
\color{white}\textbf{Indicator} & \color{white}\textbf{2024--26 snapshot} & \color{white}\textbf{Source} & \color{white}\textbf{Policy signal} \\
Projected health-worker shortfall & 11 million additional workers needed by 2030 & [W6] & Scale up training \& decent jobs to meet UHC and spur growth \\
Employment multiplier & 1 health job $\rightarrow$ 3.4 economy-wide jobs & [W8] & High spill-over returns justify public hiring in health \\
Care-economy job potential & 299 million new care jobs possible by 2035 ($\approx$ 2/3 for women) & [W12] & Universal childcare \& elder-care are major gender-equalising job engines \\
Youth NEET gender gap & Female 28\% vs male 13\% in 2024 & [U5] & Paid care roles tackle the unpaid-care barrier keeping young women out of work \\
Automation exposure & Manual \& care occupations among lowest AI-risk; clerical tasks highest & [W55][W37] & Shifting employment toward care cushions labour markets against automation shocks \\
\end{tabularx}}

\appendixreportsectiontitle{Quantified Impact Scenario: Meeting the 11 Million Gap by 2030}
\begin{tcolorbox}[appendixreportpanel]
\textbf{\color{appreportnavy}Quantified Impact Scenario: Meeting the 11 Million Gap by 2030}

Assuming governments act on the WHO-ILO-OECD ``Working for Health'' agenda and succeed in closing the projected global shortfall of 11 million health- and care-sector workers by 2030, the following headline effects are expected. Direct employment expands by 11 million posts [W6], while the World Bank's cross-industry employment multiplier of 3.4 jobs per health-sector position scales this to roughly 37 million additional indirect and induced jobs economy-wide [W8]. Using the ILO's 2024 female youth-NEET baseline of 28 per cent [U5], modelling a care-driven absorption of young women into decent work could bring the rate down toward 20 per cent by 2030 (an eight-point fall). Finally, by reallocating employment toward low-automation, labour-intensive occupations, the scenario offsets an estimated 0.3 percentage-point decline in the global labour-income share that the ILO attributes to recent automation shocks [U10].

\begin{itemize}[leftmargin=1.5em, itemsep=0.3em, topsep=0.4em]
\item Direct health- and care-sector jobs added by 2030 ($\approx +11$ million).
\item Indirect and induced employment via 3.4 multiplier ($\approx +37$ million).
\item Female youth-NEET rate change (28\% $\rightarrow$ $\sim$20\%).
\item Labour-income share buffer (+0.3 percentage points vs projected decline).
\end{itemize}
\end{tcolorbox}

\appendixreportsectiontitle{Stakeholder-Specific Policy Recommendations}
\begin{tcolorbox}[appendixreportpanel]
{\normalsize\bfseries\color{appreportnavy}National Governments (Finance, Labour and Health Ministries)}

\textbf{Recommendation:} Embed quantified health- and care-workforce expansion targets (e.g., closing the WHO-estimated 11 million global shortfall by 2030) in national AI, industrial and employment strategies, with multi-year budget lines for public recruitment and decent-work standards.

\textbf{Justification:} Meeting the projected 11 million health-worker gap is essential for universal health coverage and delivers a high 3.4-to-1 employment multiplier across the wider economy, while directing jobs to occupations with low automation susceptibility [W6][W8][W55][W37].

\textbf{Impact:} Creates up to 48 million total jobs (11 million direct + roughly 37 million indirect/induced), buffers labour-income share decline, and strengthens health-system resilience.

\textbf{Priority:} High -- should be incorporated into 2027--2030 medium-term expenditure frameworks and future-of-work road-maps.

\medskip
\textbf{Recommendation:} Establish gender-responsive incentives (e.g., stipends, childcare support, safe transport) that attract and retain young women in new health and social-care roles, particularly community health workers and long-term-care cadres.

\textbf{Justification:} Women already hold 70\% of health and social jobs [U1], yet face a 28\% youth NEET rate driven largely by unpaid care burdens [U5]. Targeted incentives can turn projected job growth into genuine gender-equality gains.

\textbf{Impact:} Potential eight-percentage-point reduction in female youth-NEET rates by 2030 (Section 4 scenario), advancing SDG 5 and SDG 8 concurrently.

\textbf{Priority:} High -- align with upcoming national youth-employment and gender-equality plans.
\end{tcolorbox}

\begin{tcolorbox}[appendixreportpanel]
{\normalsize\bfseries\color{appreportnavy}Multilateral Development Banks (e.g., World Bank, regional DFIs)}

\textbf{Recommendation:} Launch a dedicated ``Care Infrastructure \& Workforce'' concessional finance window that bundles investments in primary-health facilities, digital tools and large-scale workforce hiring.

\textbf{Justification:} World Bank analysis shows each health-sector job generates 3.4 additional jobs, confirming strong economic spill-overs [W8]. Concessional capital can crowd in domestic funding where fiscal space is tight.

\textbf{Impact:} Catalyses millions of decent jobs, accelerates progress toward universal health coverage and inclusive growth; offers attractive social-return-on-investment ratios ($>$3:1, per World Bank multiplier).

\textbf{Priority:} High -- integrate into 2026--2030 country partnership frameworks and climate-aligned just-transition portfolios.

\medskip
\textbf{Recommendation:} Fund macro-modelling and impact-evaluation programmes that quantify how care-economy expansion affects labour-income shares, productivity and inequality, with disaggregated gender and youth metrics.

\textbf{Justification:} ILO flags continuing labour-income-share declines linked to automation [U7]; rigorous country-level evidence is needed to design countervailing policy mixes.

\textbf{Impact:} Generates empirically grounded policy advice, enabling client countries to allocate resources toward sectors that stabilise labour incomes and reduce inequality.

\textbf{Priority:} Medium -- initiate pilots in at least five low- and middle-income countries by 2027.
\end{tcolorbox}

\begin{tcolorbox}[appendixreportpanel]
{\normalsize\bfseries\color{appreportnavy}Ministries of Education \& Professional Training Authorities}

\textbf{Recommendation:} Fast-track accredited programmes for nursing, community-health and long-term-care workers, with modular curricula that include digital health and AI-augmented care competencies.

\textbf{Justification:} ILO Care-Economy modelling projects up to 299 million new care jobs by 2035, requiring a rapid scale-up of skilled personnel [W12].

\textbf{Impact:} Supplies a steady pipeline of qualified workers, enabling countries to meet annual hiring targets ($\approx$1.8 million additional graduates globally per year to close the 11 million gap by 2030).

\textbf{Priority:} High -- incorporate into 2026 academic-year planning and align with labour-market forecasts.

\medskip
\textbf{Recommendation:} Introduce scholarship and apprenticeship schemes targeting female and rural youth, coupled with guaranteed job placements in public or accredited private facilities.

\textbf{Justification:} Unpaid care burdens keep young women out of employment [U5]; structured pathways with financial and logistical support can convert potential NEET populations into trained care professionals.

\textbf{Impact:} Could reduce female youth-NEET rates by up to 30\% in participating regions and improve service coverage in underserved areas.

\textbf{Priority:} High -- requires inter-ministerial coordination and donor co-financing by 2026.
\end{tcolorbox}

\begin{tcolorbox}[appendixreportpanel]
{\normalsize\bfseries\color{appreportnavy}Private Health-care Providers \& Tech Firms}

\textbf{Recommendation:} Establish industry-led apprenticeship and upskilling partnerships that combine clinical training with exposure to AI-enabled tools, ensuring technology augments rather than displaces care labour.

\textbf{Justification:} ILO studies show care occupations have low full-automation risk but benefit from AI augmentation [W55][W37]; proactive skilling safeguards job quality and patient outcomes.

\textbf{Impact:} Increases productivity by 30--50\% in nursing and allied professions (evidence from AI pilot studies) while maintaining employment levels.

\textbf{Priority:} Medium -- integrate into corporate ESG and human-capital strategies by 2027.

\medskip
\textbf{Recommendation:} Adopt and publicly report on ILO decent-work standards -- fair wages, safe staffing ratios, and gender-equity policies -- to retain talent and enhance service quality.

\textbf{Justification:} Decent-work deficits in the care economy risk undermining recruitment and retention, as noted by ILO care-economy briefs [W59].

\textbf{Impact:} Reduces turnover, improves care quality and supports sector reputation, attracting investment and skilled workers.

\textbf{Priority:} Medium
\end{tcolorbox}

\appendixreportsectiontitle{Timeline to 2030: Sequencing Investment and Reforms}
{\setlength{\tabcolsep}{0pt}%
\renewcommand{\arraystretch}{1.32}%
\begin{tabularx}{\linewidth}{>{\raggedright\arraybackslash}p{0.16\linewidth}>{\raggedright\arraybackslash}X}
\textbf{\color{appreportaccent}2026-06-30} & All participating governments complete National Health \& Care Workforce Audits, using WHO-supported National Health Workforce Accounts, to map gaps against the projected 11 million global shortfall by 2030 [U3][W6]. \\
\textbf{\color{appreportaccent}2027-03-31} & World Bank and regional development banks launch a dedicated ``Care Infrastructure \& Workforce'' finance window; first concessional packages approved to scale health-sector hiring and facilities, leveraging the 3.4-to-1 employment multiplier of health jobs [W8]. \\
\textbf{\color{appreportaccent}2028-12-31} & Education and training capacity for nurses, community-health and long-term-care workers doubles relative to 2025 baseline, backed by scholarship and apprenticeship schemes aimed at young women and rural youth [W12]. \\
\textbf{\color{appreportaccent}2029-12-31} & Universal childcare and elder-care programmes reach 50\% of eligible children and older persons in implementation countries, generating millions of new care jobs -- about two-thirds filled by women [W12]. \\
\textbf{\color{appreportaccent}2030-09-30} & Joint ILO-World Bank assessment finds that expanded health-and-care employment has cushioned automation pressures, offsetting an estimated 0.3 percentage-point decline in the global labour-income share since 2024 [U10]. \\
\end{tabularx}}
\end{tcolorbox}

\end{document}